\documentclass[sigconf]{acmart}
\acmSubmissionID{60}

\usepackage{amsmath}
\usepackage{mathrsfs}
\usepackage[linesnumbered,ruled,vlined]{algorithm2e}
\usepackage{booktabs}
\usepackage{multirow}
\usepackage{makecell}
\usepackage{float}
\usepackage{afterpage}
\usepackage{lipsum}
\usepackage{subfigure}
\usepackage{stfloats}
\usepackage{tcolorbox}
\usepackage{titlesec}
\usepackage{tabularx}
\definecolor{salmon}{HTML}{f8bec0}
\newtcbox{\besthighlight}[1][salmon]{
  on line, arc=0.75mm, colback=#1, colframe=#1, boxrule=0pt, boxsep=0pt, left=1.5pt, right=1.5pt, top=1.5pt, bottom=1.5pt
}

\definecolor{skyblue}{HTML}{bbd0e8}
\newtcbox{\secondhighlight}[1][skyblue]{
  on line, arc=0.75mm, colback=#1, colframe=#1, boxrule=0pt, boxsep=0pt, left=1.5pt, right=1.5pt, top=1.5pt, bottom=1.5pt
}

\definecolor{springgreen}{HTML}{d2f0ca}
\newtcbox{\thirdhighlight}[1][springgreen]{
  on line, arc=0.75mm, colback=#1, colframe=#1, boxrule=0pt, boxsep=0pt, left=1.5pt, right=1.5pt, top=1.5pt, bottom=1.5pt
}

\AtBeginDocument{
  \providecommand\BibTeX{{
    \normalfont B\kern-0.5em{\scshape i\kern-0.25em b}\kern-0.8em\TeX}}}

\setcopyright{acmlicensed}
\copyrightyear{2024}
\acmYear{2024}
\acmDOI{XXXXXXX.XXXXXXX}

\renewcommand\footnotetextcopyrightpermission[1]{}

\begin{document}

\title{\textbf{RDSA:} A Robust Deep Graph Clustering Framework via Dual Soft Assignment}

\author{
  \href{https://orcid.org/0009-0007-9693-7963}{Yang Xiang}\footnotemark[1]\textsuperscript{\rm 1, 2}, 
  \href{https://orcid.org/0009-0005-7234-5774}{Li Fan}\footnotemark[1]\textsuperscript{\rm 1}, 
  \href{https://orcid.org/0000-0002-3252-0997}{Tulika Saha}\textsuperscript{\rm 2},
  \href{https://orcid.org/0000-0003-3389-2263}{Xiaoying Pang}\textsuperscript{\rm 1}, 
  \href{https://orcid.org/0000-0002-6877-3937}{Yushan Pan}\textsuperscript{\rm 1}, 
  \href{https://orcid.org/0000-0002-3693-9808}{Haiyang Zhang}\textsuperscript{\rm 1}, 
  \href{https://orcid.org/0000-0001-5733-6881}{Chengtao Ji}\footnotemark[2]\textsuperscript{\rm 1}
}
\affiliation{%
  \institution{
    \textsuperscript{\rm 1} Department of Computing, Xi'an Jiaotong-Liverpool University, Suzhou \country{China}
    }
}
\email{{Yang.Xiang19, Li.Fan21}@student.xjtlu.edu.cn}
\email{{Xiaoying Pang, Yushan.Pan, Haiyang.Zhang, Chengtao.Ji}@xjtlu.edu.cn}
\affiliation{%
  \institution{
    \textsuperscript{\rm 2} Department of Computer Science, University of Liverpool, Liverpool \country{United Kingdom}
  }
}
\email{{Y.Xiang17, Tulika.Saha}@liverpool.ac.uk}

\renewcommand{\authors}{Yang Xiang, Li Fan, Tulika Saha, Yushan Pan, Haiyang Zhang, Chengtao Ji}
\renewcommand{\shortauthors}{Yang and Li, et al.}

\begin{abstract}
Graph clustering is an essential aspect of network analysis that involves grouping nodes into separate clusters. Recent developments in deep learning have resulted in graph clustering, which has proven effective in many applications. Nonetheless, these methods often encounter difficulties when dealing with real-world graphs, particularly in the presence of noisy edges. Additionally, many denoising graph clustering methods tend to suffer from lower performance, training instability, and challenges in scaling to large datasets compared to non-denoised models. To tackle these issues, we introduce a new framework called the Robust Deep Graph Clustering Framework via Dual Soft Assignment (RDSA). RDSA consists of three key components: (i) a node embedding module that effectively integrates the graph's topological features and node attributes; (ii) a structure-based soft assignment module that improves graph modularity by utilizing an affinity matrix for node assignments; and (iii) a node-based soft assignment module that identifies community landmarks and refines node assignments to enhance the model's robustness. We assess RDSA on various real-world datasets, demonstrating its superior performance relative to existing state-of-the-art methods. Our findings indicate that RDSA provides robust clustering across different graph types, excelling in clustering effectiveness and robustness, including adaptability to noise, stability, and scalability. The code is available at \url{https://github.com/EsiksonX/RDSA}.

\end{abstract}

\ccsdesc[500]{Information systems~Clustering}
\ccsdesc[500]{Mathematics of computing~Graph algorithms}
\ccsdesc[300]{Computing methodologies~Neural networks}

\keywords{Deep Graph Clustering, Graph Neural Network, Robust Learning, Soft Assignment}

\maketitle

\renewcommand{\thefootnote}{\fnsymbol{footnote}}
\footnotetext[1]{Both authors contributed equally to this research.}
\footnotetext[2]{Corresponding author.}

\section{INTRODUCTION}
Graph clustering is crucial in network analysis, spanning physics, bioinformatics \cite{haoyu}, and social sciences \cite{fanshen}. Real-world graphs naturally exhibit clusters, representing sets of nodes with similar characteristics. These clusters hold significant implications, delineating social groups in friendship networks \cite{bedi2016community}, functional modules within protein-interaction networks \cite{chen2006detecting}, and thematic groups of academic papers within citation networks \cite{sen2008collective}. Recently, breakthroughs in deep learning have led to a paradigm shift in artificial intelligence and machine learning, achieving great success in many important tasks, including graph clustering. The main techniques used in deep graph clustering are MLP-based graph clustering \cite{gao2019progan,jia2019communitygan,cui2020adaptive,liu2023simple,pal1992multilayer}, GNN-based graph clustering \cite{wang2017mgae,park2019symmetric,fan2020one2multi,cheng2021multi,xia2021self,wang2021self}, and hybrid-based graph clustering \cite{bo2020structural,peng2021attention,tu2021deep,hassani2020contrastive,zhao2021graph,liu2022towards,liu2022deep,yang2023cluster,yang2023convert,yang2022contrastive}. 

The complexities inherent in real-world graph structures present notable challenges to the effectiveness of current deep graph clustering methodologies. Real-world datasets often struggle with noisy edges, which can severely impact the performance of deep graph clustering algorithms. Because of this, most research mainly focuses on denoising, a part of robust graph clustering. For instance, RSGC \cite{RSGC} primarily endeavors to preserve the inherent structure of graphs to combat noise and outliers, MetaGC \cite{MetaGC} addresses the issue of noisy graphs through meta-learning techniques, while ROGC \cite{ROGC} tackles noise elimination by acquiring an optimal graph structure within a denoising representation framework. However, these methods often face issues such as significantly lower performance compared to non-denoised models, instability during training, and limited scalability to large datasets. As a result, they tend to achieve only denoising rather than true robustness.

In graph clustering, soft assignment means nodes are given probabilities of belonging to different clusters rather than being assigned to just one. This probabilistic approach provides a more nuanced view of data relationships, capturing the ambiguity and overlaps in many datasets. Because of this flexibility, we aim to use soft assignment to improve the denoising ability of graph clustering methods. Traditional soft assignment methodologies in graph clustering, exemplified by DGCluster \cite{bhowmick2024dgcluster} and DMoN \cite{DMoN}, typically rely on modularity. Modularity, however, primarily focuses on the structural aspects of node embedding, often overlooking important information related to node information. Hence, we propose a novel method of soft assignment grounded in nodes embedding, complementing the existing structural-based soft assignment techniques. To elaborate, our approach involves identifying pivotal nodes within distinct modules, considered landmarks due to their symbolic significance. Leveraging these landmarks, we devise a mechanism for node-based soft assignment for each node, based on the node's embedding characteristics and proximity to the identified landmarks. Through experiments, we found that this new node-based soft assignment method further enhances the results of our graph clustering approach and improves its resistance to noise.

To address the limitations discussed above, we introduce a novel model named \textbf{R}obust deep graph clustering framework via \textbf{D}ual \textbf{S}oft \textbf{A}ssignment (RDSA). Using an autoencoder, we construct graph embeddings based on structural and attribute information. To enhance the robustness of our graph clustering process, we employ two distinct types of soft assignments. The first type is based on graph structure, where we initiate a preliminary soft assignment of graph nodes by calculating a modularity score using an affinity matrix between nodes. This is then optimized through modularity maximization to achieve a coarse partitioning of nodes based on modularity. The second type of soft assignment is based on node information; we select representative nodes from each module to serve as landmarks for the respective modules. Subsequently, we compute the similarity between each node and the landmarks of each module to determine a landmark-based soft assignment of nodes. Finally, we parameterize the soft assignment matrix using a Student's t-distribution to minimize the distance between nodes and their respective cluster landmarks. Our experiments demonstrate that node-based soft assignments enhance the robustness of graph clustering in terms of denoising, training stability, and scalability. We summarise the key contributions of this paper as follows:
\begin{itemize}
    \item We present a deep graph clustering method, RDSA, which leverages dual soft assignments to enhance robustness regarding effectiveness, handling noisy graphs, training stability, and scalability.
    \item Subsequently, we use modularity maximization and the Student's t-distribution, which minimizes the distance between nodes and their respective cluster landmarks, to determine each node's final module and cluster assignment from the soft assignment results.
    \item We conduct extensive experiments on six public datasets (Cora, Citeseer, PubMed, Amazon Photo, Amazon PC, ogbn-arxiv) to demonstrate the effectiveness of our proposed approach. We then apply three noise levels to real-world graphs to assess the method's denoising capability. These experiments highlight the advantages of RDSA over the state-of-the-art models.
\end{itemize} 

\section{RELATED WORK}

\subsection{Deep Graph Clustering}
Graph clustering is aimed at partitioning the nodes in a graph into several distinct clusters, which is a fundamental yet challenging task. In recent years, considerable success has been achieved with deep graph clustering techniques, owing to the robust representational capacities of deep learning. Based on the network architecture, we can classify the deep graph clustering algorithms into three categories: MLP-based graph clustering \cite{gao2019progan,jia2019communitygan,cui2020adaptive,liu2023simple,pal1992multilayer}, GNN-based graph clustering \cite{wang2017mgae,park2019symmetric,fan2020one2multi,cheng2021multi,xia2021self,wang2021self}, and hybrid graph clustering \cite{bo2020structural,peng2021attention,tu2021deep,hassani2020contrastive,zhao2021graph,liu2022towards,liu2022deep,yang2023cluster,yang2023convert,yang2022contrastive}.

\textbf{MLP-based Graph Clustering} utilize multi-layer perceptrons (MLP) \cite{pal1992multilayer} to precisely extract information features from graphs. For instance, GraphEncoder \cite{tian2014gae} and DNGR \cite{cao2016deep} employ autoencoders to encode graph structures. Subsequently, in ProGAN \cite{gao2019progan} and CommunityGAN \cite{jia2019communitygan}, the authors implemented MLPs to design generative adversarial networks. Moreover, based on MLP, AGE \cite{cui2020adaptive} and SCGC \cite{liu2023simple} devised adaptive encoders and parameter non-sharing encoders to embed smooth node features into latent spaces. While the effectiveness of these methods has been proven, MLPs struggle to capture the non-Euclidean structural information in graphs. Therefore, in recent years, GNN-based approaches have been increasingly proposed.

\textbf{GNN-based Graph clustering} employ graph encoders such as graph convolutional networks, graph attention networks, and graph auto-encoders to model non-Euclidean graph data effectively. These methods have demonstrated significant success due to their robust representation of graph structures. For example, MGAE \cite{wang2017mgae} was introduced to simultaneously capture node attributes and graph topology using a specially designed graph auto-encoder. A unique symmetric graph auto-encoder, GALA \cite{park2019symmetric}, was developed. GNNs have also been extended to handle heterogeneous graphs, as seen in O2MAC \cite{fan2020one2multi}, MAGCN \cite{cheng2021multi}, and SGCMC \cite{xia2021self}. However, GNNs face challenges due to the intertwined processes of transformation and propagation, leading to high computational costs. To address this, SCGC \cite{liu2023simple} was developed, enhancing the efficiency and scalability of deep graph clustering by separating these processes.

\textbf{Hybrid Graph Clustering}, which integrates the advantages of MLP-based and GNN-based approaches, is also the category to which our method belongs. Notably, research such as \cite{bo2020structural} showcases moving embeddings from an auto-encoder to a GCN layer through a dedicated delivery operator. Additionally, the combination of node attribute and graph topology features has been effectively demonstrated by AGCN \cite{peng2021attention} and DFCN \cite{tu2021deep}, leveraging the synergy of auto-encoders and GCNs. Moreover, several contrastive deep graph clustering techniques, for instance \cite{hassani2020contrastive}, \cite{zhao2021graph}, \cite{liu2022towards}, \cite{liu2022deep}, \cite{yang2023cluster}, \cite{yang2023convert}, \cite{yang2022contrastive}, and \cite{gong2022attributed}, have developed a hybrid architecture that amalgamates MLPs and GNNs functionalities.

\subsection{Modularity Maximaization}

Modularity, a metric that compares the density of links within communities to those between them, is foundational for identifying closely connected groups within extensive networks. Researchers have developed various methods to optimize modularity across different scenarios. In graph clustering, "modularity maximization" is crucial for improving community detection methods.

Yang et al. \cite{yang2016modularity} introduced a method using nonlinear reconstruction through graph autoencoders, enhancing modularity by imposing constraints on node pairs \cite{brandes2007modularity}. Wu et al. \cite{wu2020deep} further developed this by leveraging the adjacency matrix and identifying opinion leaders to generate a spatial proximity matrix, with spatial eigenvectors used to optimize modularity \cite{salha2022modularity}.

Mandaglio, Amelio, and Tagarelli \cite{mandaglio2018consensus} integrated the modularity metric into their community detection and graph clustering algorithms \cite{shiokawa2013fast}. Choong, Liu, and Murata \cite{choong2018learning}, alongside Bhatia and Rani \cite{bhatia2018dfuzzy}, explored community structures without predefined boundaries. Choong et al. used a generative model based on a variational autoencoder for community detection \cite{hu2018deep}, while Bhatia and Rani focused on fine-tuning potential community counts based on modularity \cite{newman2006modularity}.

Sun et al. \cite{sun2021graph} utilized graph neural networks to optimize modularity and attribute similarity, representing a significant advancement in community detection methods \cite{watanabe2018modular}. The latest approach, DMoN by Tsitsulin et al. \cite{DMoN}, presents an architecture encoding cluster assignments with an objective function based on modularity, advancing modularity maximization in graph clustering \cite{filan2021clusterability}. These diverse approaches highlight modularity's ongoing development and application as a crucial metric in analyzing complex network structures.

\section{PRELIMINARIES}
Consider an undirected graph $\mathcal{G} = (\boldsymbol{V}, \boldsymbol{E}, \boldsymbol{X})$, where $\boldsymbol{V}$ denotes nodes, $\boldsymbol{E}$ represents edges, and $\boldsymbol{X}$ contains node attributes. The adjacency matrix $\boldsymbol{A}$ has elements $\boldsymbol{a}_{ij} = 1$ if there exists an edge between nodes $i$ and $j$, otherwise $0$. The degree matrix $\boldsymbol{D}$ is diagonal, with $\boldsymbol{d}_{i}$ as the sum of connections for node $i$. $\boldsymbol{x}_i$ denotes the attributes of node $i$, and $\boldsymbol{\hat{X}}$ is its reconstructed version. $\boldsymbol{H}$ contains embedded nodes, with $\boldsymbol{h}_i$ as the embedding of node $i$. $\boldsymbol{C}$ is the affinity matrix, where $\boldsymbol{c}_{ij}$ is the probability of node $i$ associated with cluster $j$. $\boldsymbol{U}$ represents community landmarks, with $\boldsymbol{u}_i$ indicating the $i$-th landmark. $\boldsymbol{W}$ is the soft assignment matrix for community landmarks, where $\boldsymbol{w}_{ij}$ signifies the probability of node $i$ belonging to community landmark $j$. $\boldsymbol{\tilde{W}}$ is the sharpen version of $\boldsymbol{W}$.

\section{PROPOSED METHOD}

\begin{figure*}[ht]
  \centering
  \includegraphics[width=\linewidth]{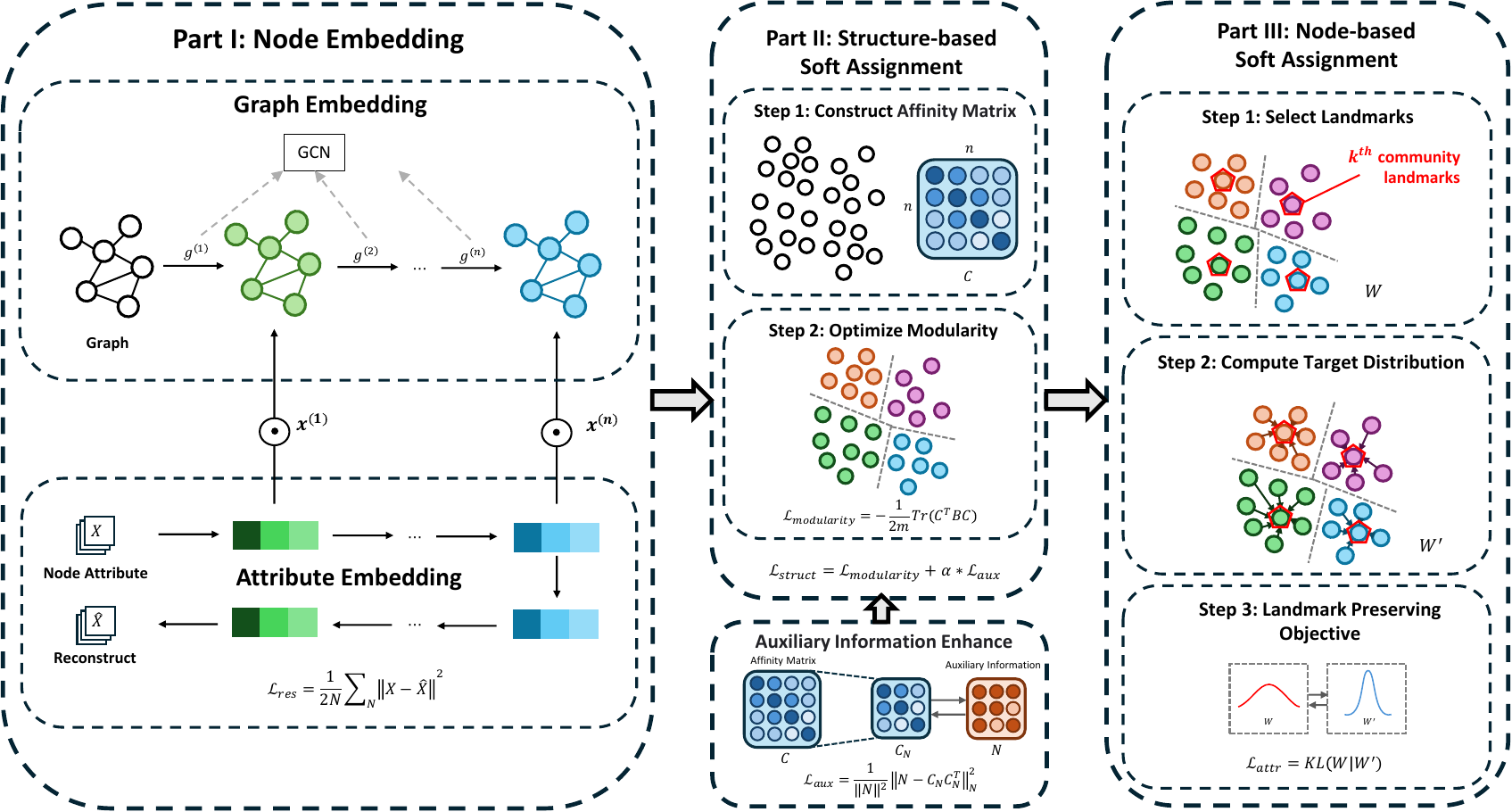}
  \caption[Short description of the image]{The overall framework of RDSA.}
  \label{fig:framework}
\end{figure*}

In this section, we introduce our proposed model, RDSA, designed to address the limitations and improve the robustness of existing deep-graph clustering methods. As illustrated in Figure \ref{fig:framework}, RDSA comprises three key components: the node embedding module, the graph structural-based soft assignment module, and the node attribute-based soft assignment module. The node embedding module captures the graph's topological information and node attributes, embedding them effectively. The structural-based soft assignment module optimizes community partitioning by maximizing modularity, while the attribute-based soft assignment module integrates critical node attribute information, identifying key community landmarks and refining node allocations. This comprehensive approach ensures robust clustering by aligning similar nodes with their community landmarks, enhancing the model's overall effectiveness and robustness.

\subsection{Node Embedding Module}
The node embedding module in deep graph clustering aims to map the input graph data to vector representations. In our proposed framework, we employ both multilayer perceptron (MLP) and graph neural network (GNN) techniques to embed nodes. We use an autoencoder (AE) to encode node attributes and a graph convolutional network (GCN) to capture the graph's topological information into embedded representations. The AE effectively embeds node attributes, while the GCN captures the graph's structural information. The output of the node embedding module for node \(u\) at the \(l\)-th layer is formulated as follows:
\begin{equation}
  \begin{aligned}
    \boldsymbol{h}^{(l)}_u =  &\sigma * \phi_{AE}(\boldsymbol{x}^{(l-1)}_u) + (1 - \sigma) * \\
    &\phi_{GCN}\left(\boldsymbol{h}^{(l-1)}_u, \left\{\boldsymbol{h}^{(l-1)}_v \mid v \in \mathcal{N}_u\right\}\right),
  \end{aligned}
\end{equation}
where \(\boldsymbol{x}^{(l-1)}_u\) is the output of the encoder at the \((l-1)\)-th layer, \(\boldsymbol{h}^{(l-1)}_u\) is the output of the GCN at the \((l-1)\)-th layer, \(\phi\) denotes the activation functions (specifically, ReLU \cite{glorot2011RELU}, LeakyReLU \cite{maas2013leakyrelu}, SeLU \cite{klaumbauer2017selu}), \(\mathcal{N}_u\) is the set of neighbors of node \(u\), and \(\sigma\) is a hyperparameter controlling the weight of the node attributes information and graph topology information. The output of the node embedding module is the embedded node representation \(\boldsymbol{H} = \boldsymbol{h}^{(L)}\). We use the mean squared error (MSE) to minimize the reconstruction difference:
\begin{equation}
  \begin{aligned}
    \mathcal{L}_{res} = \frac{1}{2N}||\boldsymbol{X} - \hat{\boldsymbol{X}}||^2,
  \end{aligned}
\end{equation}
where \(\boldsymbol{X}\) and \(\hat{\boldsymbol{X}}\) represent the input and reconstructed node attributes, $N$ is the number of nodes. This MLP-GNN mixture architecture enables us to obtain embedded node representations that encompass both node attributes and graph topology information. In this work, we consider a basic autoencoder and two representative GCN architectures: GraphSAGE \cite{Hamilton2017GraphSAGE} and GCN \cite{kipf2016semi}. Other MLP and GNN architectures can also be incorporated within our proposed framework.

\subsection{Structure-based Soft Assignment}
Structure-based soft assignment primarily focuses on leveraging the structural information of graphs to partition them into distinct communities. To quantify and evaluate structural information, we use modularity, a measure introduced by Newman \cite{newman2006modularity}, which evaluates graph partitions based on the density of edges intracommunity versus intercommunity. A higher modularity value indicates a better partitioning of the graph structure \cite{newman2006finding}. In this module, our objective is to maximize the modularity of the structure-based soft assignment to achieve an optimal partitioning of the graph. We employ the reformulated matrix form of modularity \cite{DMoN} to calculate the modularity of these assignments. Given the modularity matrix $\boldsymbol{B}$, defined as $\boldsymbol{B} = \boldsymbol{A} - \frac{\boldsymbol{D} \boldsymbol{D}^\top}{2m}$, where $\boldsymbol{A}$ is the adjacency matrix of the graph, $\boldsymbol{D}$ is the degree matrix, and $m$ is the number of edges in the graph, the modularity can be formally defined as follows:
\begin{equation}
  \begin{aligned}
    \boldsymbol{\mathcal{Q}} = \frac{1}{2m} \text{Tr}(\boldsymbol{C}^\top \boldsymbol{B} \boldsymbol{C}), \text{ s.t. } \boldsymbol{C} = \frac{\text{tanh}(\boldsymbol{H})^2}{\| \text{tanh}(\boldsymbol{H})^2 \|_1}.
  \end{aligned}
\end{equation}
Here, $\boldsymbol{C}$ represents the assignment matrix, where $C(i,j)$ indicates the probability of nodes $i$ and $j$ belonging to the same cluster. $H$ denotes the embedded nodes and $| \cdot |_1$ denotes the $L_1$ norm. To achieve our objective of maximizing modularity, we can minimizing the negative of modularity:
\begin{equation}
  \begin{aligned}
    \mathcal{L}_{struct} = -\boldsymbol{\mathcal{Q}}.
  \end{aligned}
\end{equation}
The random initial state in clustering tasks can lead to an unstable training process, causing convergence to a local minimum. To mitigate this problem, we use an auxiliary information loss, inspired by DGCluster \cite{bhowmick2024dgcluster}, to guide the model toward global optimal convergence and improve stability. Let $\boldsymbol{M} \in \boldsymbol{V}$ represent a subset of nodes, and $\boldsymbol{N}$ denote the pairwise membership matrix of $\boldsymbol{M}$, where $\boldsymbol{N}(i,j) = 1$ if nodes $i$ and $j$ belong to the same cluster, and $\boldsymbol{N}(i,j) = 0$ otherwise. The auxiliary information loss is defined as:
\begin{equation}
  \begin{aligned}
    \mathcal{L}_{aux} = \frac{1}{\|\boldsymbol{N}\|^2}\|\boldsymbol{N} - \boldsymbol{C}_N \boldsymbol{C}^\top_N\|_F^2.
  \end{aligned}
\end{equation}
However, some datasets lack additional information. In these cases, clusters tend to show higher structural density in their central regions. For datasets without this information, these central regions can act as substitutes for the missing data. Let $\alpha$ represent the weight of auxiliary information loss. Often, the reliability of auxiliary information is uncertain. When we are unsure about the confidence of this information, we typically assign a lower weight to mitigate the effects of this uncertainty \cite{bhowmick2024dgcluster}. The total loss function for the structure-based soft assignment can be formulated as:
\begin{equation}
  \begin{aligned}
    \mathcal{L}_{struct} = \mathcal{L}_{mod} + \alpha \mathcal{L}_{aux}.
  \end{aligned}
\end{equation}

\subsection{Node-based Soft Assignment}
In the proposed framework, we introduce a structure-based soft assignment method focused on modularity maximization. However, this primarily considers only the graph structure. Inspired by SLIM \cite{zhu2022landmarking}, which identifies representative nodes (landmarks) for each community to characterize the entire graph, The landmarks are dynamically updated during training, which is different from traditional fixed centroids. Community landmarks effectively capture the distribution of communities and improve the ability to distinguish between different communities, which means that landmarks can provide strong discriminative power in unsupervised clustering. Therefore, landmarks can be introduced to improve the robustness of graph clustering.

The node-based soft assignment process begins with the selection of landmarks. Specifically, the $k$ most representative nodes within each community are chosen based on connectivity, assessed via modularity. As landmarks are central to representing community structures, they are chosen to be well-connected within their respective communities to provide a stable representation. By selecting nodes with the highest modularity scores, each landmark effectively captures the local community structure and nuances, performing more robustly than centroids and being less affected by outliers. This landmark selection process can be formalized as follows:
\begin{equation}
  \begin{aligned}
    U=\{u_1, \cdots, u_k\}=\underset{U \subseteq H, |U|=k}{\arg \max } \sum_{i} \sum_{j} \boldsymbol{B}(i, j),
  \end{aligned}
\end{equation}
where $U$ is the set of landmarks, $u_k$ is the $k$-th community landmark, $\boldsymbol{H}$ denotes the embedded nodes, and $\boldsymbol{B}$ denotes the modularity matrix. 

Following the identification of landmarks, we need to assign embedded nodes softly to these landmarks. To achieve this, we utilize the t-student distribution to calculate the assignment probability between embedded nodes and landmarks:
\begin{equation}
  \label{eq:soft_assignment2}
  \begin{aligned}
    \boldsymbol{W}(i, k)=\frac{\left(1+\frac{\left\|\boldsymbol{H}(i,:)-u_k\right\|^2}{\upsilon} \right)^{-\frac{\upsilon+1}{2}}}{\sum_{k^{\prime}}\left(1+\frac{\left\|\boldsymbol{H}(i,:)-u_{k^{\prime}}\right\|^2}{\upsilon} \right)^{-\frac{\upsilon+1}{2}}},
  \end{aligned}
\end{equation}
where $\boldsymbol{W}(i, k)$ represents the probability of node $i$ being assigned to landmark $k$, and $\upsilon$ is a hyperparameter controlling the tail thickness of the t-distribution. Although other distributions such as Gaussian and Gamma have been widely used in clustering tasks \cite{young2019finite,yang2019deep}, we choose the t-distribution due to its heavy tails, which enhance the model's robustness to outliers and noise. A lower $\upsilon$ results in thicker tails, meaning that embedded nodes are more tightly distributed within clusters. However, an excessively small $\upsilon$ may increase the sensitivity to pairwise distances, making the assignment more similar to hard assignment. To balance robustness and sensitivity, we set $\upsilon$ to 1, ensuring an optimal distribution of nodes within clusters \cite{liu2024robust}.

To enhance the connections between nodes and landmarks within communities, we introduce an objective to sharpen the soft assignment matrix:
\begin{equation}
  \begin{aligned}
    \mathcal{L}_{attr}=\text{KL}(\boldsymbol{W} || \tilde{\boldsymbol{W}}),
  \end{aligned}
\end{equation}
where $\tilde{\boldsymbol{W}}(i, k)$ represents the sharpened probability of node $i$ being assigned to landmark $k$:
\begin{equation}
  \begin{aligned}
    \tilde{\boldsymbol{W}}(i, k)=\frac{\boldsymbol{W}^{2}(i, k)/\sum_{n}\boldsymbol{W}(n, k)}{\sum_{k^{\prime}}[\boldsymbol{W}^{2}(i, k^{\prime})/\sum_{n}\boldsymbol{W}(n, k^{\prime})]}.
  \end{aligned}
\end{equation}
The sharpening process amplifies the assignment probabilities, making nodes with higher probabilities more likely to be assigned to only a few landmarks. This ensures that the distribution of embedded nodes is more centralized and sharper. Normalization guarantees that the sharpened matrix still adheres to the properties of a probability distribution. We then minimize the Kullback-Leibler (KL) divergence between the soft assignment matrix $\boldsymbol{W}$ and its sharpened version $\tilde{\boldsymbol{W}}$ to strengthen the node-landmark connections within communities.

\subsection{Training Loss}
In this section, we will introduce our joint training objective. The overall training objective of RDSA is the combination of the structural-based soft assignment objective $\mathcal{L}_{struct}$, the node-based soft assignment objective $\mathcal{L}_{attr}$, and the reconstruction loss $\mathcal{L}_{res}$, defined as:
\begin{equation}
  \begin{aligned}
    \mathcal{L} = \mathcal{L}_{res} + \mathcal{L}_{struct} + \mathcal{L}_{attr},
  \end{aligned}
\end{equation}
where $\mathcal{L}_{res}$ denote the Mean Squared Error (MSE) between the input node attributes and the reconstructed node attributes, $\mathcal{L}_{struct}$ denote the structural-based soft assignment objective to maximize the modularity of the partition of the graph structure, and $\mathcal{L}_{attr}$ denote the node-based soft assignment objective utilize the Kullback-Leibler (KL) divergence between the soft assignment matrix and the sharpened version to enhance the node-landmark connections within communities.

\section{COMPLEXITY ANALYSIS}
\textbf{\textit{Space Complexity}}: The space complexity of RDSA is given by $\mathcal{O}(m + Nd + b^2)$, where $m$ is the number of edges, $N$ is the number of nodes, $d$ is the dimension of the node attributes, and $b$ is the batch size. For all datasets, we store the sparse adjacency matrix $\boldsymbol{A}$, the node attributes $\boldsymbol{X}$, and the mini-batch modularity matrix $\boldsymbol{B}$, which require storage spaces of $\mathcal{O}(m)$, $\mathcal{O}(Nd)$, and $\mathcal{O}(b^2)$, respectively.
\textbf{\textit{Time Complexity}}: The time complexity of RDSA is $\mathcal{O}(rbd + rbd^2)$, where $r$ is the number of epochs. For each epoch, given a batch, the forward and backward computations for the MLP and GNN require $\mathcal{O}(bd)$ and $\mathcal{O}(bd^2)$, respectively.

\section{EXPERIMENTS}
This section presents the reality datasets used for our experiments, the experimental settings, including hardware and software settings, the evaluation metrics, the baseline models, and the implementation and hyperparameter settings; finally, we present the experimental results and analysis.

\subsection{\textbf{Datasets}}
We evaluate our proposed method on two medium-scale datasets (Cora and Citeseer), three large-scale datasets (Pubmed, Amazon Photo, and Amazon Computers), and two extra large scale dataset (ogbn arxiv, and ogbn products ). The statistics of these datasets are summarized in Table \ref{table:datasets}.
.

\begin{table}[!h]
  \centering
  \caption{Statistics of the datasets.}
  \label{table:datasets}
  \resizebox{\linewidth}{!}{
    \begin{tabular}{l|c|c|c|c}
      \toprule
      \textbf{Dataset} & \textbf{\# Nodes} & \textbf{\# Edges} & \textbf{\# Features} & \textbf{\# Clusters} \\
      \midrule
      \midrule
      Cora & 2,708 & 5,429 & 1,433 & 7 \\
      Citeseer & 3,327 & 4,732 & 3,703 & 6 \\
      Pubmed & 19,717 & 44,324 & 500 & 3 \\
      Amazon Photo & 7,650 & 119,081 & 745 & 8 \\
      Amazon Computers & 13,752 & 245,778 & 767 & 10 \\
      ogbn arxiv & 169,343 & 1,166,243 & 128 & 40 \\
      ogbn products & 2,449,029 & 61,465,973 & 100 & 47 \\
      \bottomrule
    \end{tabular}
  }
\end{table}

\subsubsection*{\textbf{Noise Generation}}
To simulate real-world scenarios, we introduce noise edges to the datasets. Specifically, we add $30\%$ (Level I), $60\%$ (Level II), and $90\%$ (Level III) noisy edges to the original datasets. The noise edges are randomly generated between nodes that do not belong to the same class. The noisy datasets are used to evaluate the noise adaptability of our proposed method in Sec. \ref{sec:adaptability}.

\subsection{Experimental Setup}
\subsubsection*{\textbf{Hardware and Software}}
We conduct our experiments on a HPC server with an AMD EPYC 9754, 60GB RAM, and an NVIDIA GeForce RTX 4090 D GPU with 24GB memory. The software environment includes Python 3.11, Pytorch 2.3, and CUDA 12.1.

\subsubsection*{\textbf{Metrics}}
We adopt four metrics to evaluate the performance of our proposed method: Accuracy (ACC), Normalized Mutual Information (NMI), Adjusted Rand Index (ARI), and macro F1 score. The average and standard deviation of these metrics are calculated over ten runs.

\subsubsection*{\textbf{Baseline Models}}
We compare our proposed method with several state-of-the-art deep graph clustering methods:
\begin{itemize}
  \item AGE \cite{cui2020adaptive} is an adaptive graph encoder that learns node representations by adaptively adjusting the weights of the encoder;
  \item SCGC \cite{liu2023simple} is a simple contrastive graph clustering method that separates the transformation and propagation processes to enhance efficiency;
  \item DCRN \cite{liu2022deep} is a deep clustering method that employs a deep clustering network to learn node representations;
  \item ARGA \cite{ARGA2020shiyu} is an attributed graph autoencoder that learns node representations by reconstructing node attributes;
  \item MAGI \cite{liu2024revisiting} is a contrastive learning-based method that utilizes modularity maximization as a pretext task to enhance node representation learning and improve graph clustering performance;
  \item SDCN \cite{bo2020structural} is a structural deep clustering network that learns node representations by combining structural and attribute information;
  \item SUBLIME \cite{liu2022towards} is a deep graph clustering method that learns node representations by combining structural and attribute information;
  \item AGC-DRR \cite{gong2022attributed} is an attributed graph clustering method that learns node representations by reconstructing node attributes;
  \item CONVERT \cite{yang2023convert} is a deep graph clustering method that learns node representations by combining structural and attribute information.
\end{itemize}
The hyperparameters of the baselines are set according to the original papers.

\begin{table*}[ht]
  \centering
  \caption{Performance comparison with state-of-the-art methods on six datasets, \besthighlight{red} indicates the best performance, \secondhighlight{blue} indicates the second-best performance, \thirdhighlight{green} indicates the third-best performance, and OOM denotes out of memory.}
  \label{tabel:performance}
  \resizebox{\linewidth}{!}{
      \begin{tabular}{c|c|>{\centering\arraybackslash}m{1.2cm}>{\centering\arraybackslash}m{1.2cm}|>{\centering\arraybackslash}m{1.2cm}>{\centering\arraybackslash}m{1.2cm}>{\centering\arraybackslash}m{1.2cm}|>{\centering\arraybackslash}m{1.2cm}>{\centering\arraybackslash}m{1.4cm}>{\centering\arraybackslash}m{1.5cm}>{\centering\arraybackslash}m{1.4cm}|>{\centering\arraybackslash}m{1.2cm}}
          \hline
          \multirow{3}*{\textbf{Dataset}} & \multirow{3}*{\textbf{Metrics}} & \multicolumn{2}{c|}{\textbf{MLP-based Methods}} & \multicolumn{3}{c|}{\textbf{GNN-based Methods}} & \multicolumn{4}{c|}{\textbf{Hybrid Methods}} & \textbf{Ours} \\
          \cline{3-12}
          \multicolumn{1}{c|}{} & \multicolumn{1}{c|}{} &
          \textbf{AGE} & \textbf{SCGC} &
          \textbf{DCRN} & \textbf{ARGA} & \textbf{MAGI} &
          \textbf{SDCN} & \textbf{SUBLIME} & \textbf{AGC-DRR} & \textbf{CONVERT} & \multirow{2}*{\textbf{RDSA}}\\
          \multicolumn{1}{c|}{} & \multicolumn{1}{c|}{} &
          \cite{cui2020adaptive} & \cite{liu2023simple} &
          \cite{liu2022deep} & \cite{ARGA2020shiyu} & \cite{liu2024revisiting} &
          \cite{bo2020structural} & \cite{liu2022towards} & \cite{gong2022attributed} & \cite{yang2023convert} & \\
          \hline
          \multirow{4}*{\textbf{Cora}} & \textbf{ACC} & 
          $73.50$ & $73.88$ & $59.34$ & $71.04$ & \secondhighlight{$76.00$} & $35.60$ & $71.14$ & $40.62$ & \thirdhighlight{$74.07$} & \besthighlight{$81.40$} \\
          \multicolumn{1}{c|}{} & \textbf{NMI} & 
          \thirdhighlight{$57.58$} & $56.10$ & $44.53$ & $51.06$ & \secondhighlight{$59.70$} & $14.28$ & $53.88$ & $18.74$ & $55.57$ & \besthighlight{$69.52$} \\
          \multicolumn{1}{c|}{} & \textbf{ARI} & 
          $50.10$ & \thirdhighlight{$51.79$} & $33.34$ & $47.71$ & \secondhighlight{$57.30$} & $07.78$ & $50.15$ & $14.80$ & $50.58$ & \besthighlight{$70.97$} \\
          \multicolumn{1}{c|}{} & \textbf{F1} & 
          $69.28$ & $70.81$ & $50.00$ & $69.27$ & \secondhighlight{$73.90$} & $24.37$ & $63.11$ & $31.23$ & \thirdhighlight{$72.92$} & \besthighlight{$77.42$} \\
          \hline
          \multirow{4}*{\textbf{Citeseer}} & \textbf{ACC} & 
          $68.73$ & \secondhighlight{$71.02$} & $57.74$ & $61.07$ & \thirdhighlight{$70.60$} & $65.96$ & $64.14$ & $68.32$ & $68.43$ & \besthighlight{$72.14$} \\
          \multicolumn{1}{c|}{} & \textbf{NMI} & 
          $44.93$ & \secondhighlight{$45.25$} & $37.01$ & $34.40$ & \thirdhighlight{$45.20$} & $38.71$ & $39.08$ & $40.28$ & $41.62$ & \besthighlight{$51.95$} \\
          \multicolumn{1}{c|}{} & \textbf{ARI} & 
          $45.31$ & \thirdhighlight{$46.29$} & $33.38$ & $34.32$ & \secondhighlight{$46.80$} & $40.17$ & $39.27$ & $45.34$ & $42.77$ & \besthighlight{$53.24$} \\
          \multicolumn{1}{c|}{} & \textbf{F1} & 
          \thirdhighlight{$64.45$} & \secondhighlight{$64.80$} & $46.21$ & $58.23$ & \secondhighlight{$64.80$} & $63.62$ & $61.00$ & \besthighlight{$64.82$} & $62.39$ & $63.97$ \\
          \hline
          \multirow{4}*{\textbf{PubMed}} & \textbf{ACC} &
          $45.96$ & $67.51$ & \secondhighlight{$69.87$} & $65.34$ & $63.78$ & $50.36$ & $59.91$ & $40.00$ & \thirdhighlight{$69.60$} & \besthighlight{$85.45$} \\
          \multicolumn{1}{c|}{} & \textbf{NMI} &
          $12.29$ & \thirdhighlight{$30.66$} & \secondhighlight{$32.20$} & $25.04$ & $25.10$ & $15.53$ & $22.38$ & $00.23$ & $29.97$ & \besthighlight{$52.39$} \\
          \multicolumn{1}{c|}{} & \textbf{ARI} &
          $5.57$ & $29.68$ & \secondhighlight{$31.41$} & $24.56$ & $23.00$ & $11.74$ & $19.47$ & $-00.32$ & \thirdhighlight{$30.09$} & \besthighlight{$61.73$} \\
          \multicolumn{1}{c|}{} & \textbf{F1} &
          $41.29$ & $67.30$ & \secondhighlight{$68.94$} & $65.51$ & $63.37$ & $42.58$ & $60.69$ & $24.88$ & \thirdhighlight{$68.20$} & \besthighlight{$84.27$} \\
          \hline
          \multirow{4}*{\makecell{\textbf{Amazon}\\\textbf{Photo}}} & \textbf{ACC} &
          $75.60$ & \secondhighlight{$77.48$} & $72.90$ & $69.28$ & \besthighlight{$79.00$} & $53.44$ & $27.22$ & $76.81$ & \thirdhighlight{$77.19$} & $74.63$ \\
          \multicolumn{1}{c|}{} & \textbf{NMI} &
          $64.87$ & \thirdhighlight{$67.67$} & $60.82$ & $58.36$ & \secondhighlight{$71.60$} & $44.85$ & $06.37$ & $66.54$ & $67.20$ & \besthighlight{$75.61$} \\
          \multicolumn{1}{c|}{} & \textbf{ARI} &
          $54.82$ & $58.48$ & $50.21$ & $44.18$ & \secondhighlight{$61.50$} & $31.21$ & $05.36$ & $60.15$ & \thirdhighlight{$60.79$} & \besthighlight{$63.77$} \\
          \multicolumn{1}{c|}{} & \textbf{F1} &
          \thirdhighlight{$72.85$} & $72.22$ & $67.61$ & $64.30$ & \secondhighlight{$72.90$} & $50.66$ & $15.97$ & $71.03$ & \besthighlight{$74.03$} & $66.10$ \\
          \hline
          \multirow{4}*{\makecell{\textbf{Amazon}\\\textbf{PC}}} & \textbf{ACC} &
          \secondhighlight{$69.09$} & $62.42$ & \thirdhighlight{$64.21$} & \multirow{4}*{OOM} & $62.00$ & $43.72$ & $30.78$ & $51.07$ & $55.25$ & \besthighlight{$70.52$} \\
          \multicolumn{1}{c|}{} & \textbf{NMI} &
          $48.53$ & \thirdhighlight{$51.60$} & $48.90$ & \multicolumn{1}{c}{} & \secondhighlight{$59.20$} & $37.04$ & $01.76$ & $45.83$ & $51.36$ & \besthighlight{$63.02$} \\
          \multicolumn{1}{c|}{} & \textbf{ARI} &
          \secondhighlight{$49.75$} & \thirdhighlight{$48.40$} & $34.73$ & \multicolumn{1}{c}{} & $46.20$ & $28.57$ & $06.53$ & $31.15$ & $35.91$ & \besthighlight{$61.06$} \\
          \multicolumn{1}{c|}{} & \textbf{F1} &
          $48.12$ & \thirdhighlight{$50.65$} & \secondhighlight{$51.85$} & \multicolumn{1}{c}{} & \besthighlight{$57.40$} & $26.42$ & $11.77$ & $36.50$ & $48.61$ & $40.89$ \\
          \hline
          \multirow{4}*{\textbf{ogbn-arxiv}} & \textbf{ACC} &
          \multirow{4}*{OOM} & \multirow{4}*{OOM} & \multirow{4}*{OOM} & \multirow{4}*{OOM} & \secondhighlight{$38.80$} & \multirow{4}*{OOM} & \multirow{4}*{OOM} & \multirow{4}*{OOM} & \multirow{4}*{OOM} & \besthighlight{$42.12$} \\
          \multicolumn{1}{c|}{} & \textbf{NMI} &
          \multicolumn{1}{c}{} & \multicolumn{1}{c|}{} & \multicolumn{1}{c}{} & \multicolumn{1}{c}{} & \besthighlight{$46.90$} & \multicolumn{1}{c}{} & \multicolumn{1}{c}{} & \multicolumn{1}{c}{} & \multicolumn{1}{c|}{} & \secondhighlight{$35.79$} \\
          \multicolumn{1}{c|}{} & \textbf{ARI} &
          \multicolumn{1}{c}{} & \multicolumn{1}{c|}{} & \multicolumn{1}{c}{} & \multicolumn{1}{c}{} & \secondhighlight{$31.00$} & \multicolumn{1}{c}{} & \multicolumn{1}{c}{} & \multicolumn{1}{c}{} & \multicolumn{1}{c|}{} & \besthighlight{$33.70$} \\
          \multicolumn{1}{c|}{} & \textbf{F1} &
          \multicolumn{1}{c}{} & \multicolumn{1}{c|}{} & \multicolumn{1}{c}{} & \multicolumn{1}{c}{} & \besthighlight{$26.60$} & \multicolumn{1}{c}{} & \multicolumn{1}{c}{} & \multicolumn{1}{c}{} & \multicolumn{1}{c|}{} & \secondhighlight{$14.72$} \\
          \hline
          \multirow{4}*{\makecell{\textbf{ogbn}\\\textbf{products}}} & \textbf{ACC} &
          \multirow{4}*{OOM} & \multirow{4}*{OOM} & \multirow{4}*{OOM} & \multirow{4}*{OOM} & \secondhighlight{$42.50$} & \multirow{4}*{OOM} & \multirow{4}*{OOM} & \multirow{4}*{OOM} & \multirow{4}*{OOM} & \besthighlight{$48.02$} \\
          \multicolumn{1}{c|}{} & \textbf{NMI} &
          \multicolumn{1}{c}{} & \multicolumn{1}{c|}{} & \multicolumn{1}{c}{} & \multicolumn{1}{c}{} & \besthighlight{$55.10$} & \multicolumn{1}{c}{} & \multicolumn{1}{c}{} & \multicolumn{1}{c}{} & \multicolumn{1}{c|}{} & \secondhighlight{$54.58$} \\
          \multicolumn{1}{c|}{} & \textbf{ARI} &
          \multicolumn{1}{c}{} & \multicolumn{1}{c|}{} & \multicolumn{1}{c}{} & \multicolumn{1}{c}{} & \secondhighlight{$21.50$} & \multicolumn{1}{c}{} & \multicolumn{1}{c}{} & \multicolumn{1}{c}{} & \multicolumn{1}{c|}{} & \besthighlight{$32.04$} \\
          \multicolumn{1}{c|}{} & \textbf{F1} &
          \multicolumn{1}{c}{} & \multicolumn{1}{c|}{} & \multicolumn{1}{c}{} & \multicolumn{1}{c}{} & \besthighlight{$27.60$} & \multicolumn{1}{c}{} & \multicolumn{1}{c}{} & \multicolumn{1}{c}{} & \multicolumn{1}{c|}{} & \secondhighlight{$22.72$} \\
          \hline
      \end{tabular}
  }
\end{table*}
\begin{figure*}[h]
\centering
\caption{Scatter plots showing the embedding results of six methods (SCGC, SDCN, SUBLIME, AGC-DRR, CONVERT, and RDSA) after dimensionality reduction using the t-SNE algorithm. The colors represent the ground truth labels.}
\label{fig:embedded_space}
\subfigure[SCGC]{
  \begin{minipage}[c][\height][c]{0.15\linewidth}
    \centering
    \rotatebox{90}{\scriptsize{\textbf{\quad \quad \quad Cora}}}
    \includegraphics[width=0.9\textwidth]{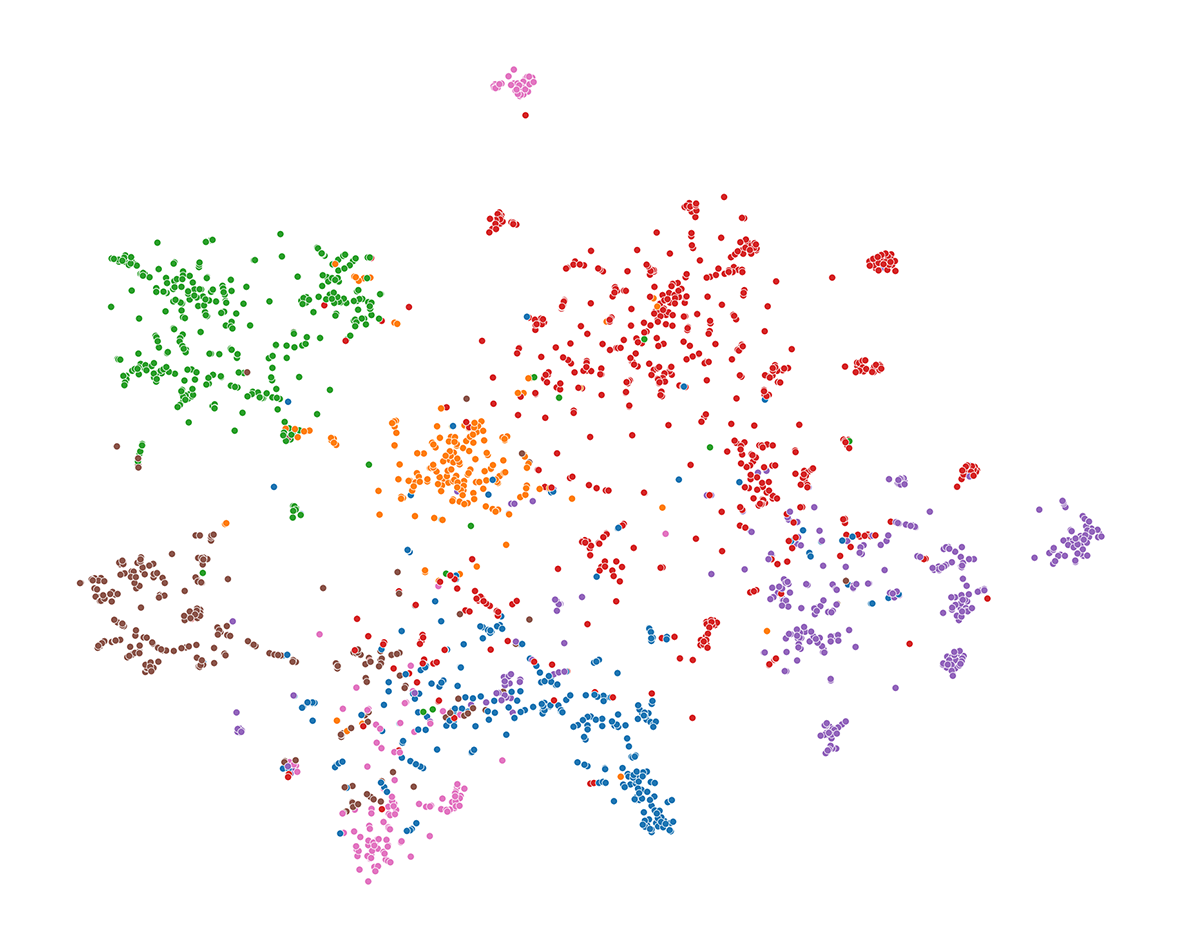} \\
    \rotatebox{90}{\scriptsize{\textbf{\quad \quad \; Citeseer}}}
    \includegraphics[width=0.9\textwidth]{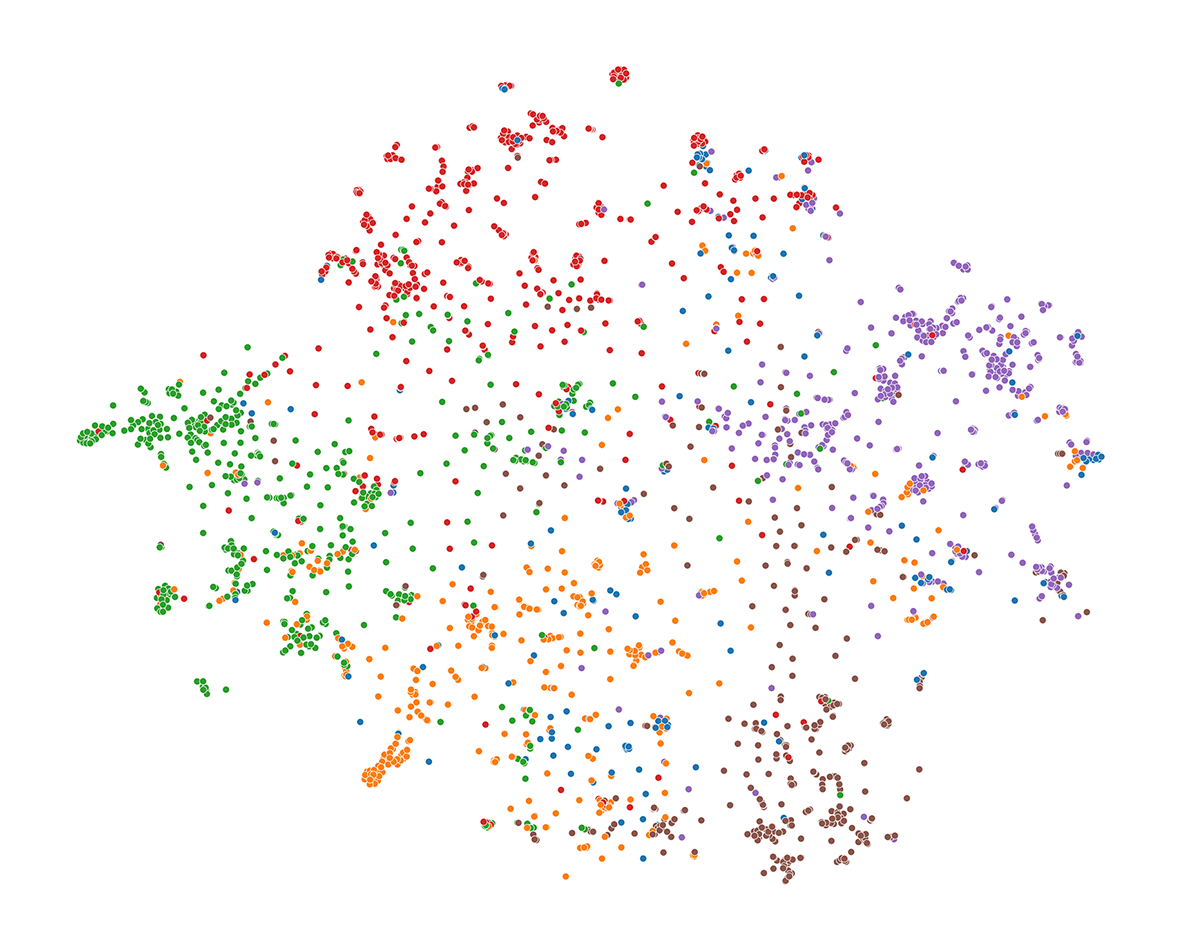} \\
    \rotatebox{90}{\scriptsize{\textbf{\quad \quad \ PubMed}}}
    \includegraphics[width=0.9\textwidth]{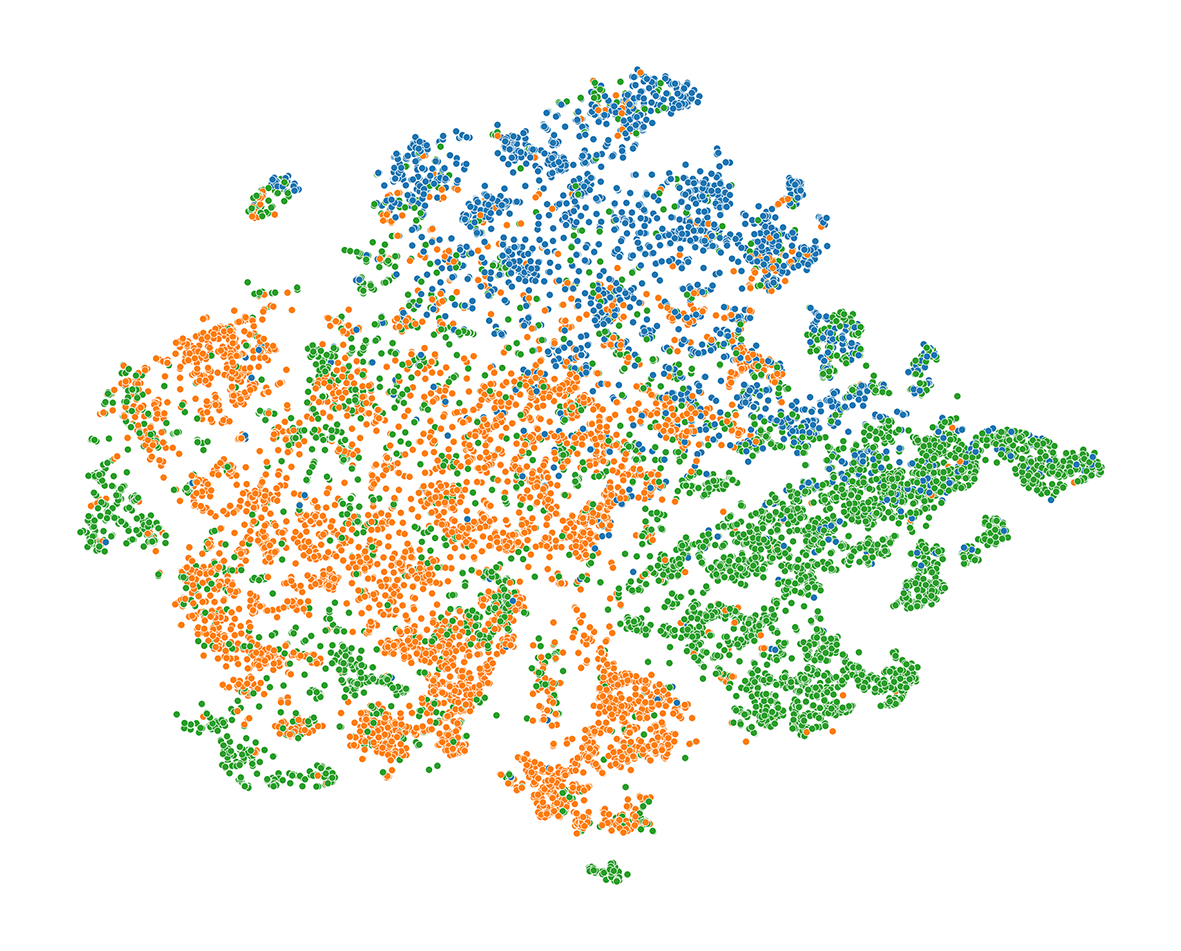} \\
    \rotatebox{90}{\scriptsize{\textbf{\quad \quad \quad Photo}}}
    \includegraphics[width=0.9\textwidth]{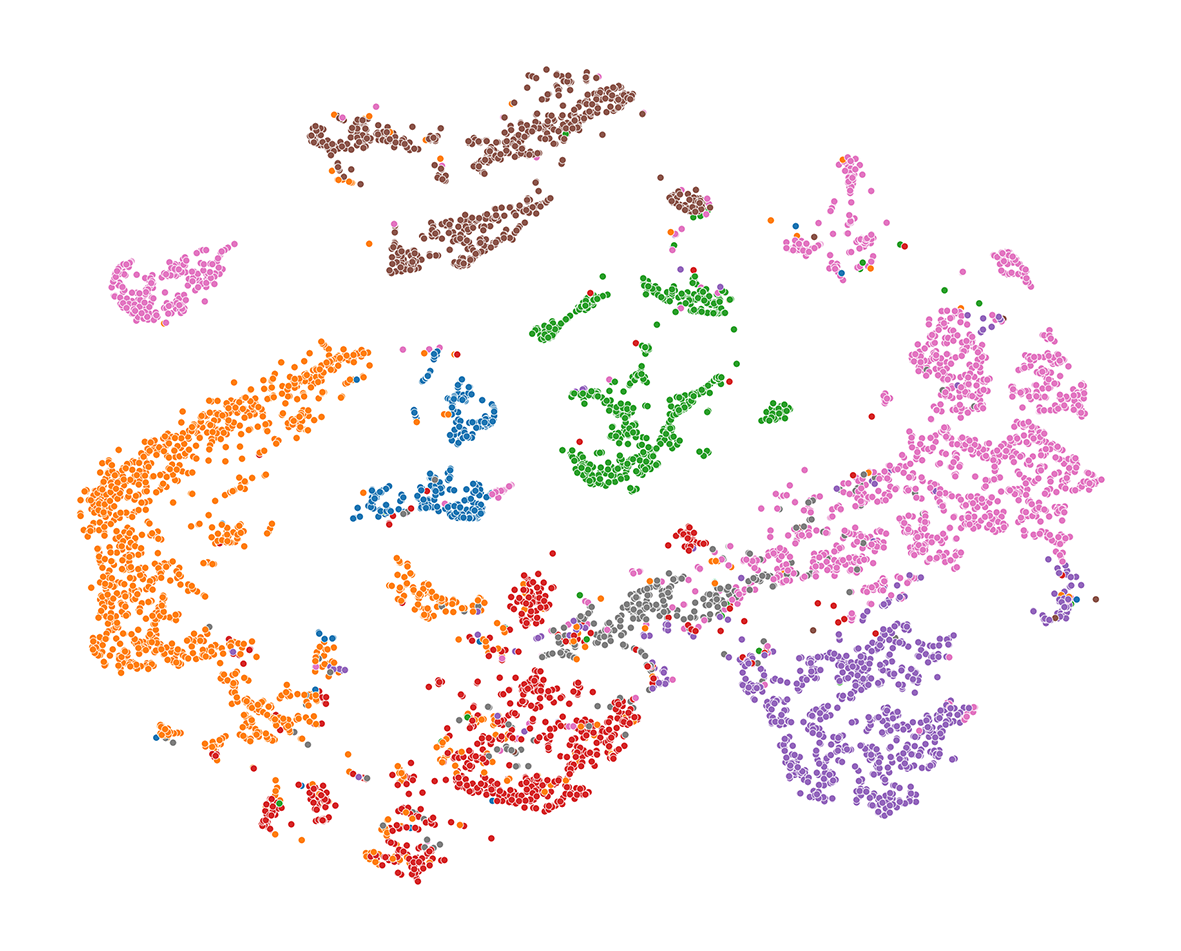} \\
    \rotatebox{90}{\scriptsize{\textbf{\quad \quad Computers}}}
    \includegraphics[width=0.9\textwidth]{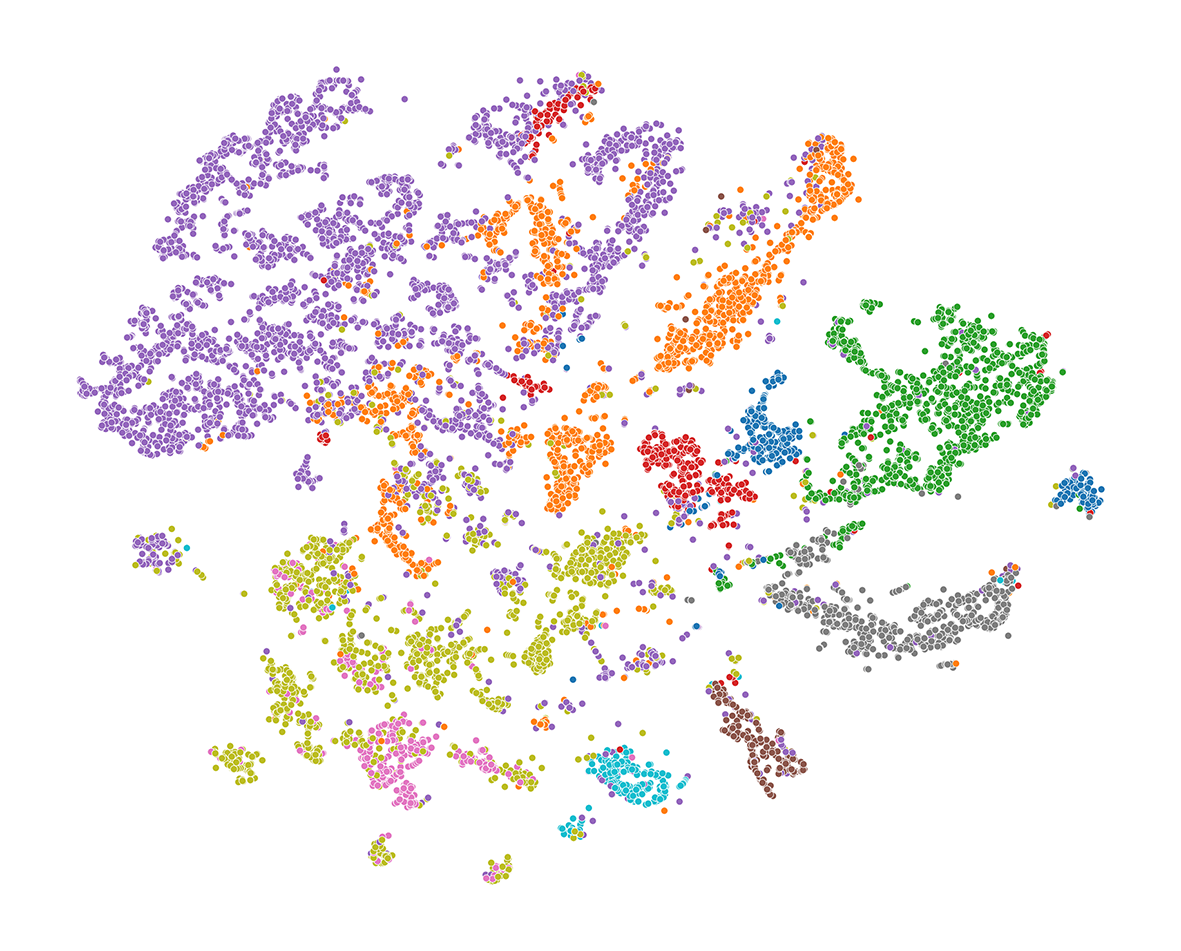}
  \end{minipage}
}
\subfigure[SDCN]{
  \begin{minipage}[c][\height][c]{0.15\linewidth}
    \centering
    \includegraphics[width=0.9\textwidth]{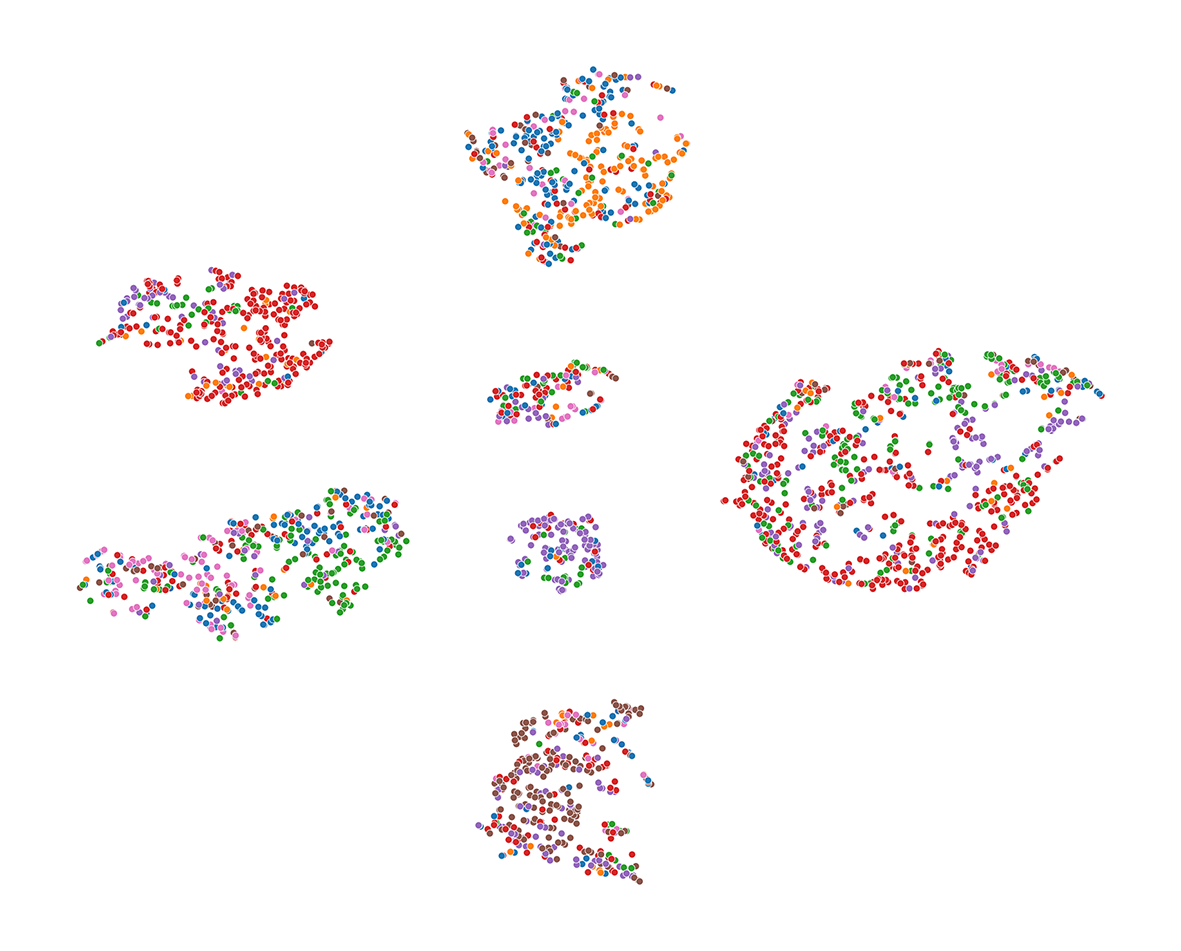} \\
    \includegraphics[width=0.9\textwidth]{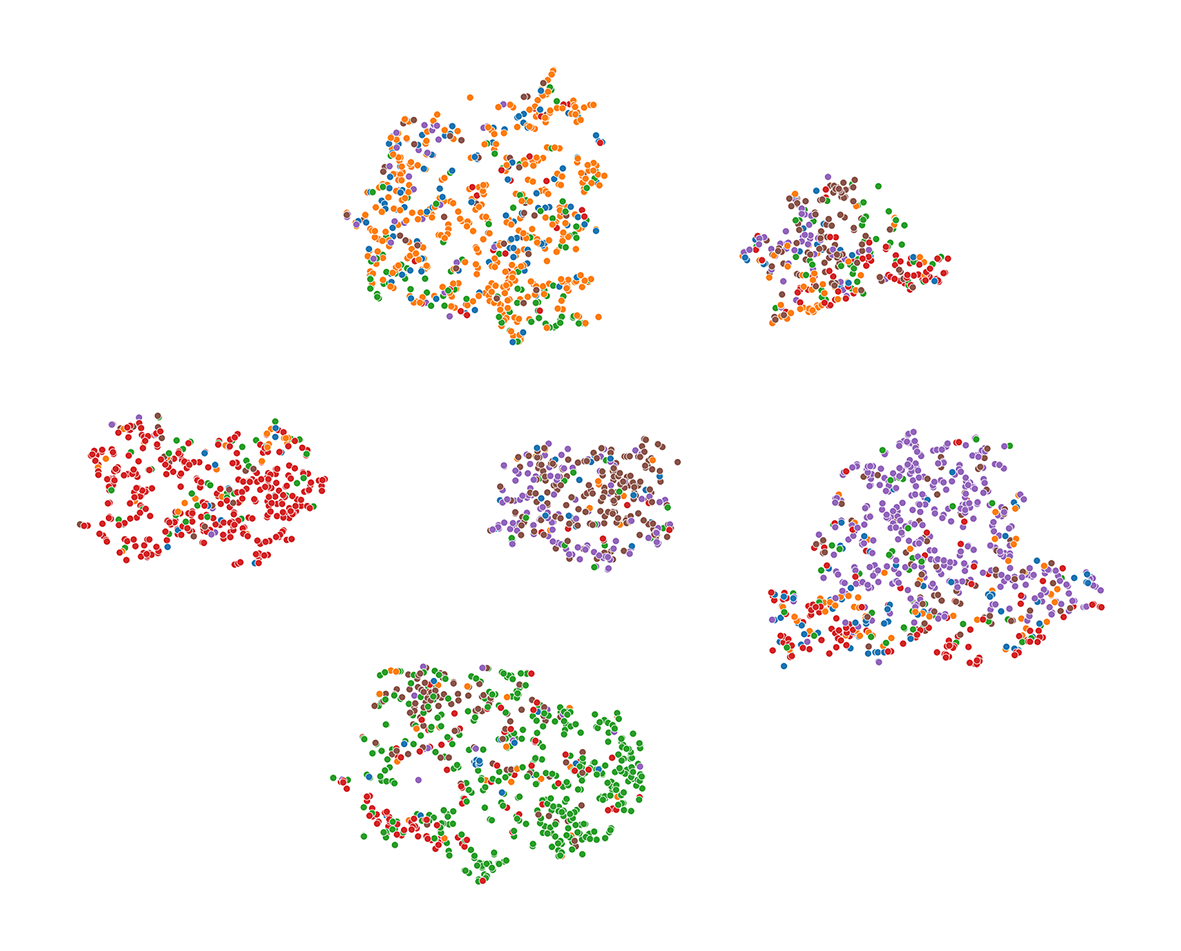} \\
    \includegraphics[width=0.9\textwidth]{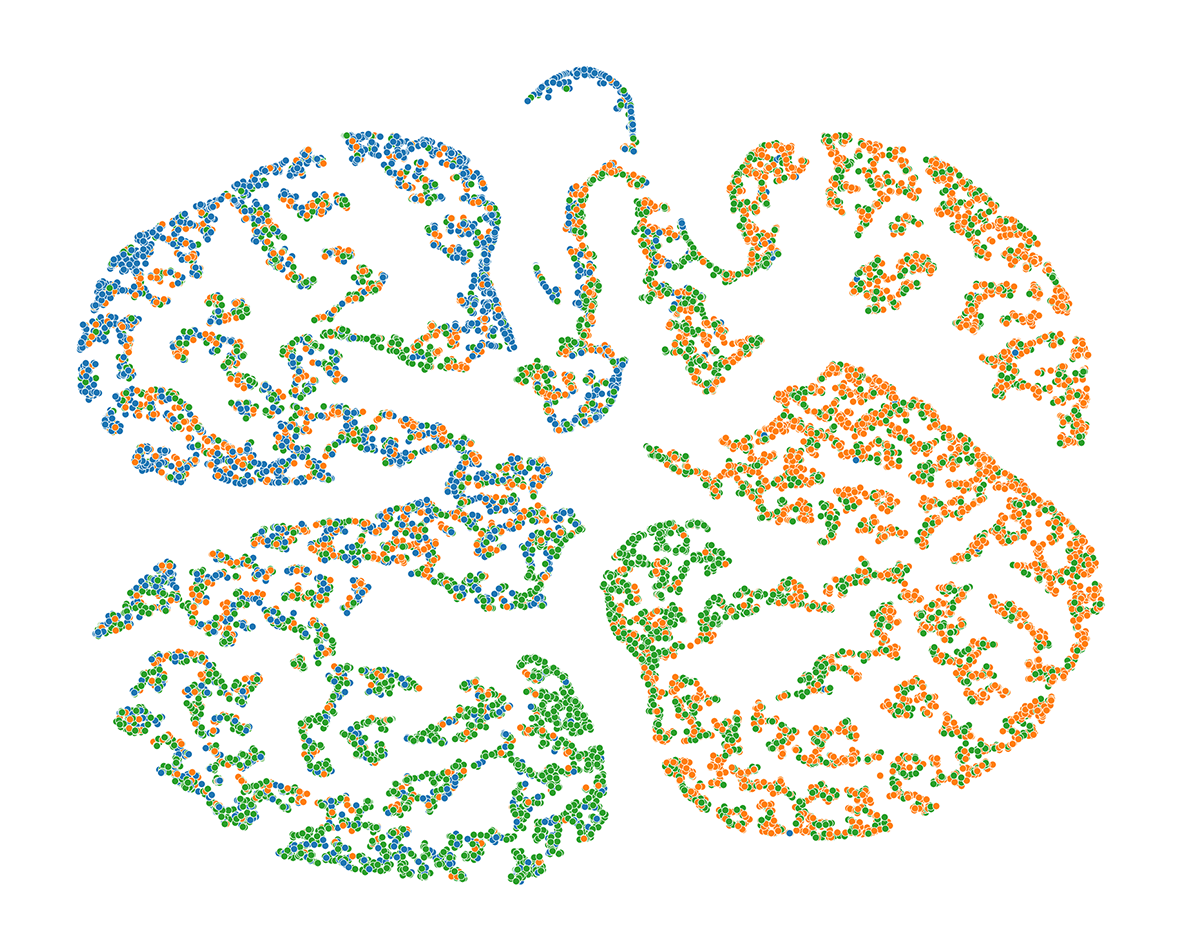} \\
    \includegraphics[width=0.9\textwidth]{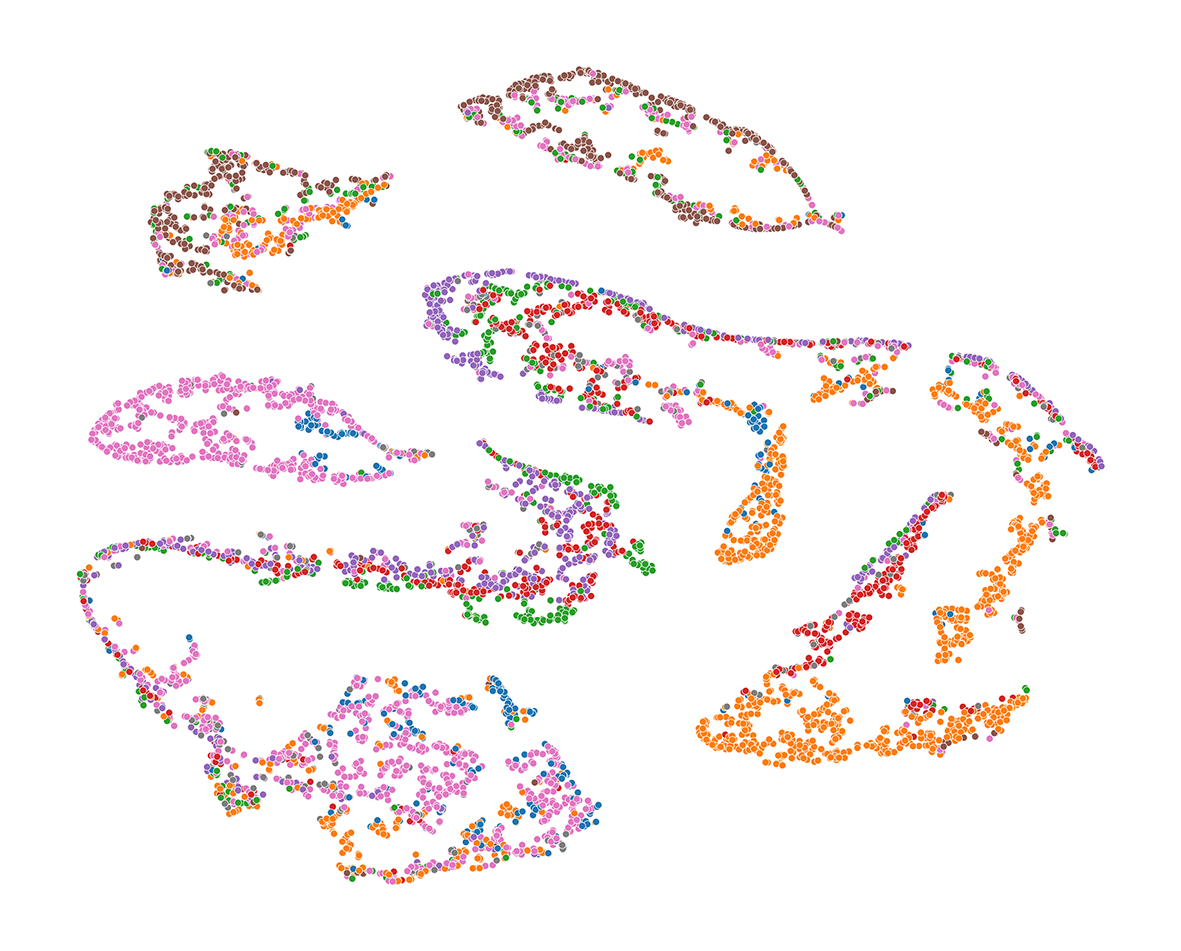} \\
    \includegraphics[width=0.9\textwidth]{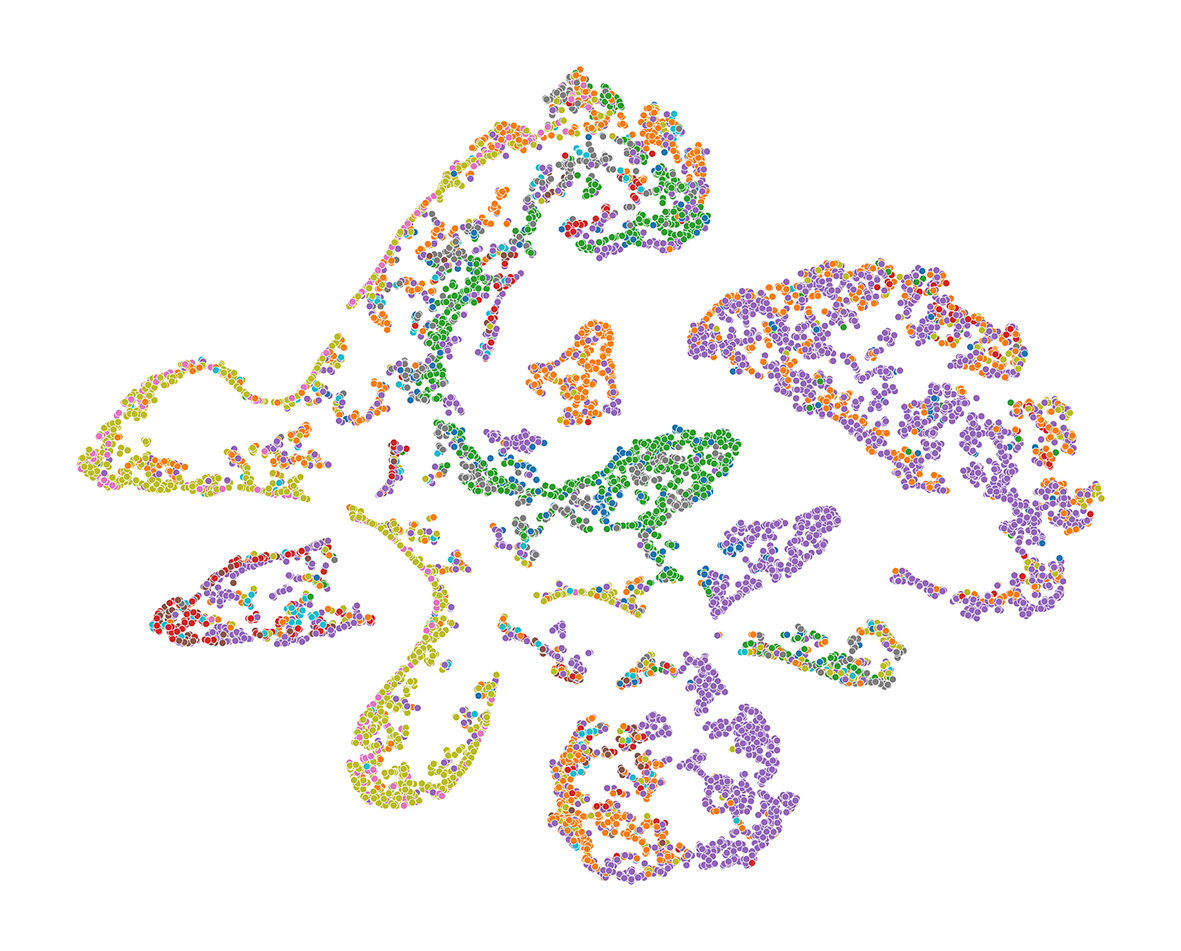}
  \end{minipage}
}
\subfigure[SUBLIME]{
  \begin{minipage}[c][\height][c]{0.15\linewidth}
    \centering
    \includegraphics[width=0.9\textwidth]{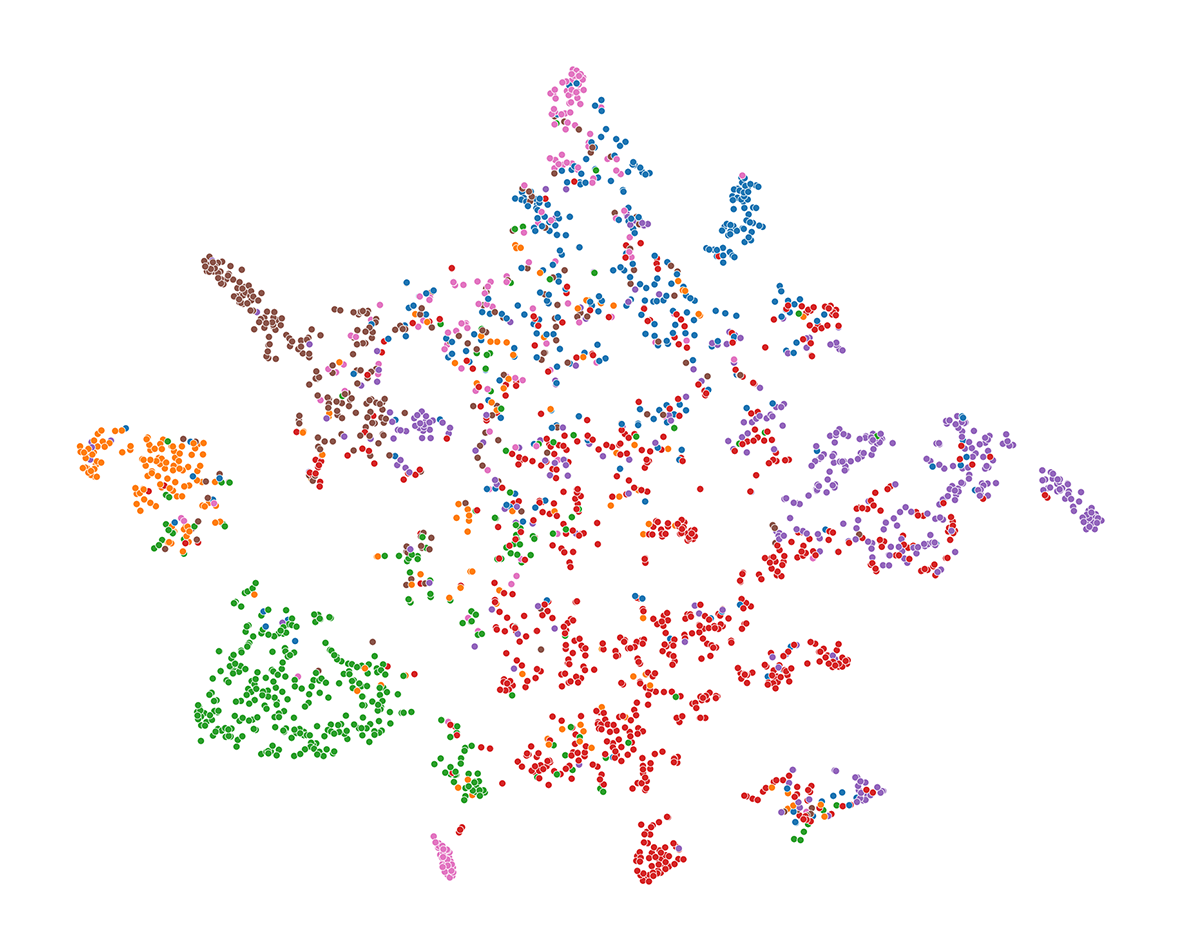} \\
    \includegraphics[width=0.9\textwidth]{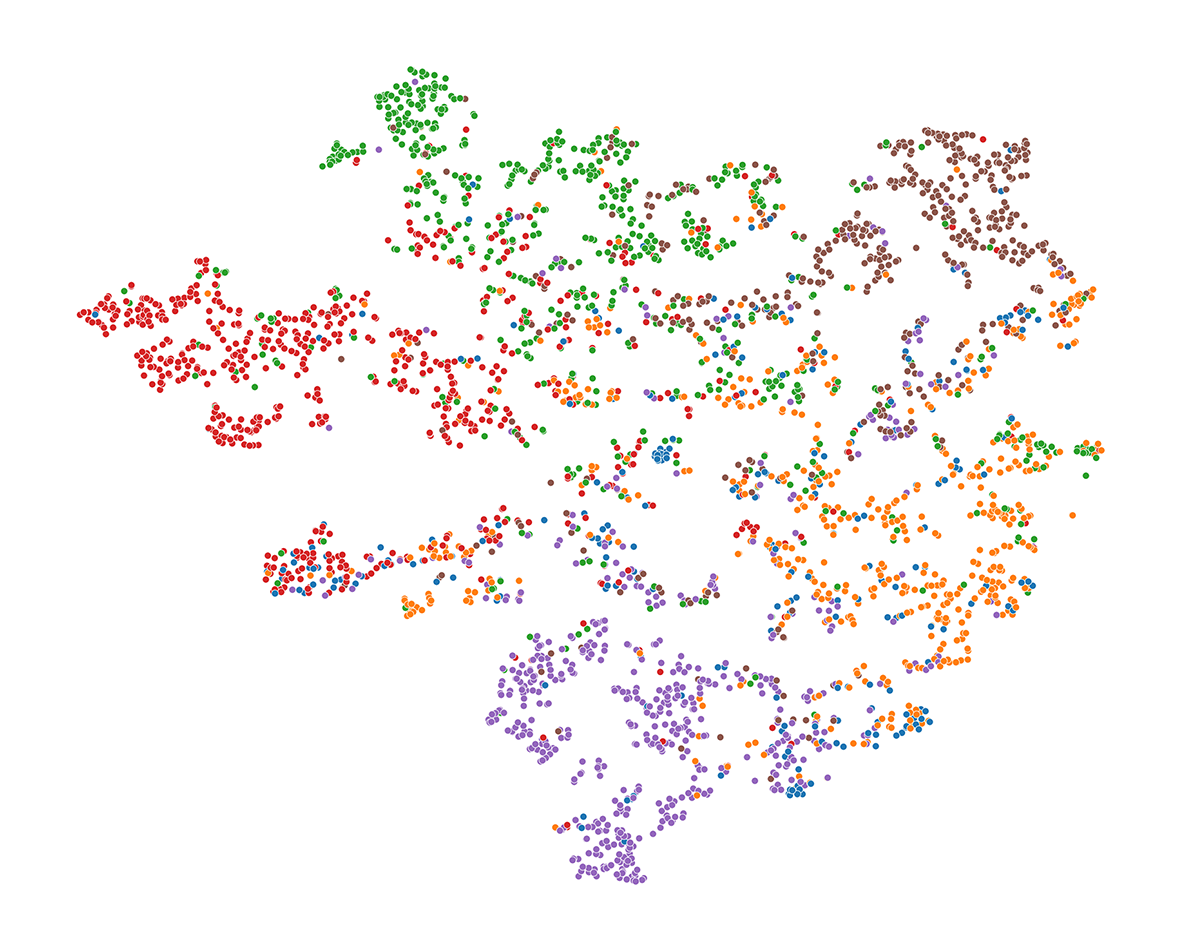} \\
    \includegraphics[width=0.9\textwidth]{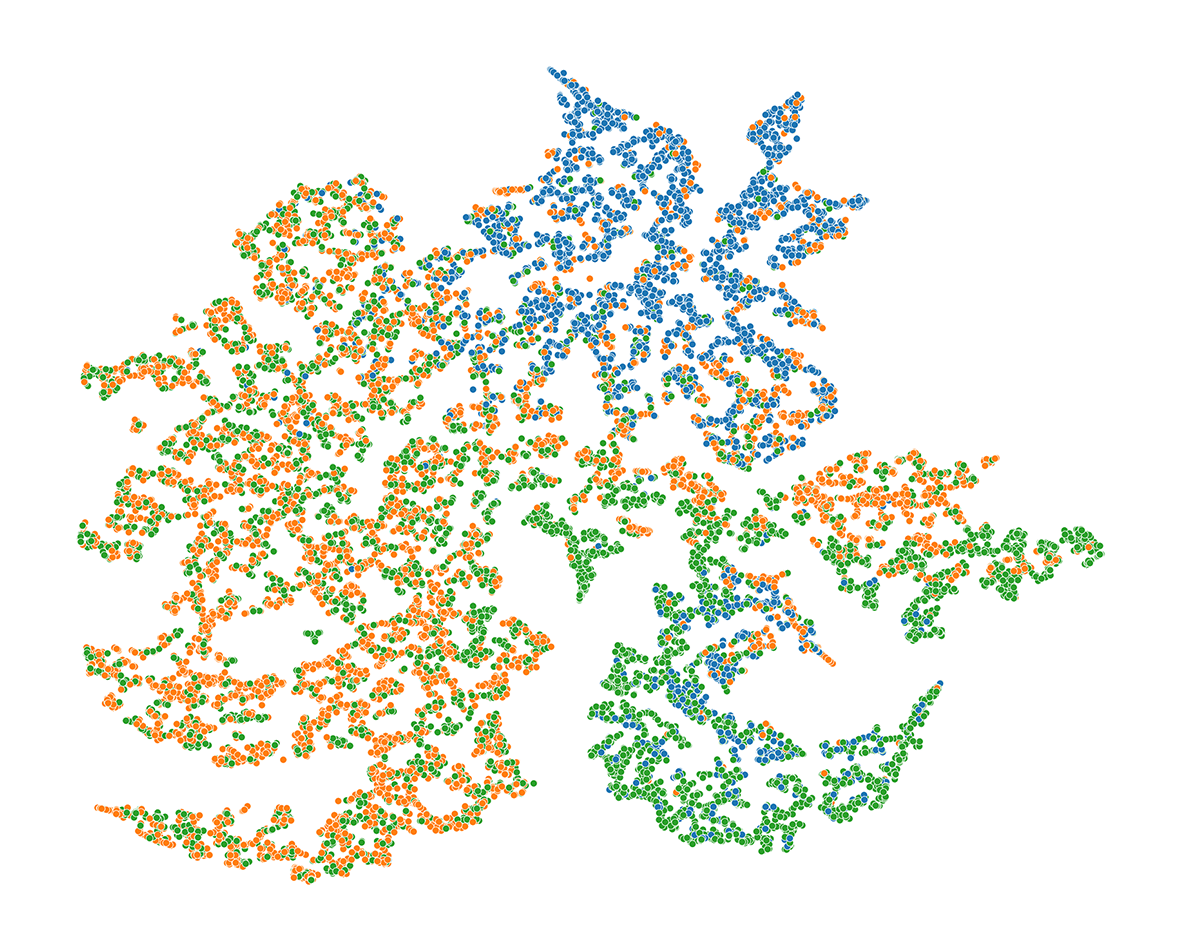} \\
    \includegraphics[width=0.9\textwidth]{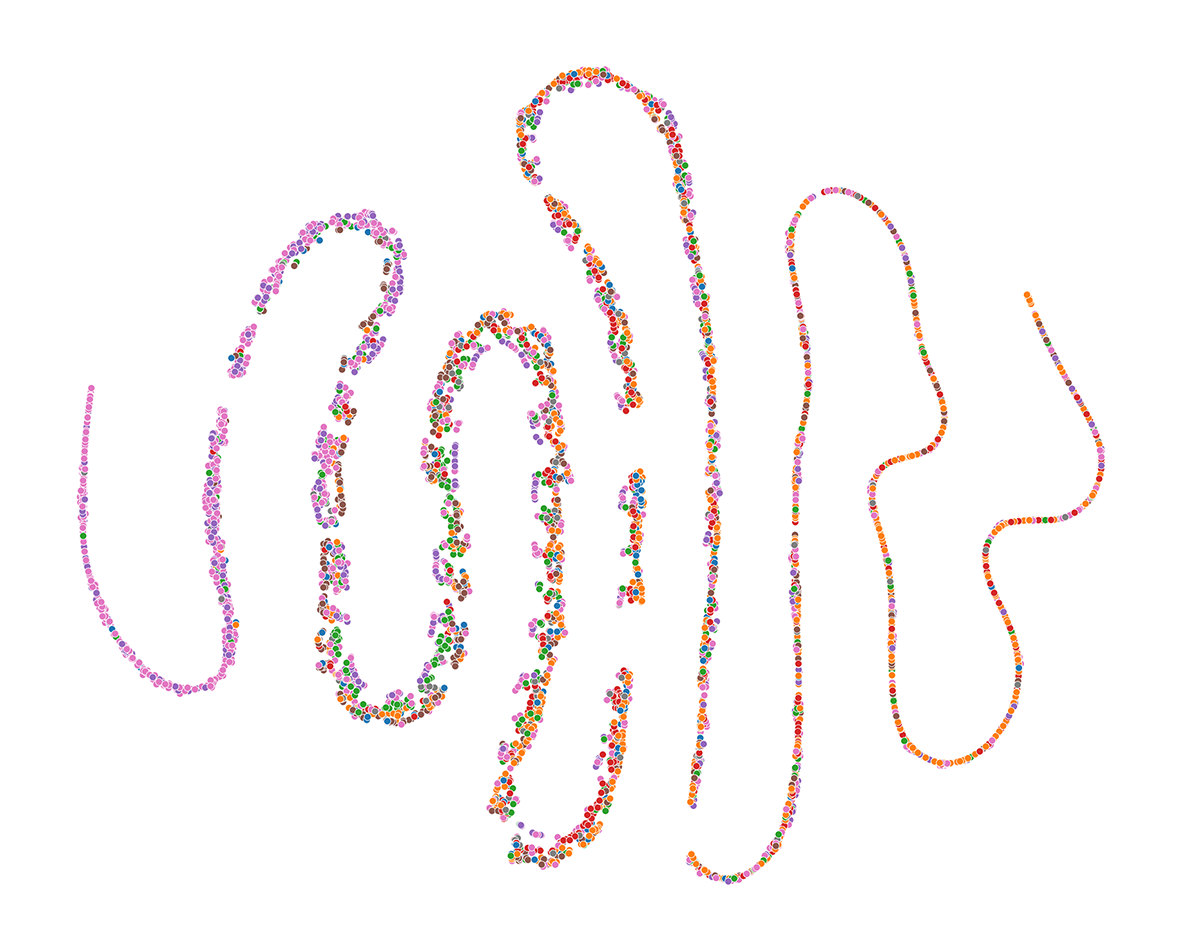} \\
    \includegraphics[width=0.9\textwidth]{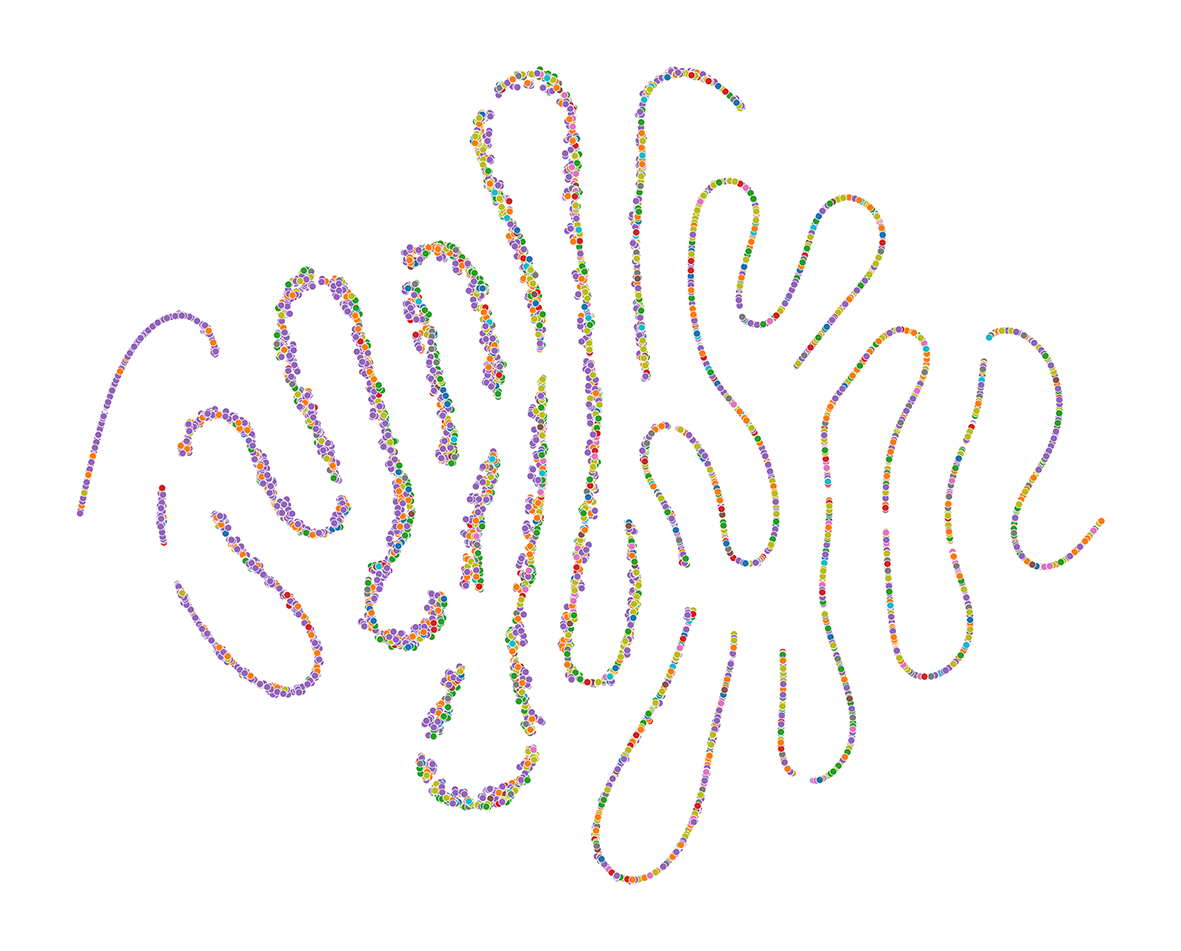}
  \end{minipage}
}
\subfigure[AGC-DRR]{
  \begin{minipage}[c][\height][c]{0.15\linewidth}
    \centering
    \includegraphics[width=0.9\textwidth]{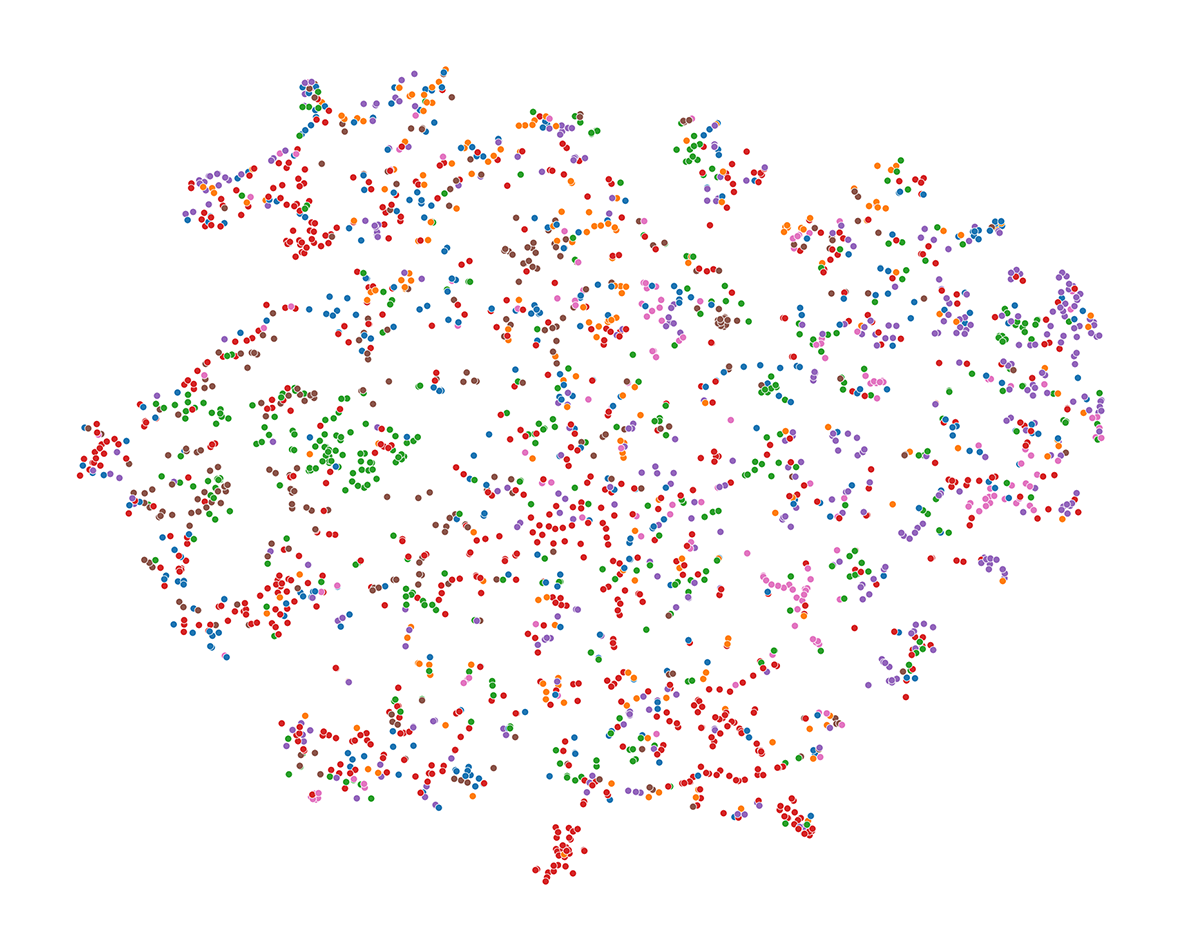} \\
    \includegraphics[width=0.9\textwidth]{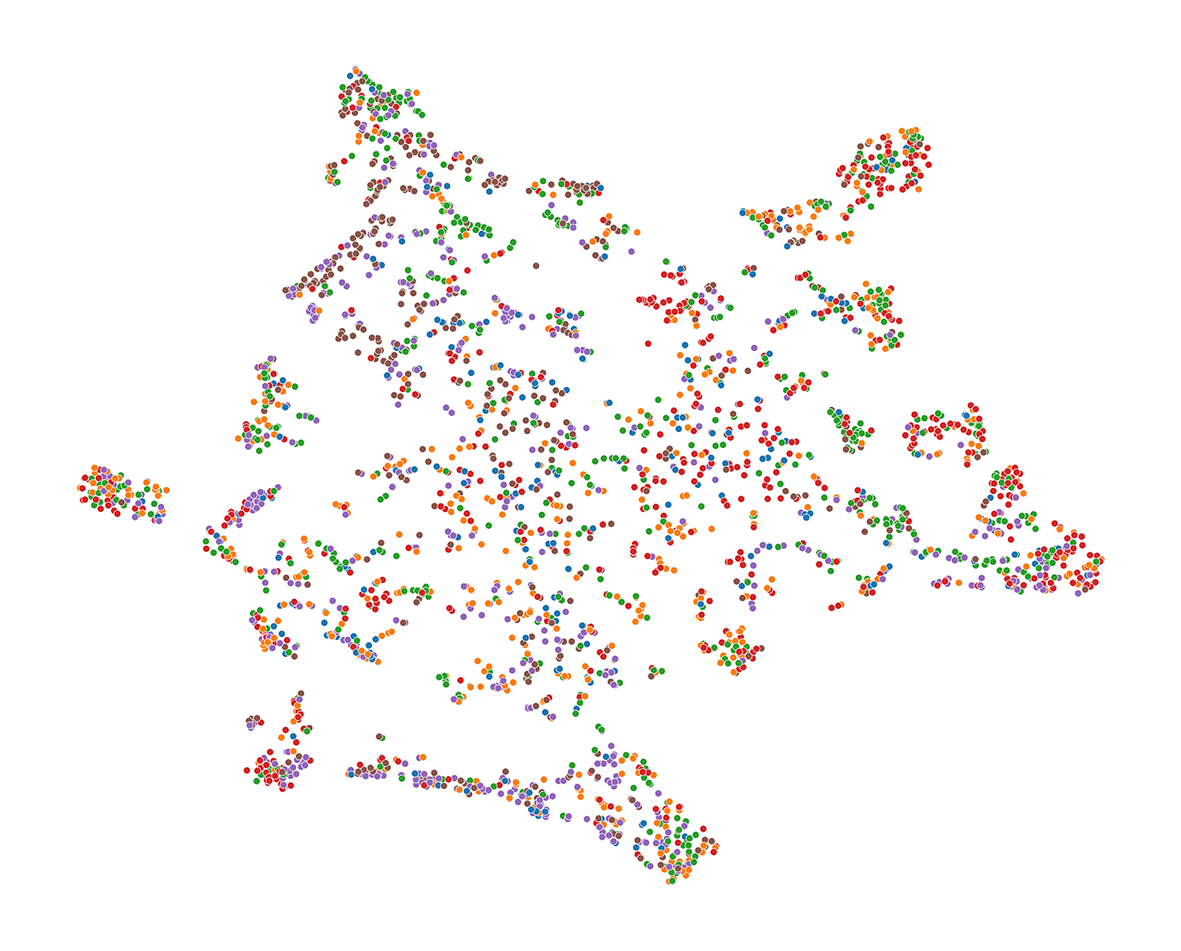} \\
    \includegraphics[width=0.9\textwidth]{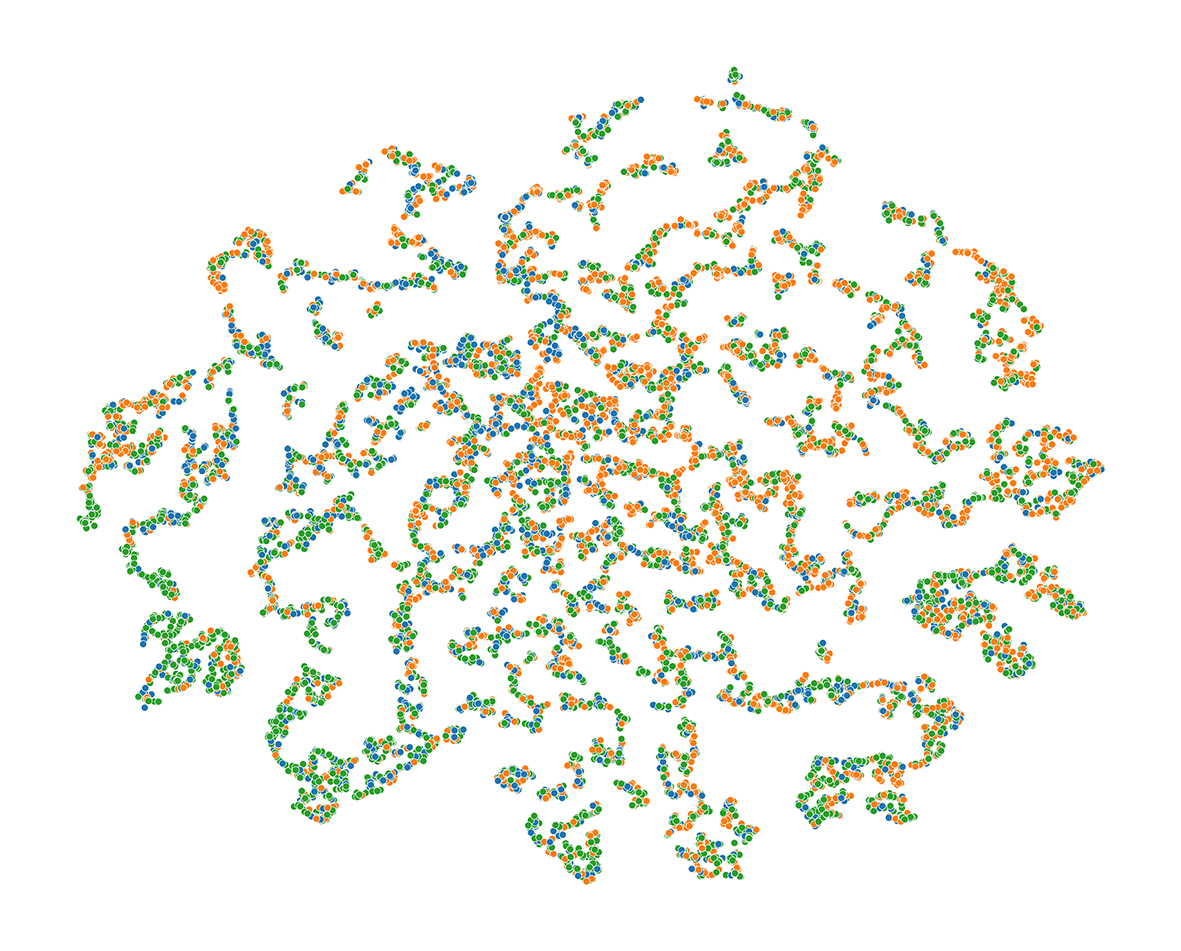} \\
    \includegraphics[width=0.9\textwidth]{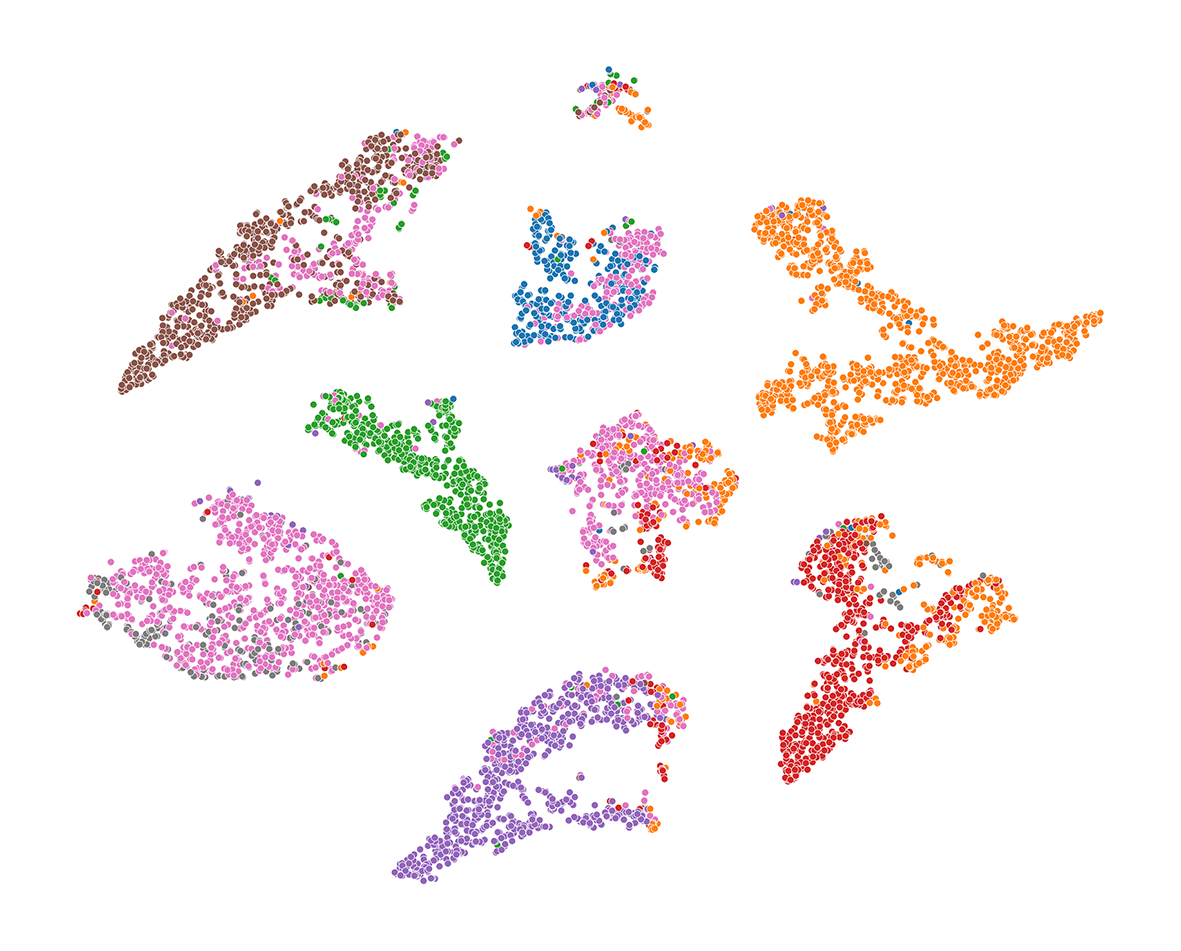} \\
    \includegraphics[width=0.9\textwidth]{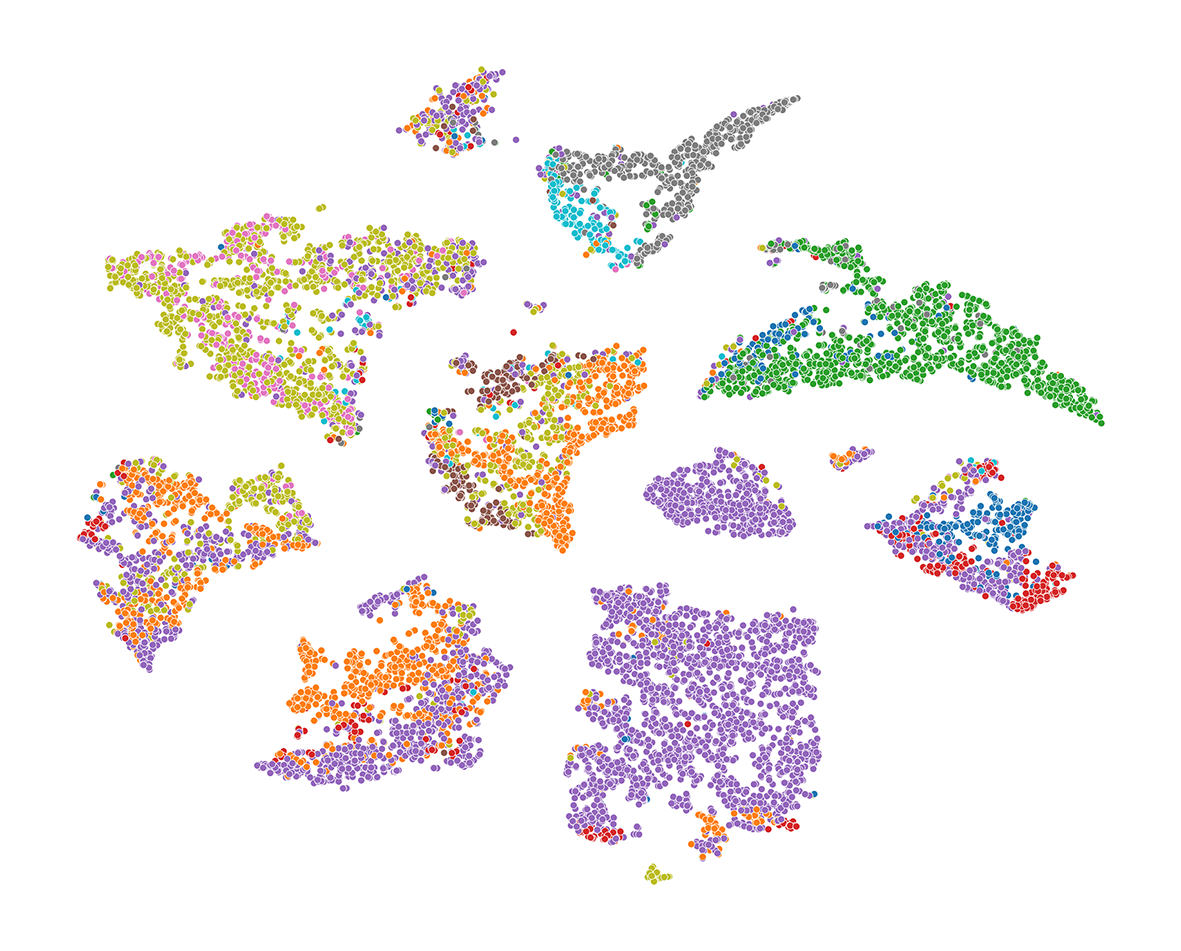}
  \end{minipage}
}
\subfigure[CONVERT]{
  \begin{minipage}[c][\height][c]{0.15\linewidth}
    \centering
    \includegraphics[width=0.9\textwidth]{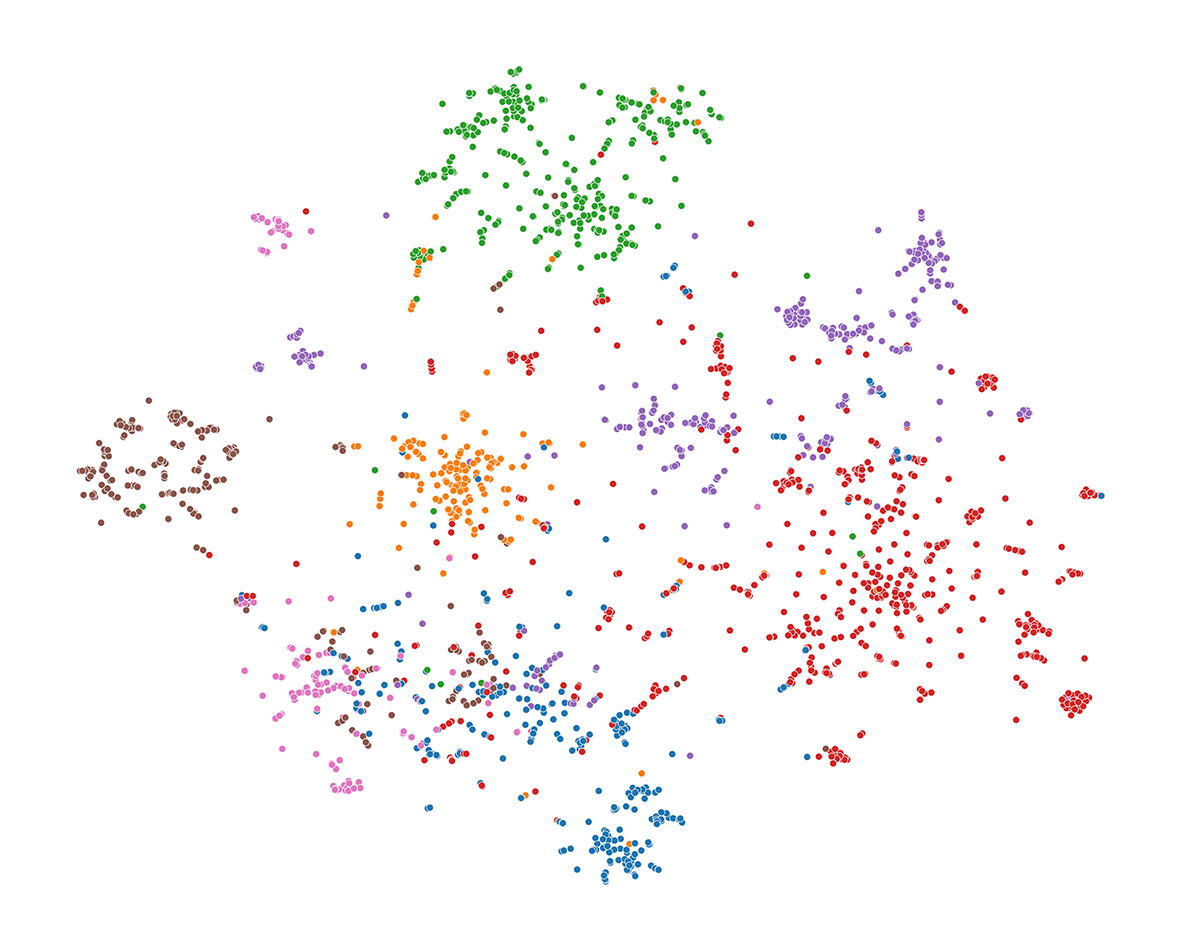} \\
    \includegraphics[width=0.9\textwidth]{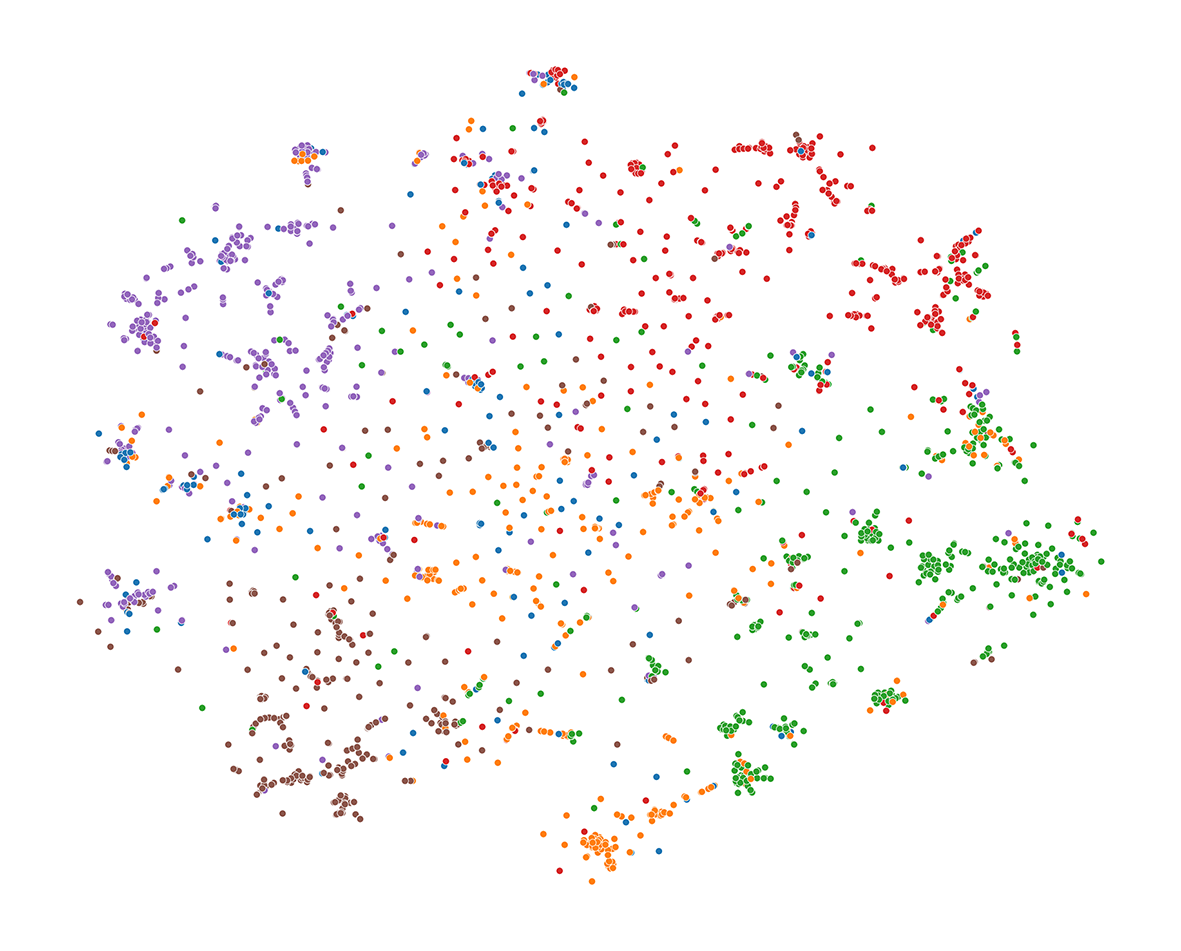} \\
    \includegraphics[width=0.9\textwidth]{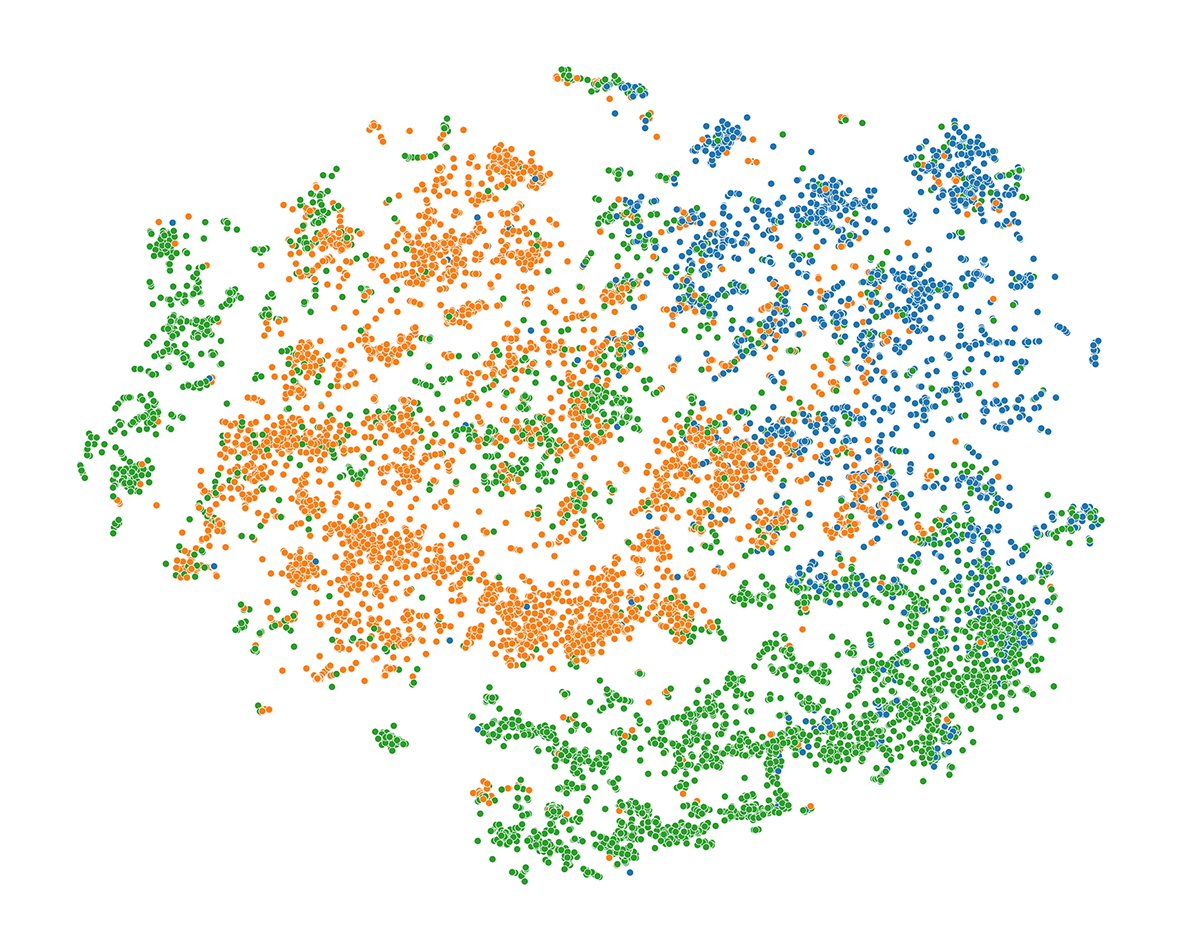} \\
    \includegraphics[width=0.9\textwidth]{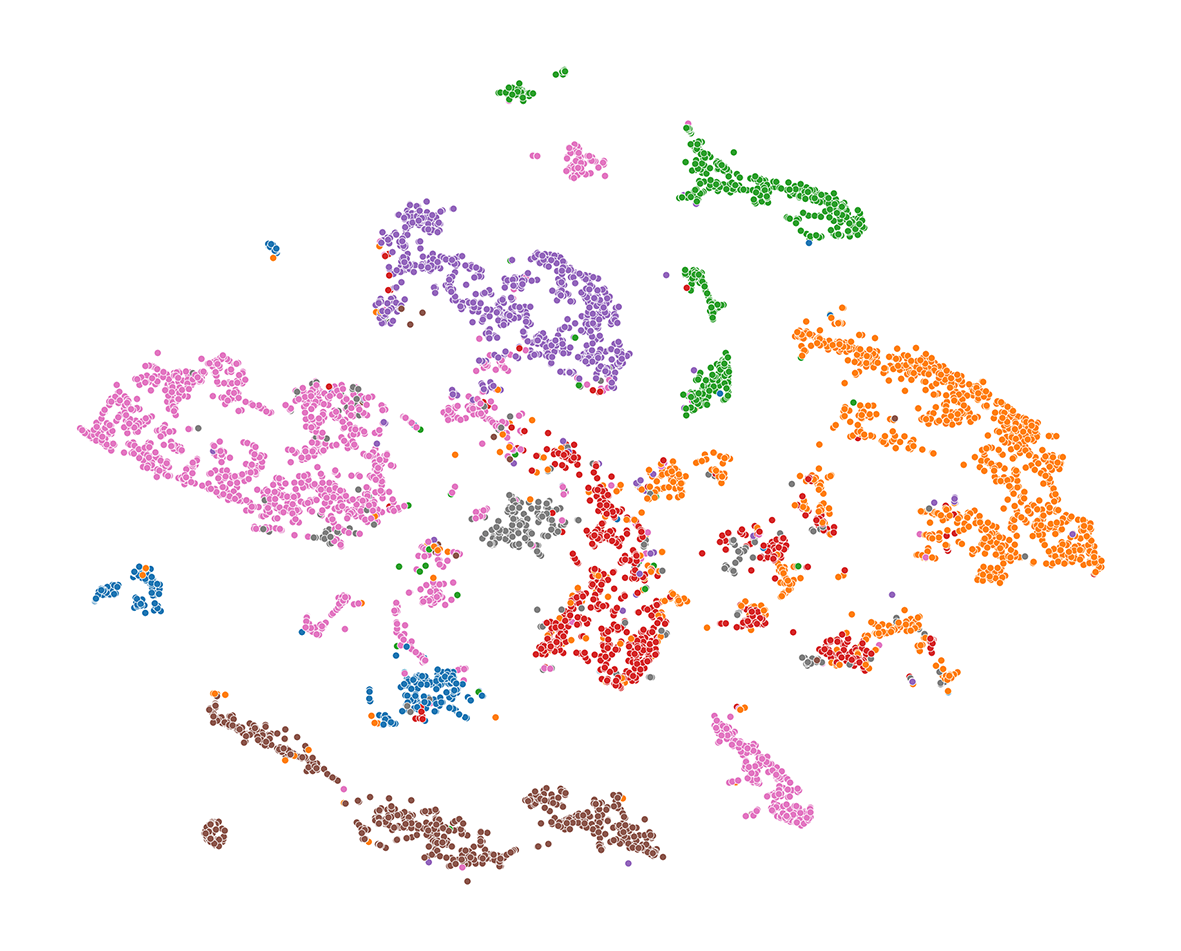} \\
    \includegraphics[width=0.9\textwidth]{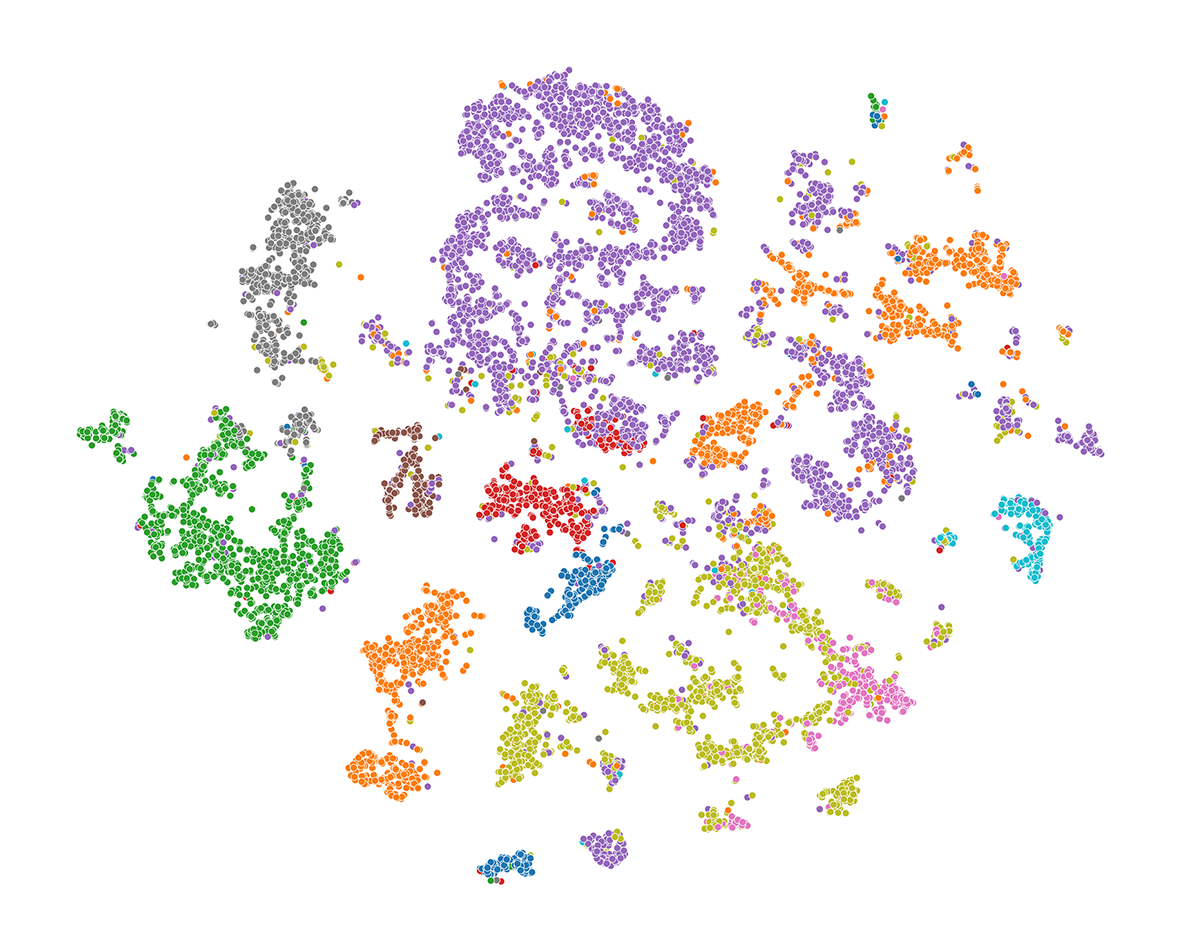}
  \end{minipage}
}
\subfigure[RDSA]{
  \begin{minipage}[c][\height][c]{0.15\linewidth}
    \centering
    \includegraphics[width=0.9\textwidth]{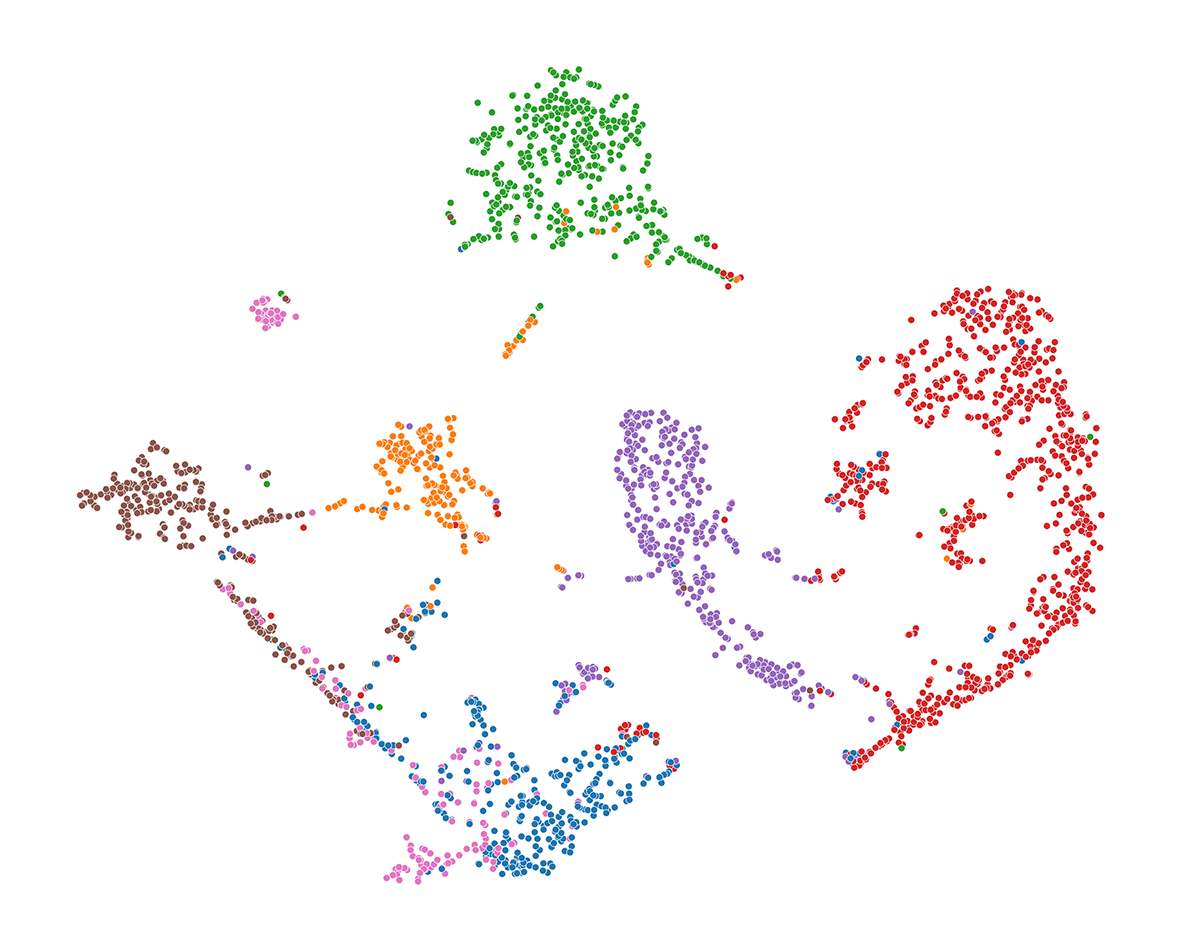} \\
    \includegraphics[width=0.9\textwidth]{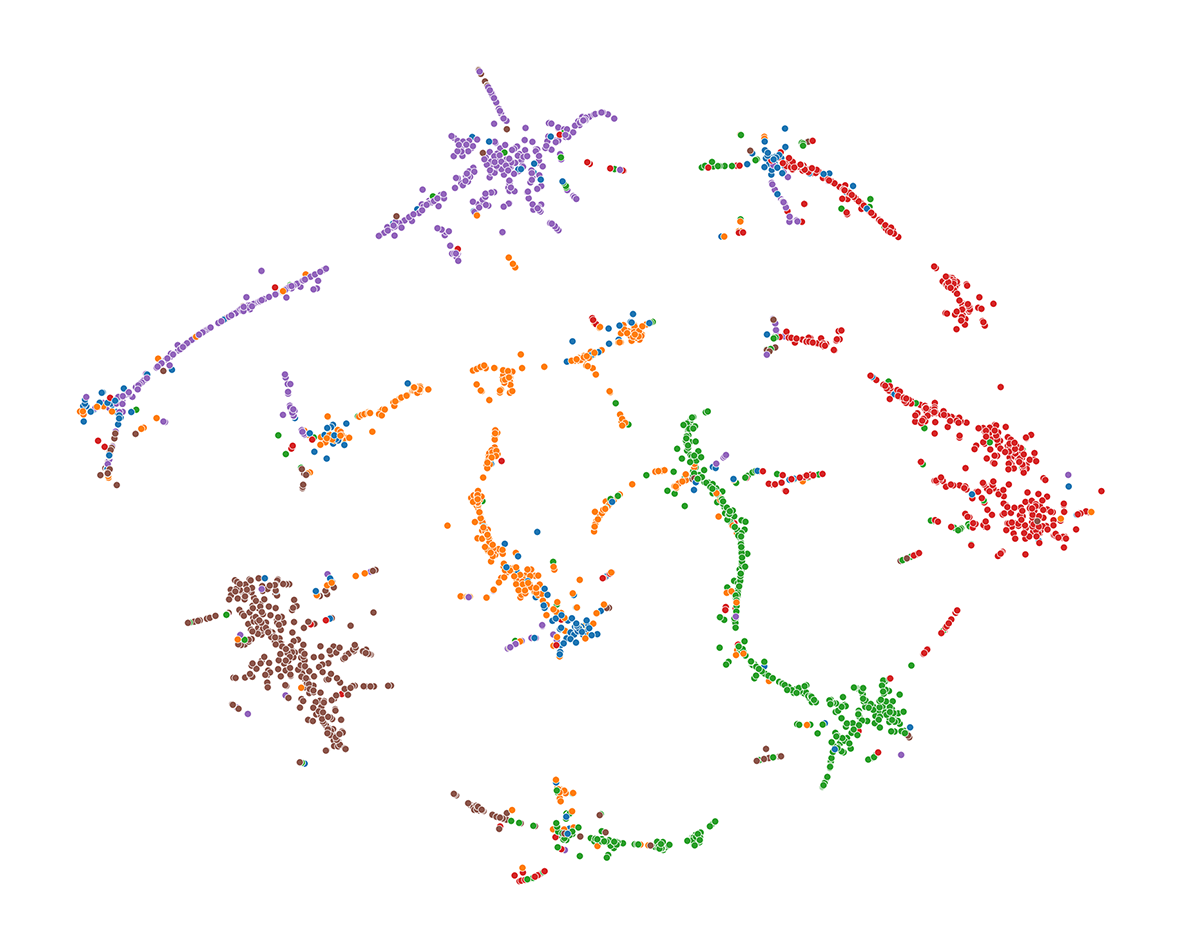} \\
    \includegraphics[width=0.9\textwidth]{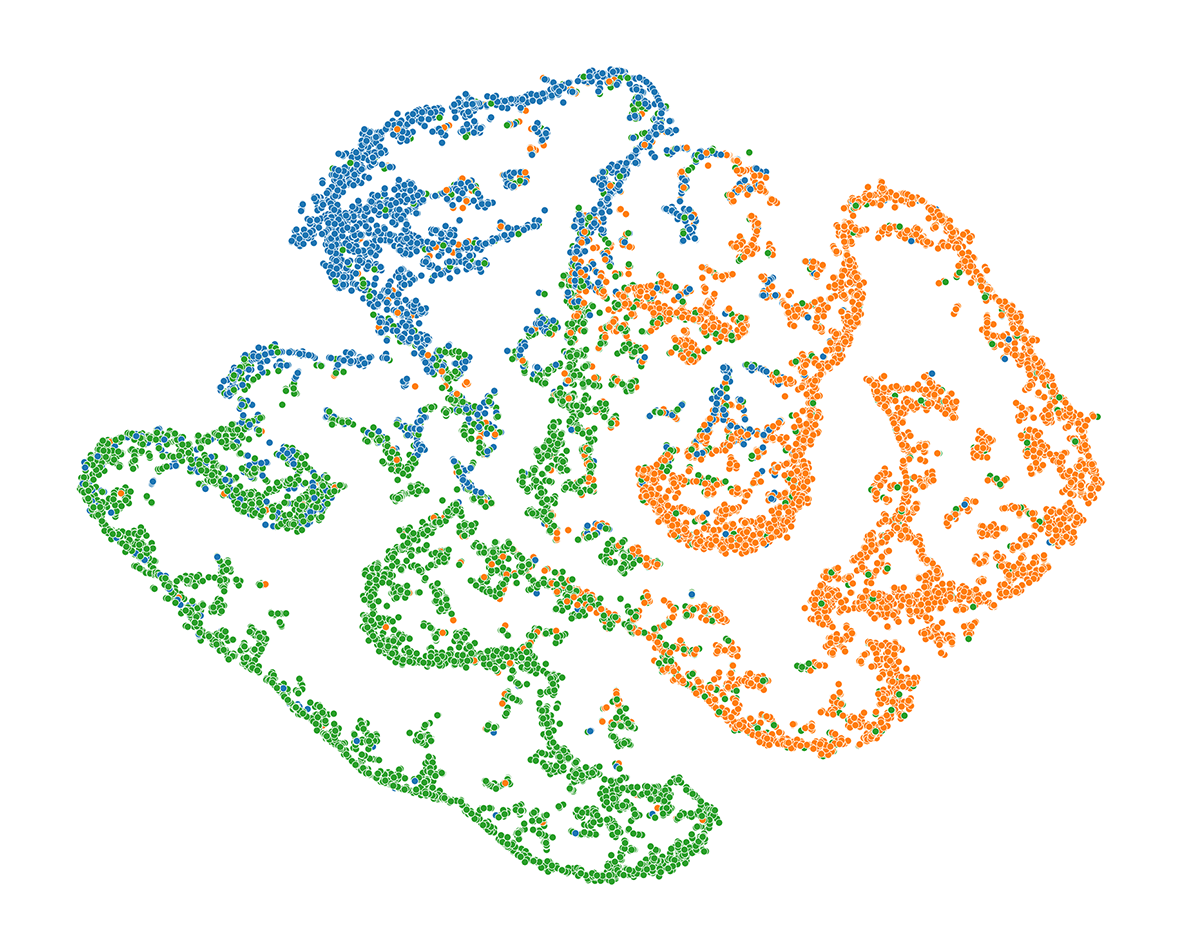} \\
    \includegraphics[width=0.9\textwidth]{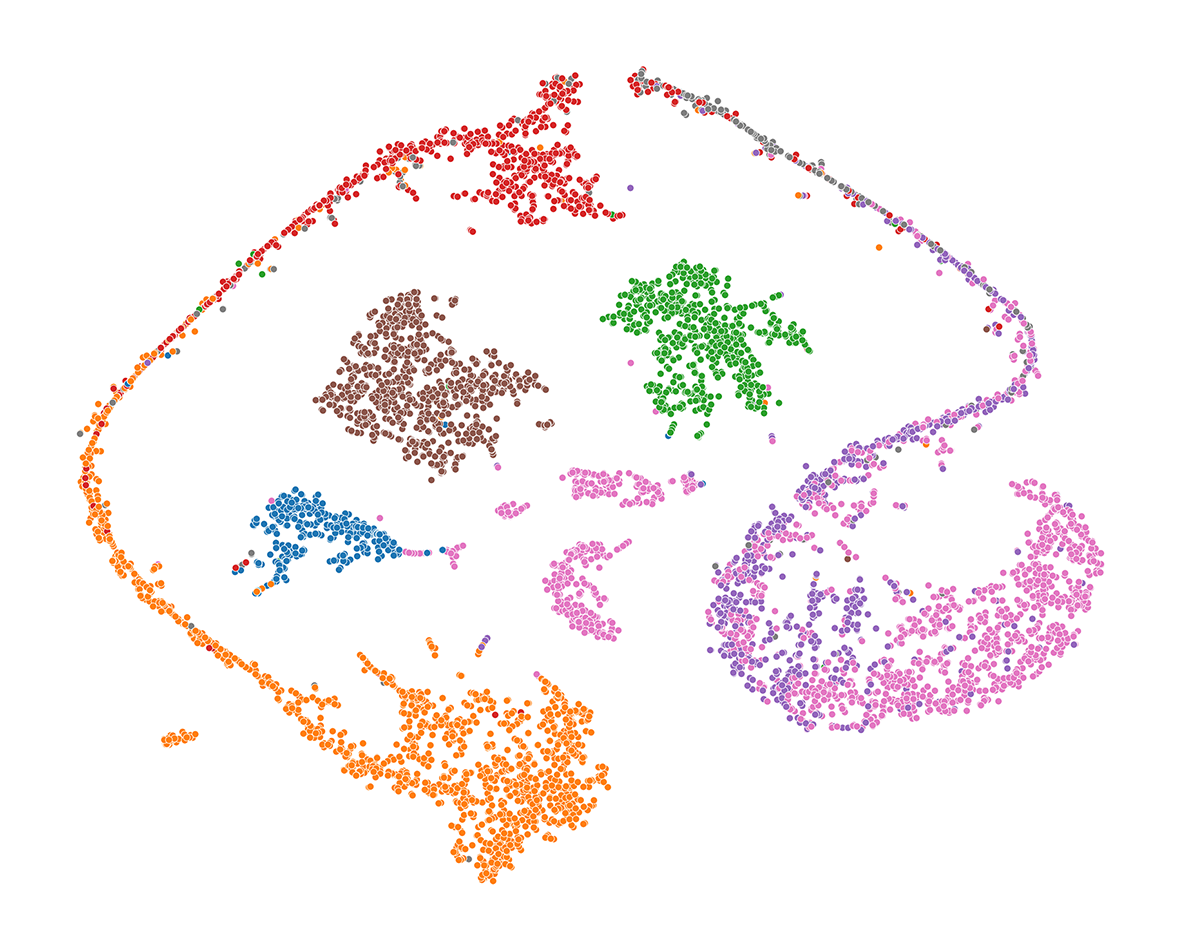} \\
    \includegraphics[width=0.9\textwidth]{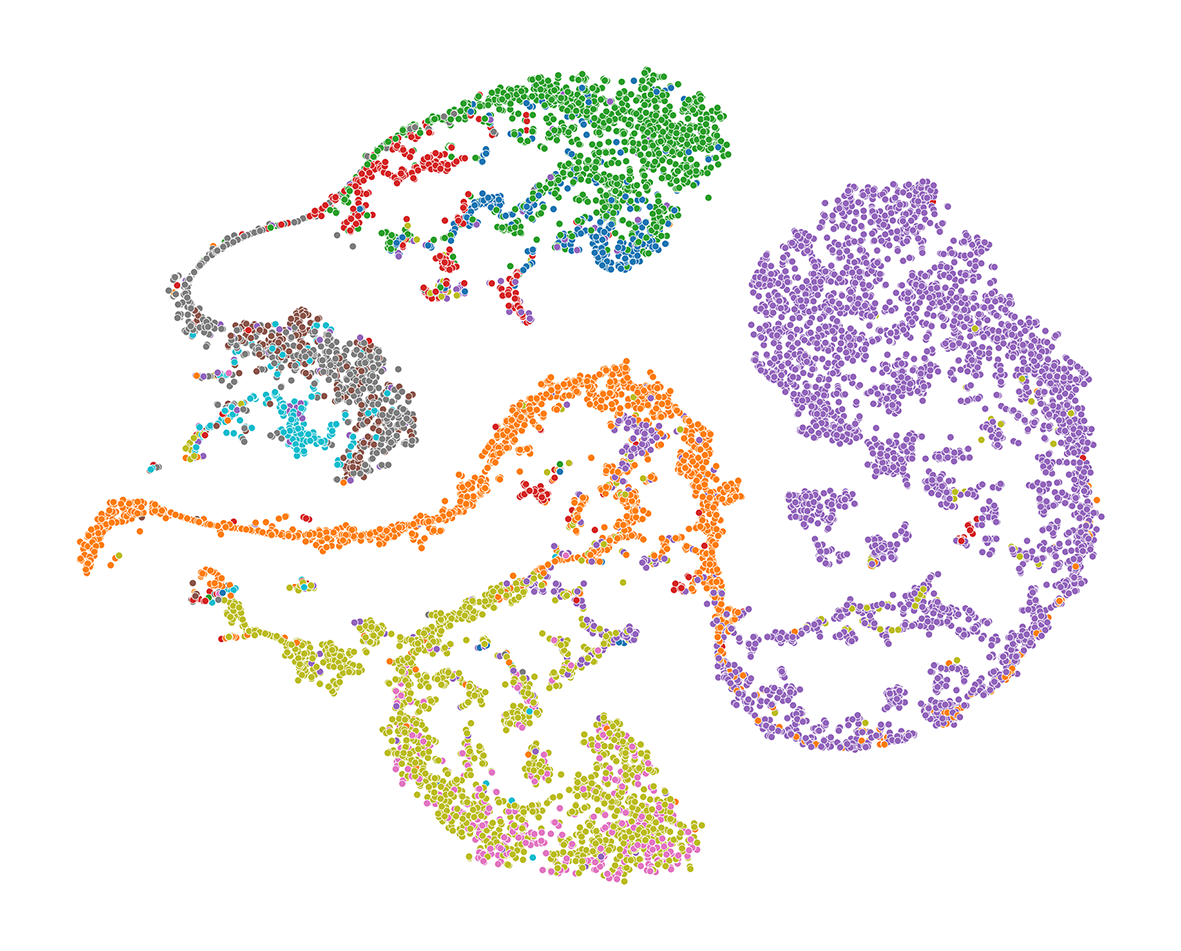}
  \end{minipage}
}
\end{figure*}

\subsubsection*{\textbf{Implementation and Hyperparameter Settings}}
The autoencoder's encoder and decoder utilize a three-layer MLP, while the GNN employs GraphSAGE with three layers, each possessing hidden dimensions of 256, 128, and 64 correspondingly. ReLU serves as the activation function for the autoencoder, while SeLU is employed for the GNN. The Adam optimizer operates with a learning rate of 0.001 throughout. The training process spans 300 epochs. Notably, for the Amazon Computers dataset, the hyperparameter $\sigma$ is set at $0.4$, while for other datasets, it stands at $0.5$. The auxiliary information loss weight $\alpha$ is set at $0.2$ in all experiments.

\subsection{Baselines Comparison}
\label{sec:baseline}
In this section, we present experiments to evaluate the performance of RDSA compared with eight baseline models across seven datasets. Table \ref{table:performance} provides a summary of the experimental results. The primary findings are as follows: \textbf{1)} Our method outperforms existing graph clustering techniques, setting new performance benchmarks; \textbf{2)} Hybrid approaches generally exceed the performance of both MLP-based and GNN-based methods on most datasets, suggesting that MLP and GNN models may not effectively capture the diversity within graph data; \textbf{3)} RDSA uses dual soft assignment objectives to incorporate both structural and attribute information from graph data, leading to improved performance over other hybrid methods. Specifically, our approach achieves the top scores in $85\%$ of evaluation metrics across the tested datasets. For instance, on the citation network datasets, our method provides a $4.96\%$ increase in ACC, a $6.00\%$ improvement in NMI, a $9.00\%$ increase in ARI, and a $1.89\%$ increase in F1 score over the second-best method, supporting the effectiveness of our framework. 

Additionally, we performed experiments on two large-scale datasets. While all baseline methods, except MAGI, encounter memory limitations on these datasets, our method consistently delivers the best performance. On the ogbn-arxiv dataset, RDSA surpasses MAGI by $8.99\%$ in accuracy and $8.12\%$ in ARI, while on the ogbn-products dataset, RDSA exceeds MAGI by $11.5\%$ in accuracy and $32.9\%$ in ARI, indicating RDSA's higher consistency in clustering boundaries. Although minor differences in information alignment and small-cluster identification result in NMI and F1 scores that are slightly lower than those of MAGI, these variations do not affect the overall clustering quality. These results underscore the efficacy of our proposed framework in enhancing clustering performance within deep graph clustering frameworks.

\begin{figure*}[t]
  \centering
  \caption{Line plots showing the stability during training of four methods (SCGC, SUBLIME, AGC-DRR, CONVERT, and RDSA) on four datasets (Cora, PubMed, and Amazon Computers).}
  \label{fig:stability}
  \vspace{-2em}
  \begin{minipage}[t]{0.33\linewidth}
    \centering
    \caption*{Cora}
    \includegraphics[width=0.48\textwidth]{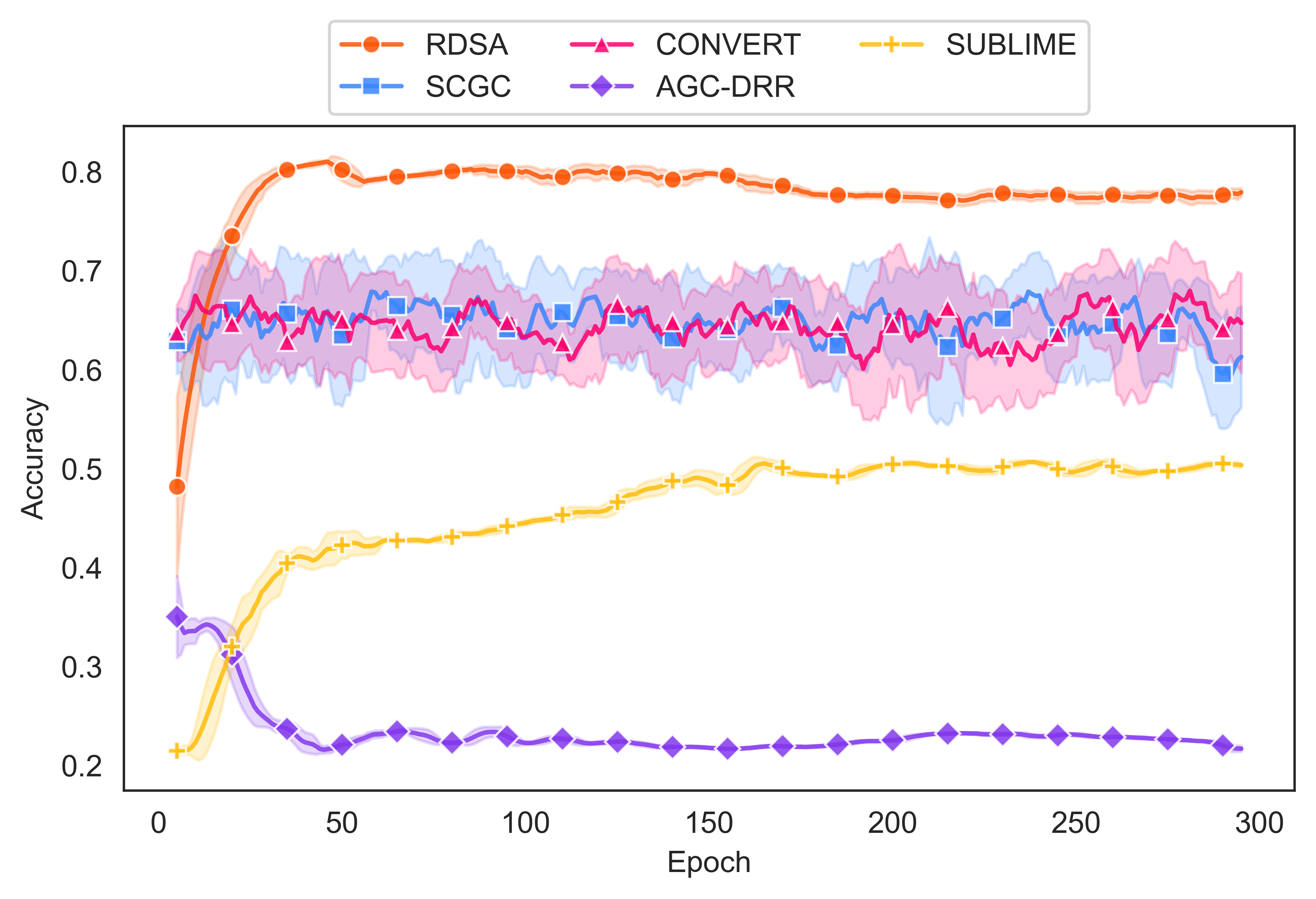}
    \includegraphics[width=0.48\textwidth]{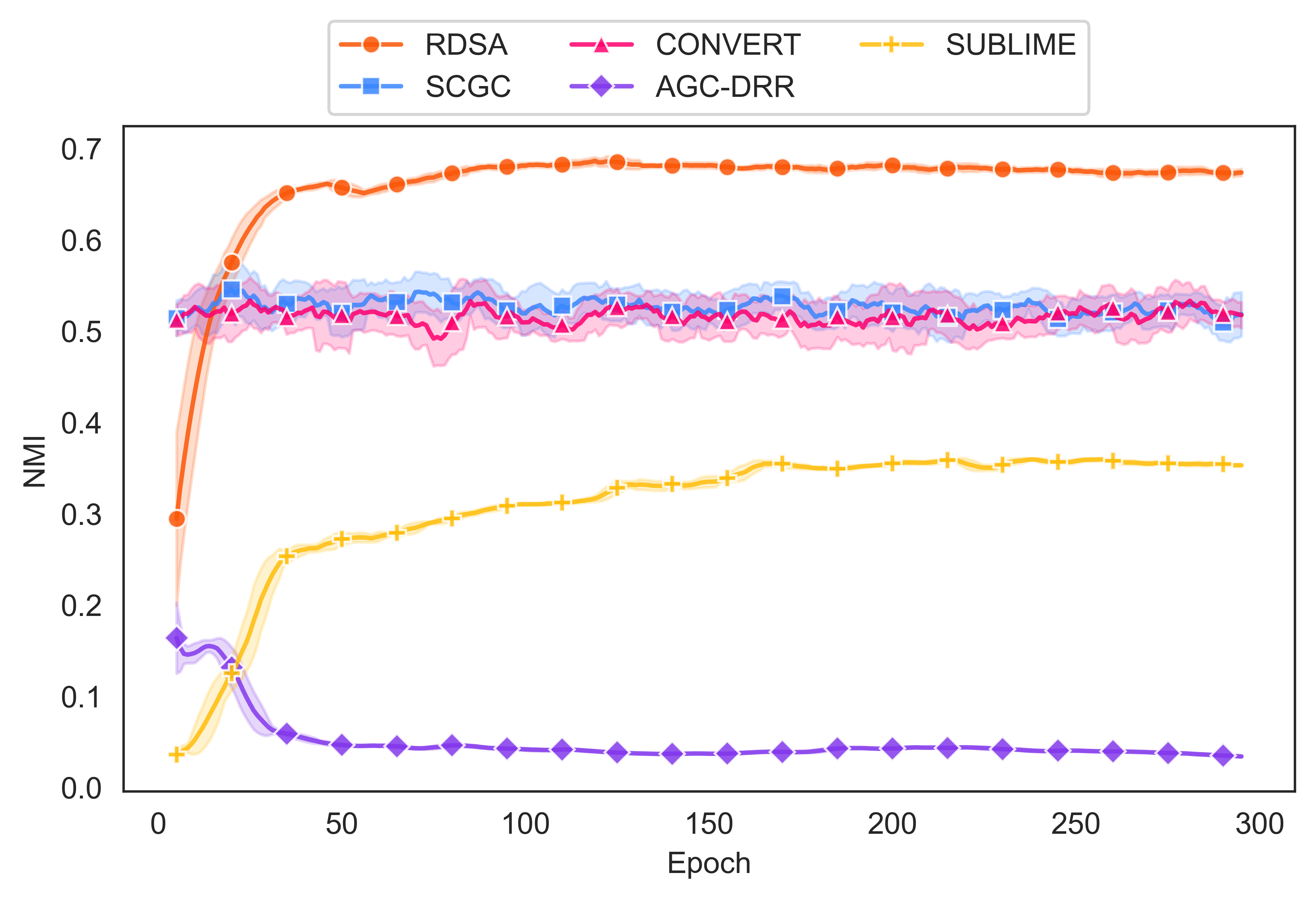} \\
    \includegraphics[width=0.48\textwidth]{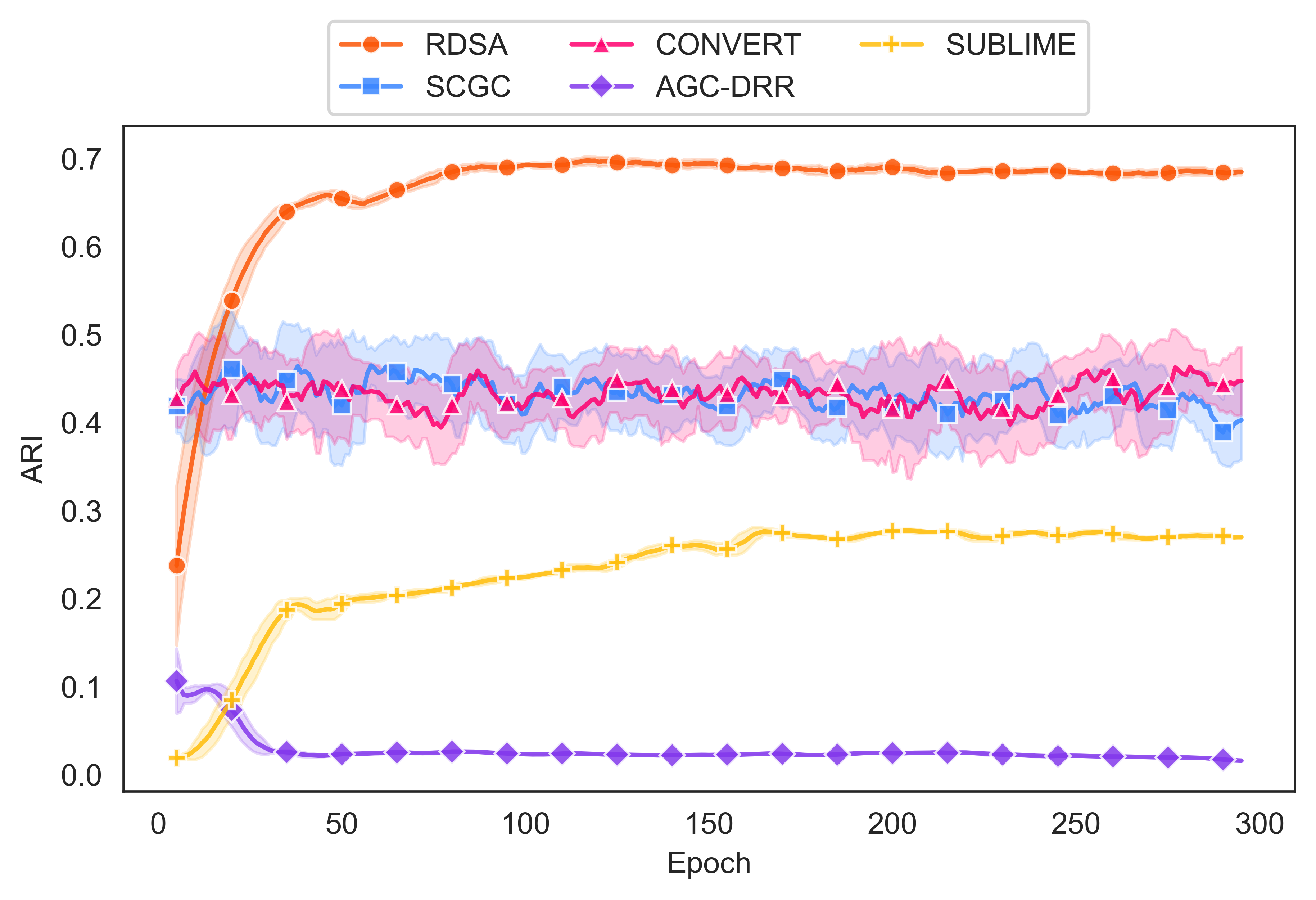}
    \includegraphics[width=0.48\textwidth]{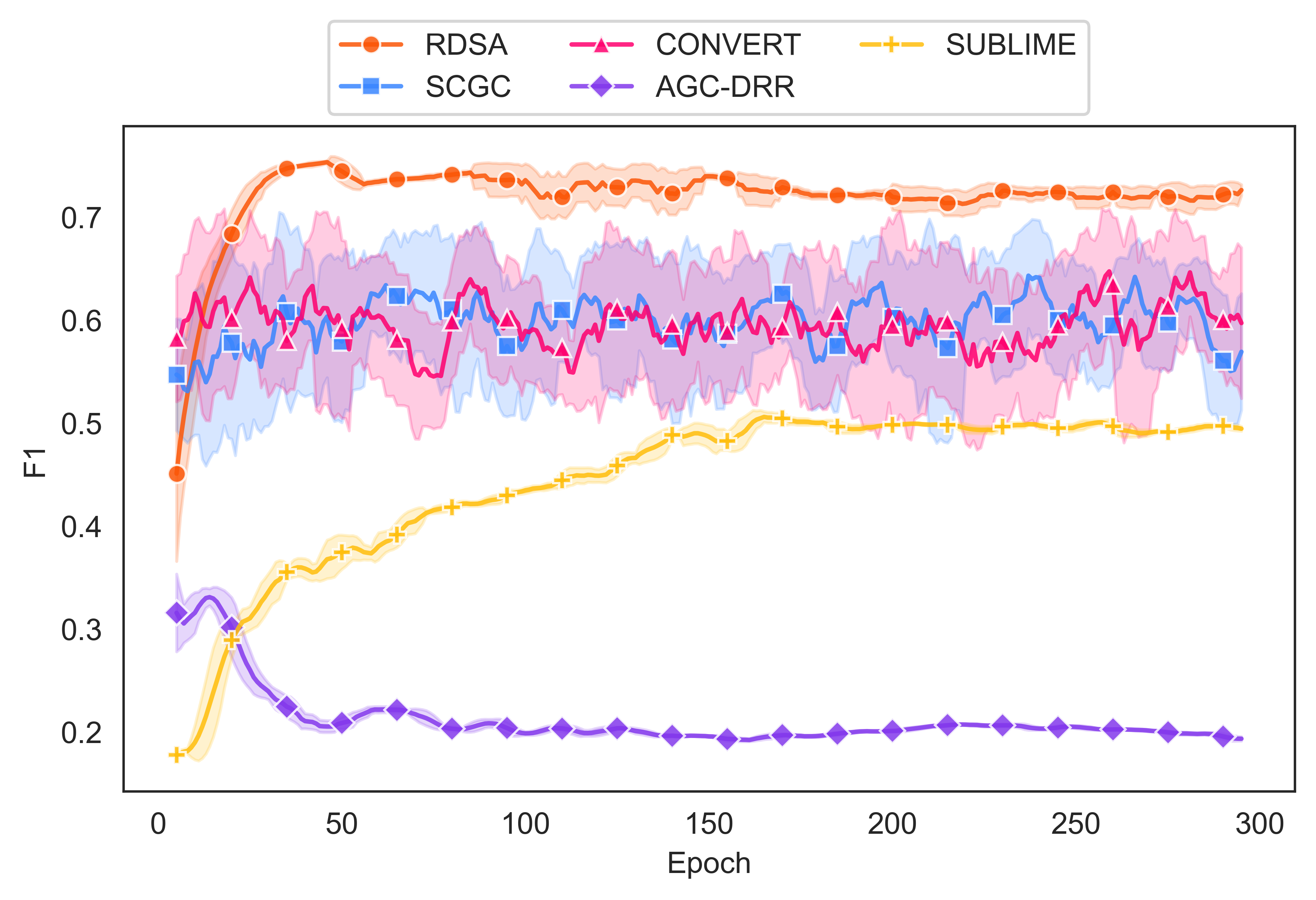}
  \end{minipage}
  \begin{minipage}[t]{0.33\linewidth}
    \centering
    \caption*{PubMed}
    \includegraphics[width=0.48\textwidth]{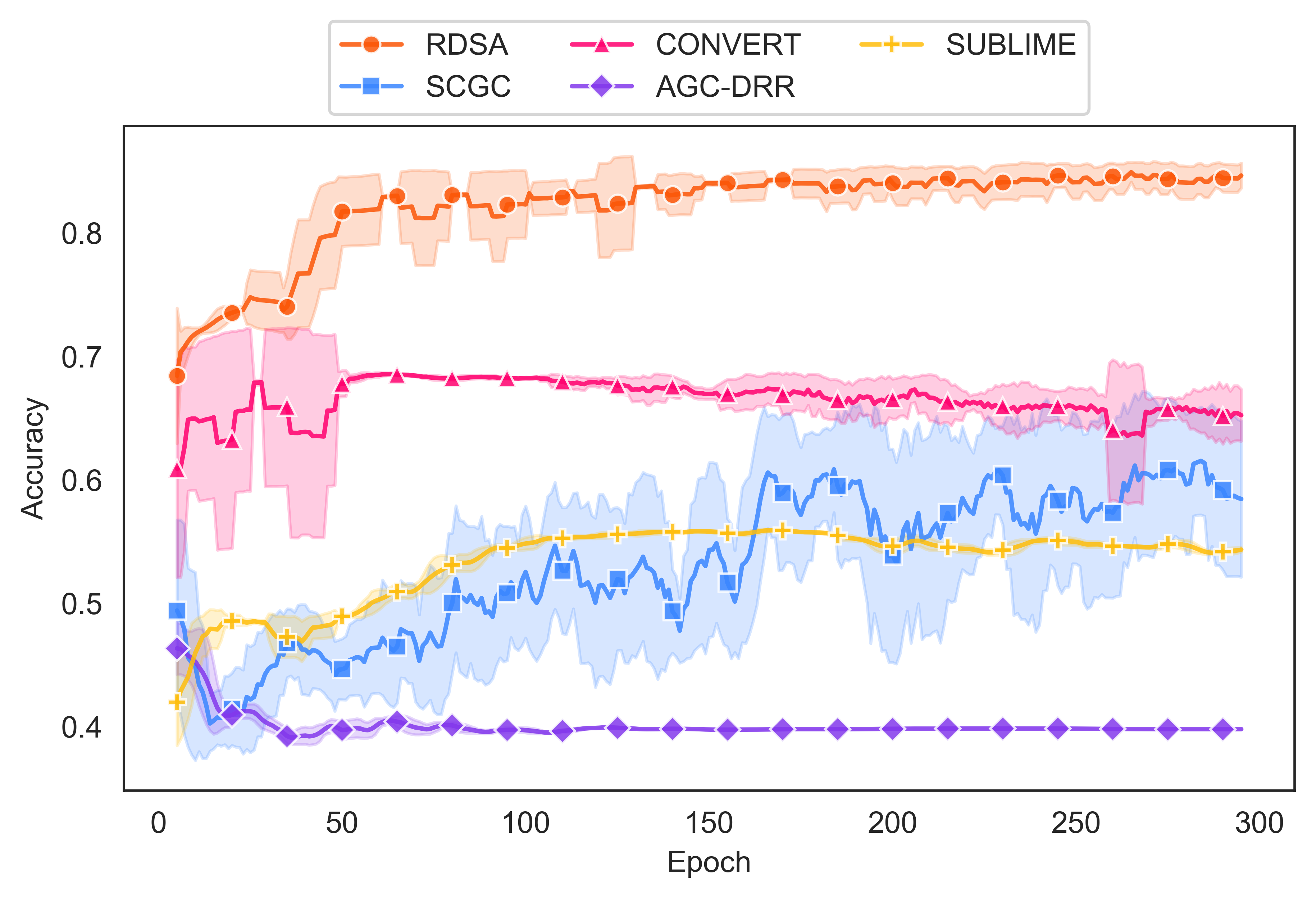}
    \includegraphics[width=0.48\textwidth]{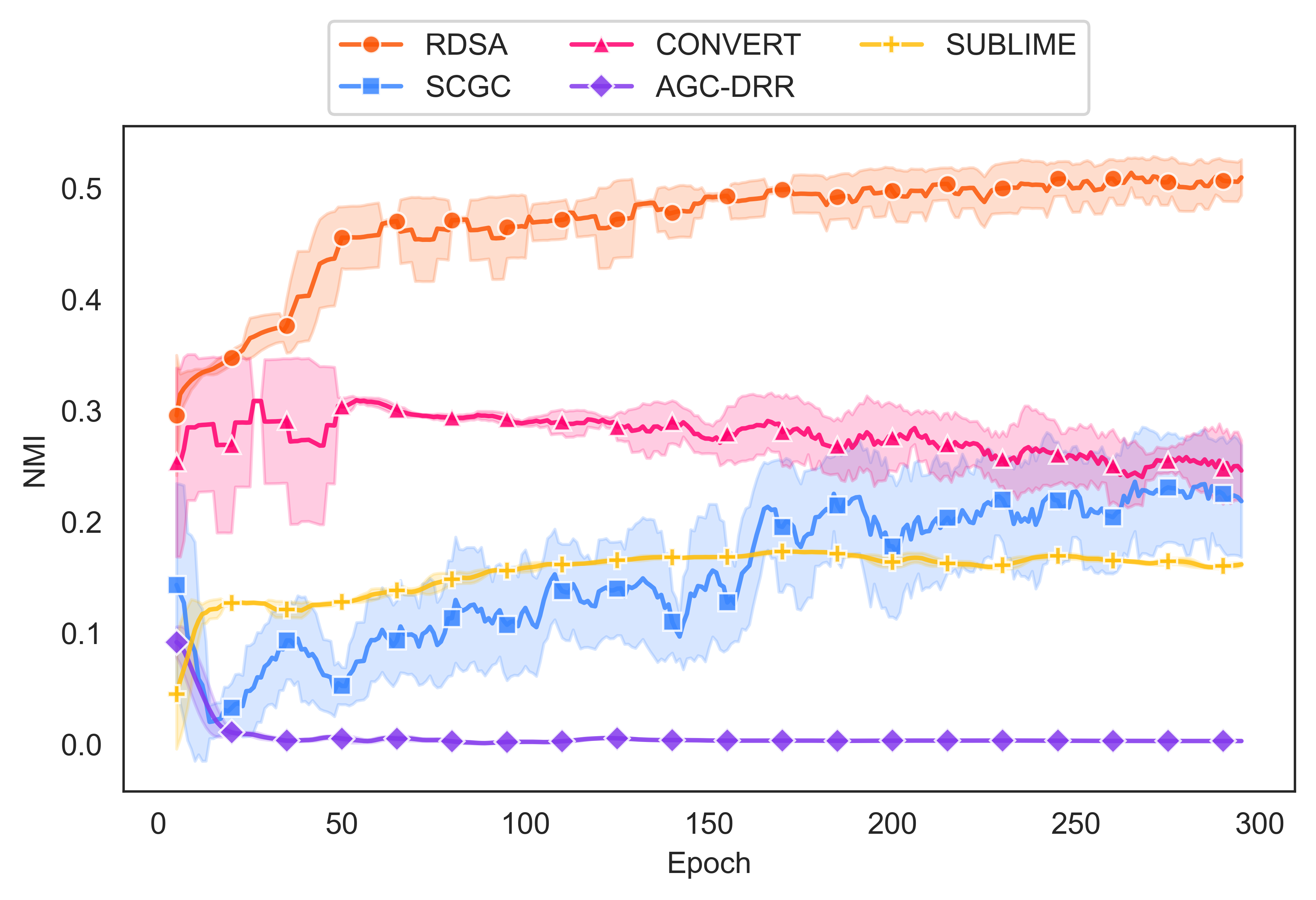} \\
    \includegraphics[width=0.48\textwidth]{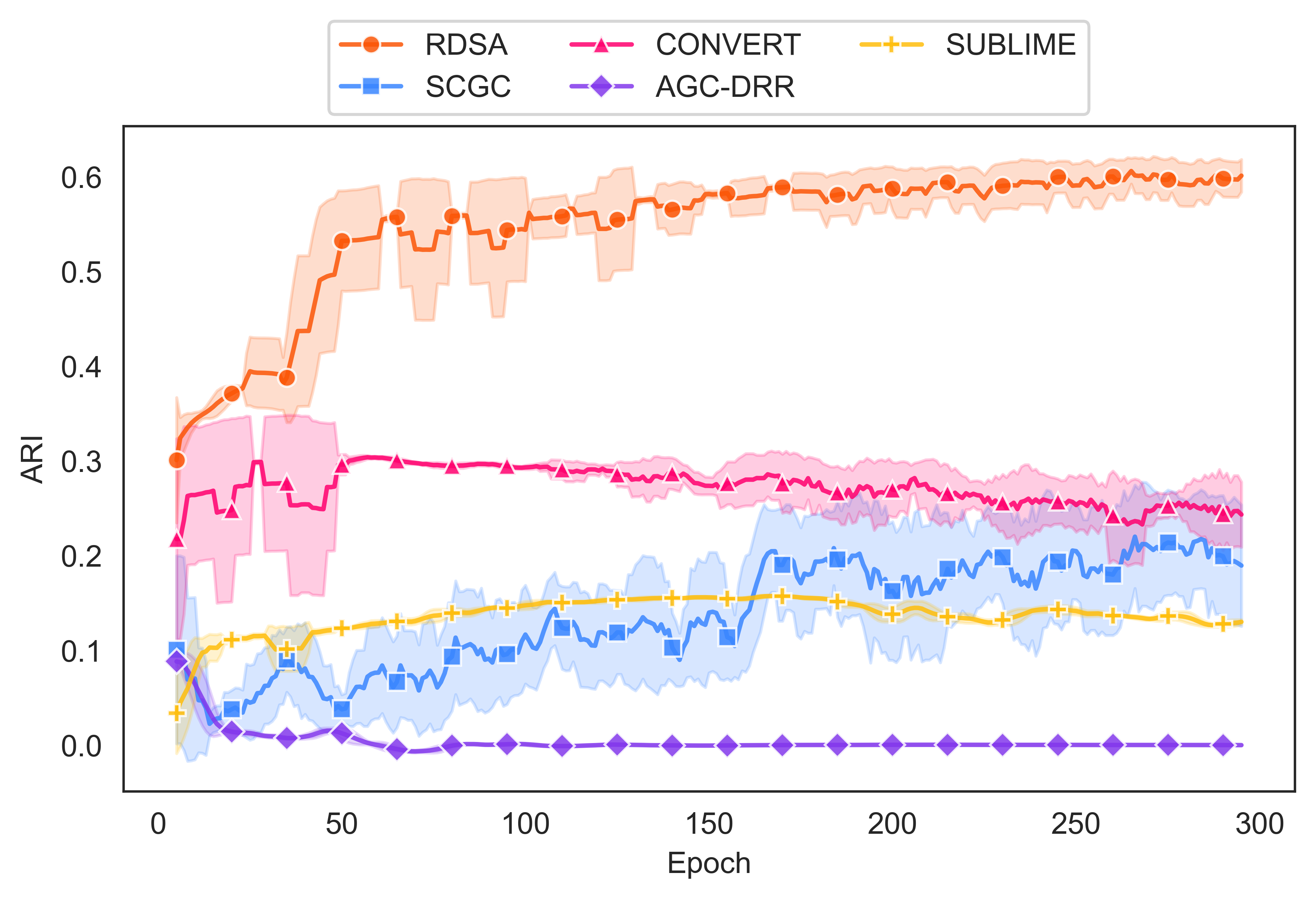}
    \includegraphics[width=0.48\textwidth]{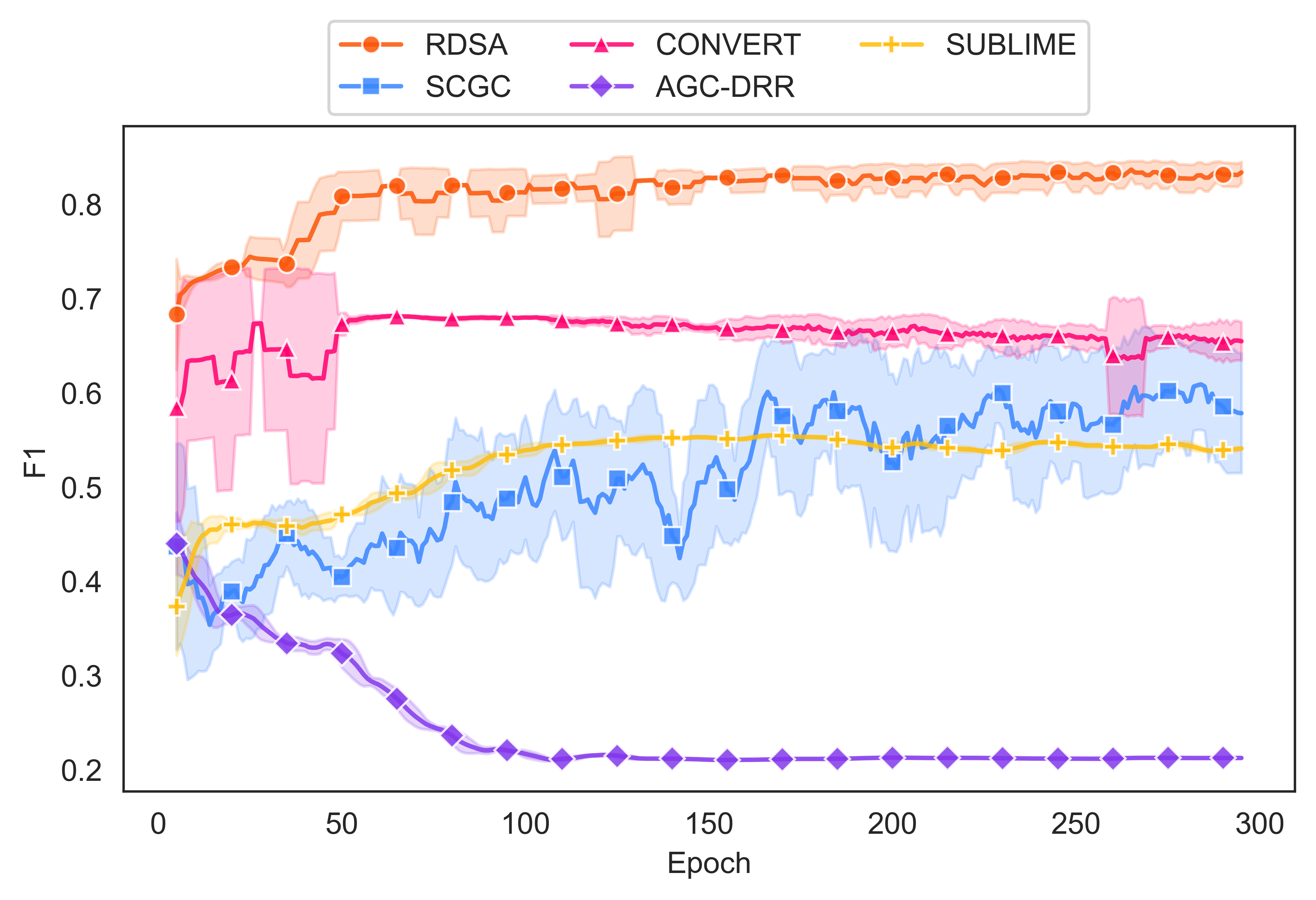}
  \end{minipage}
  \begin{minipage}[t]{0.33\linewidth}
    \centering
    \caption*{Amazon Computers}
    \includegraphics[width=0.48\textwidth]{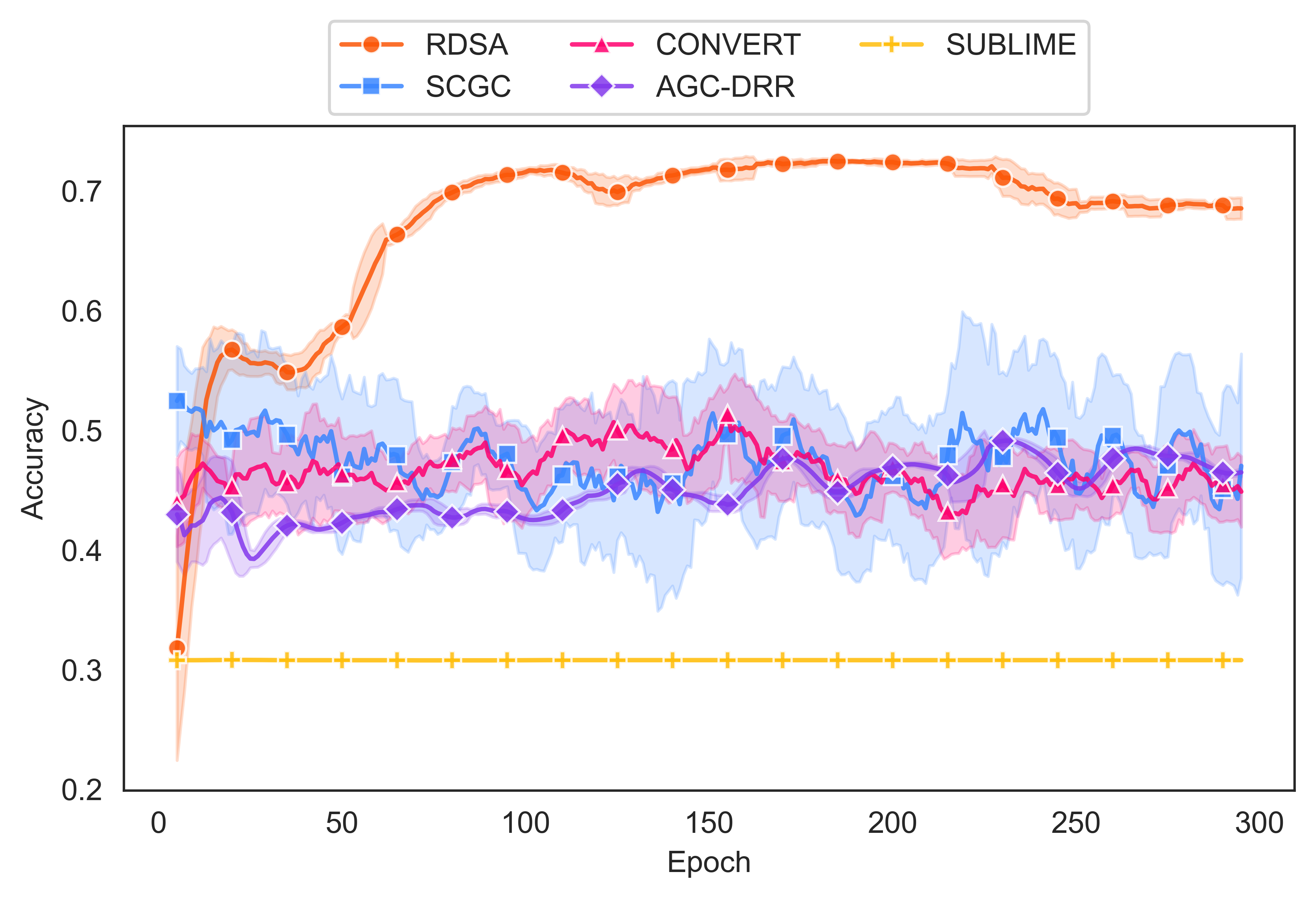}
    \includegraphics[width=0.48\textwidth]{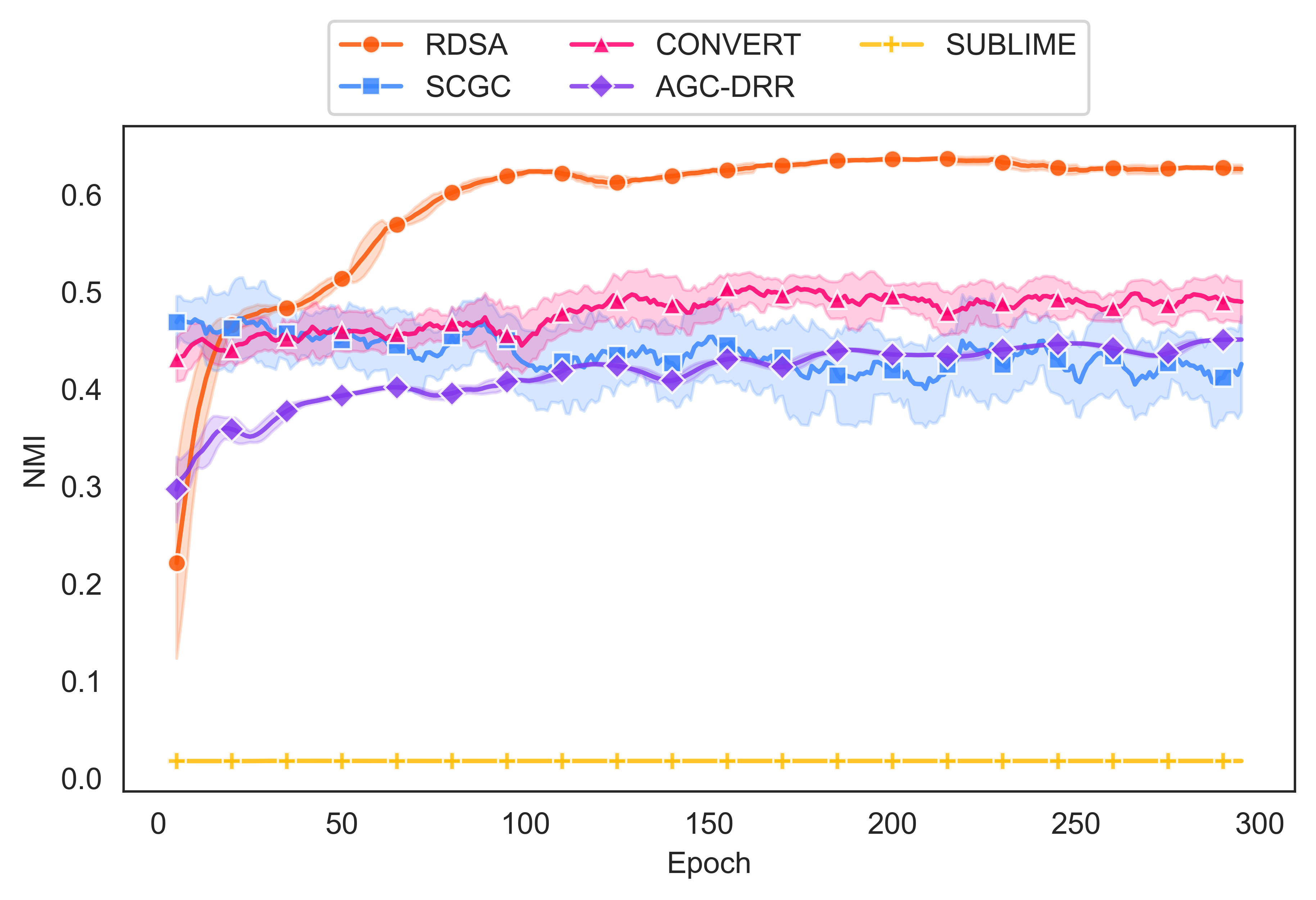} \\
    \includegraphics[width=0.48\textwidth]{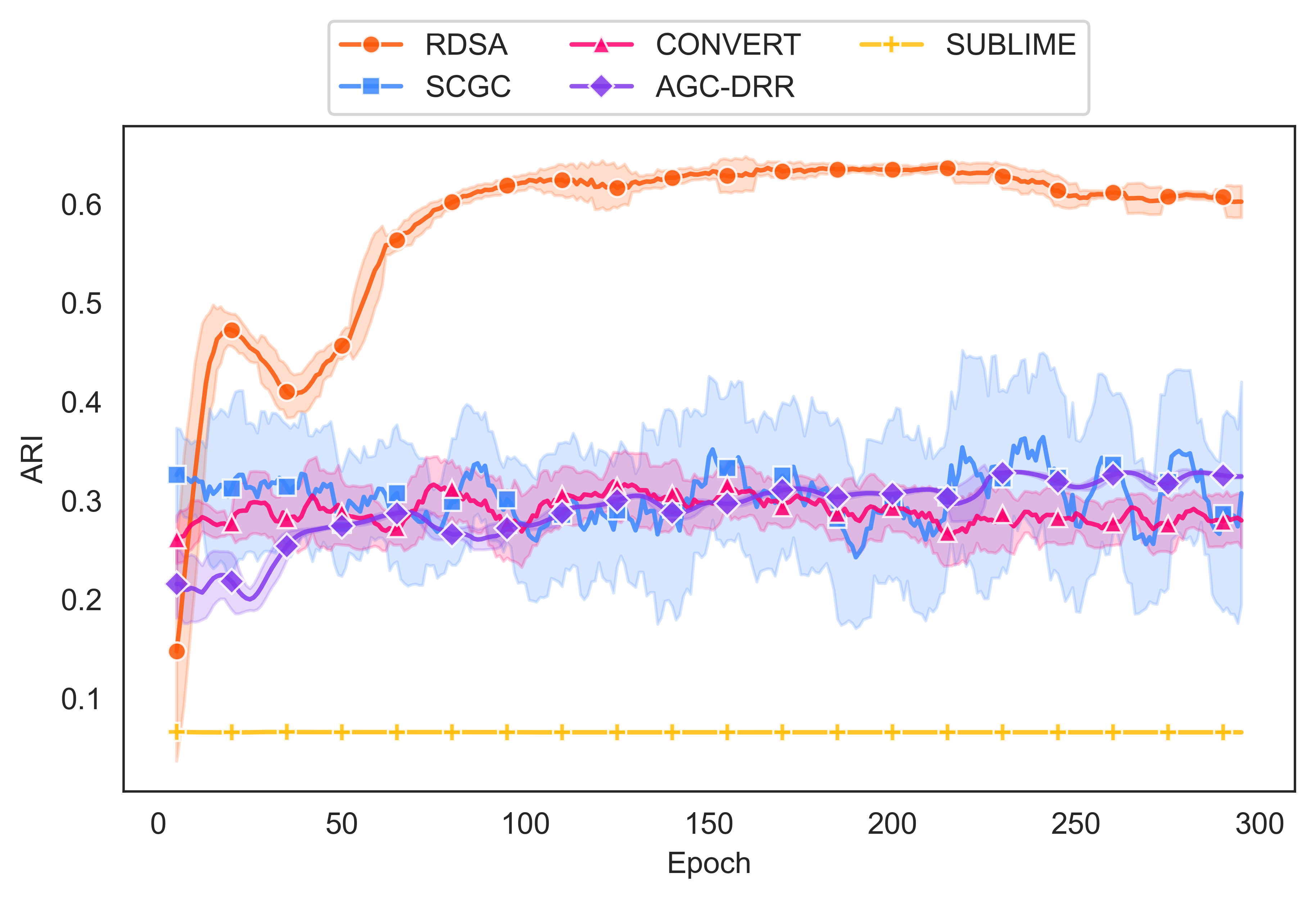}
    \includegraphics[width=0.48\textwidth]{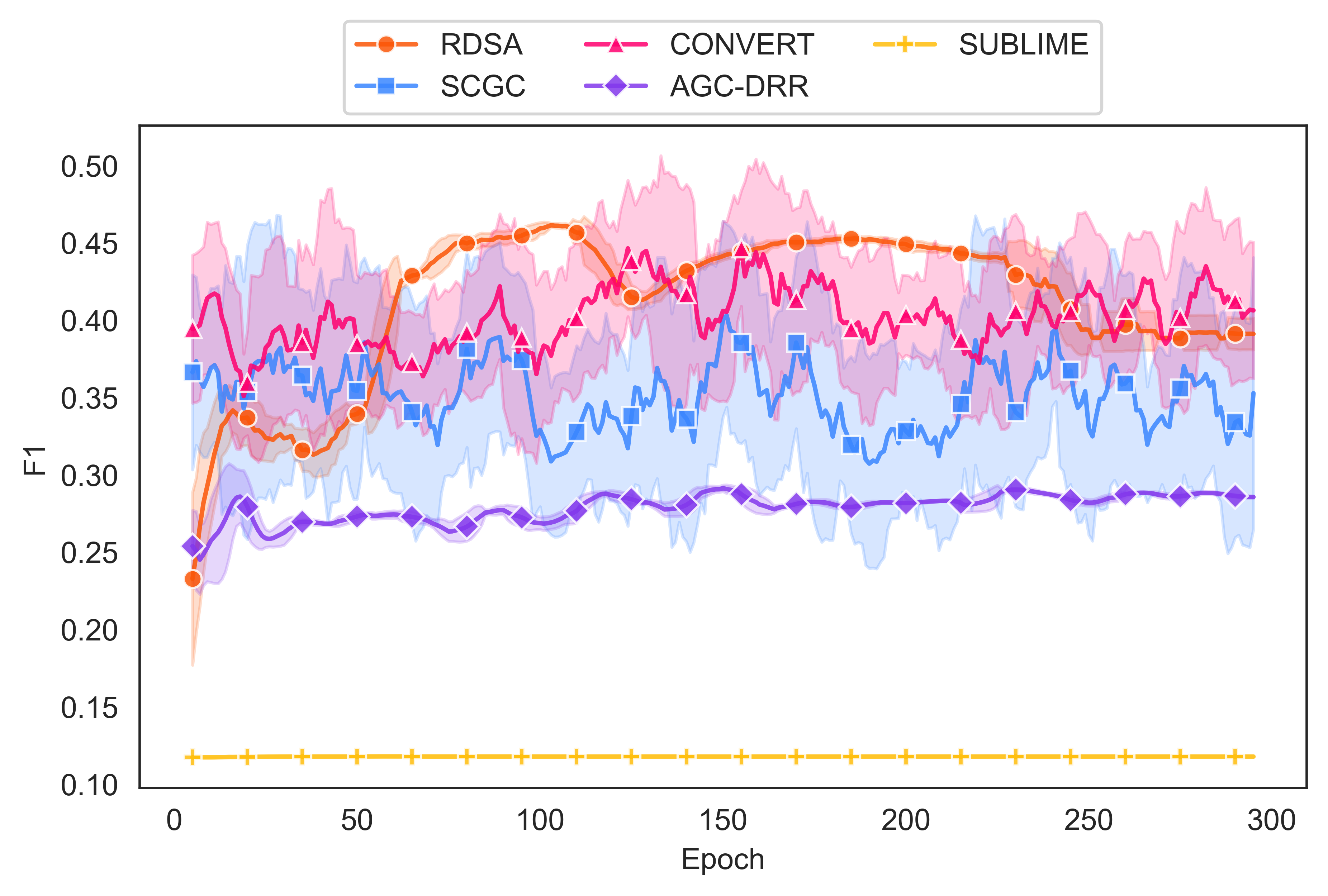}
  \end{minipage}
\end{figure*}

\subsection{Robustness Analysis}
In this section, we evaluate the robustness of our proposed method by comparing it with other state-of-the-art methods from the perspective of noise adaptability, model stability, and model scalability.
\subsubsection*{\textbf{Noise Adaptability Analysis}}
\label{sec:adaptability}
We evaluated our proposed method and three baseline approaches with denoising blocks across three datasets characterized by varying noise levels. The summarized results are presented in Table \ref{table:robustness}. Across all datasets, our method consistently achieves superior performance, highlighted by the best scores in nearly all metrics, especially under higher noise levels. For instance, under the highest noise level (III) on the Cora dataset, our method achieves an ACC of $75.63\%$, which is a mere $8\%$ degradation compared to its performance in the base experiment. In contrast, the second-best method, CONVERT, experiences a $17\%$ degradation, with an ACC of $61.18\%$. Similarly, on the PubMed dataset under noise level II, our method shows an ACC of $83.28\%$, with a minimal $2.5\%$ degradation, significantly outperforming the second-best method, which has an ACC of $58.99\%$ and a $15.3\%$ degradation. This trend is consistent across the other metrics and datasets. These results clearly demonstrate the robustness of our method against noisy data, maintaining high performance with minimal degradation compared to the baseline methods. This adaptability to noise is crucial for real-world applications where data is often imperfect, further validating the effectiveness of our approach.

\begin{table}
  \centering
  \caption{Performance comparison with state-of-the-art methods on three datasets with different noise levels, \besthighlight{red} indicates the best performance, and \secondhighlight{blue} indicates the second-best performance. The values in parentheses represent the degradation percentage compared to the base experiment.}
  \resizebox{\linewidth}{!}{
  \label{table:robustness}
    \begin{tabular}{c|c|c|cccc}
      \hline
      \multirow{2}*{\textbf{dataset}} & \multirow{2}*{\makecell{\textbf{Noise} \\ \textbf{Level}}} & \multirow{2}*{\textbf{Metrics}} & \multicolumn{4}{c}{\textbf{Methods}} \\
      \cline{4-7}
      & & & \textbf{SCGC} & \textbf{MAGI} & \textbf{CONVERT} & \textbf{RDSA} \\
      \hline
      \multirow{12}*{\textbf{Cora}} & \multirow{4}*{I} & \textbf{ACC} & \secondhighlight{$67.84$} $(-9\%)$ & $65.67$ $(-13\%)$ & $67.31$ $(-9\%)$ & \besthighlight{$77.40$} $(-5\%)$ \\
      & & \textbf{NMI} & $44.86$ $(-21\%)$ & $43.52$ $(-27\%)$ & \secondhighlight{$44.92$} $(-19\%)$ & \besthighlight{$60.78$} $(-12\%)$ \\
      & & \textbf{ARI} & $40.74$ $(-22\%)$ & $40.27$ $(-30\%)$ & \secondhighlight{$40.83$} $(-19\%)$ & \besthighlight{$65.10$} $(-8\%)$ \\
      & & \textbf{F1} & $64.04$ $(-10\%)$ & $58.87$ $(-20\%)$ & \secondhighlight{$64.05$} $(-12\%)$ & \besthighlight{$71.39$} $(-8\%)$ \\
      \cline{2-7}
      & \multirow{4}*{II} & \textbf{ACC} & $56.98$ $(-23\%)$ & $57.51$ $(-19\%)$ & \secondhighlight{$63.73$} $(-14\%)$ & \besthighlight{$76.81$} $(-6\%)$ \\
      & & \textbf{NMI} & $35.54$ $(-22\%)$ & $33.60$ $(-44\%)$ & \secondhighlight{$38.58$} $(-31\%)$ & \besthighlight{$56.55$} $(-18\%)$ \\
      & & \textbf{ARI} & $31.55$ $(-39\%)$ & $25.15$ $(-56\%)$ & \secondhighlight{$34.85$} $(-31\%)$ & \besthighlight{$61.61$} $(-13\%)$ \\
      & & \textbf{F1} & $50.55$ $(-28\%)$ & $52.47$ $(-29\%)$ & \secondhighlight{$60.50$} $(-17\%)$ & \besthighlight{$71.00$} $(-8\%)$ \\
      \cline{2-7}
      & \multirow{4}*{III} & \textbf{ACC} & $53.18$ $(-28\%)$ & $42.65$ $(-44\%)$ & \secondhighlight{$61.18$} $(-17\%)$ & \besthighlight{$75.63$} $(-8\%)$ \\
      & & \textbf{NMI} & $28.70$ $(-49\%)$ & $21.00$ $(-65\%)$ & \secondhighlight{$34.39$} $(-38\%)$ & \besthighlight{$53.82$} $(-22\%)$ \\
      & & \textbf{ARI} & $21.81$ $(-58\%)$ & $13.83$ $(-76\%)$ & \secondhighlight{$31.17$} $(-38\%)$ & \besthighlight{$59.31$} $(-16\%)$ \\
      & & \textbf{F1} & $52.98$ $(-25\%)$ & $42.79$ $(-42\%)$ & \secondhighlight{$59.57$} $(-18\%)$ & \besthighlight{$71.53$} $(-7\%)$ \\
      \hline
      \multirow{12}*{\textbf{PubMed}} & \multirow{4}*{I} & \textbf{ACC} & $53.78$ $(-20\%)$ & $43.39$ $(-32\%)$ & \secondhighlight{$64.30$} $(-7.6\%)$ & \besthighlight{$83.90$} $(-1.8\%)$ \\
      & & \textbf{NMI} & $17.21$ $(-44\%)$ & $06.67$ $(-73.4\%)$ & \secondhighlight{$20.49$} $(-31.6\%)$ & \besthighlight{$48.89$} $(-6.7\%)$ \\
      & & \textbf{ARI} & $15.19$ $(-49\%)$ & $02.86$ $(-87.6\%)$ & \secondhighlight{$22.45$} $(-25.4\%)$ & \besthighlight{$58.11$} $(-5.9\%)$ \\
      & & \textbf{F1} & $50.78$ $(-24\%)$ & $34.82$ $(-45.1\%)$ & \secondhighlight{$63.98$} $(-6.0\%)$ & \besthighlight{$82.77$} $(-1.8\%)$ \\
      \cline{2-7}
      & \multirow{4}*{II} & \textbf{ACC} & $49.79$ $(-26.3\%)$ & $41.51$ $(-34.9\%)$ & \secondhighlight{$58.99$} $(-15.3\%)$ & \besthighlight{$83.28$} $(-2.5\%)$ \\
      & & \textbf{NMI} & $12.12$ $(-60.5\%)$ & $04.63$ $(-81.6\%)$ & \secondhighlight{$14.75$} $(-50.8\%)$ & \besthighlight{$47.54$} $(-9.3\%)$ \\
      & & \textbf{ARI} & $11.62$ $(-60.9\%)$ & $01.51$ $(-93.4\%)$ & \secondhighlight{$15.89$} $(-46.4\%)$ & \besthighlight{$56.63$} $(-8.3\%)$ \\
      & & \textbf{F1} & $46.54$ $(-30.9\%)$ & $33.45$ $(-47.2\%)$ & \secondhighlight{$58.39$} $(-8.6\%)$ & \besthighlight{$82.19$} $(-2.5\%)$ \\
      \cline{2-7}
      & \multirow{4}*{III} & \textbf{ACC} & $46.71$ $(-30.8\%)$ & $40.85$ $(-35.9\%)$ & \secondhighlight{$55.02$} $(-20.9\%)$ & \besthighlight{$83.19$} $(-2.6\%)$ \\
      & & \textbf{NMI} & $03.58$ $(-88.3\%)$ & $03.44$ $(-86.3\%)$ & \secondhighlight{$14.25$} $(-52.5\%)$ & \besthighlight{$47.52$} $(-6.5\%)$ \\
      & & \textbf{ARI} & $03.20$ $(-89.2\%)$ & $01.19$ $(-94.8\%)$ & \secondhighlight{$11.95$} $(-60\%)$ & \besthighlight{$56.42$} $(-8.6\%)$ \\
      & & \textbf{F1} & $37.16$ $(-44.8\%)$ & $33.45$ $(-47.2\%)$ & \secondhighlight{$54.23$} $(-20.2\%)$ & \besthighlight{$71.53$} $(-15.2\%)$ \\
      \hline
      \multirow{12}*{\makecell{\textbf{Amazon}\\\textbf{PC}}} & \multirow{4}*{I} & \textbf{ACC} & $39.11$ $(-37.3\%)$ & \secondhighlight{$48.40$} $(-21.9\%)$ & $47.58$ $(-13.9\%)$ & \besthighlight{$51.29$} $(-27.3\%)$ \\
      & & \textbf{NMI} & $28.96$ $(-43.8\%)$ & \besthighlight{$42.94$} $(-27.5\%)$ & $35.63$ $(-30.6\%)$ & \secondhighlight{$41.83$} $(-33.6\%)$ \\
      & & \textbf{ARI} & $22.31$ $(-53.9\%)$ & \secondhighlight{$29.48$} $(-36.2\%)$ & $26.45$ $(-26.4\%)$ & \besthighlight{$56.02$} $(-8.2\%)$ \\
      & & \textbf{F1} & $28.66$ $(-43.4\%)$ & \secondhighlight{$43.08$} $(-25.0\%)$ & $39.02$ $(-19.7\%)$ & \besthighlight{$43.28$} $(-14.6\%)$ \\
      \cline{2-7}
      & \multirow{4}*{II} & \textbf{ACC} & $34.95$ $(-44.0\%)$ & \secondhighlight{$43.72$} $(-29.5\%)$ & $40.43$ $(-26.9\%)$ & \besthighlight{$49.10$} $(-30.4\%)$ \\
      & & \textbf{NMI} & $20.33$ $(-60.6\%)$ & \secondhighlight{$38.42$} $(-35.1\%)$ & $25.97$ $(-49.4\%)$ & \besthighlight{$39.59$} $(-37.2\%)$ \\
      & & \textbf{ARI} & $10.18$ $(-79.0\%)$ & \secondhighlight{$24.47$} $(-47.0\%)$ & $20.13$ $(-56.4\%)$ & \besthighlight{$50.80$} $(-16.8\%)$ \\
      & & \textbf{F1} & $22.44$ $(-55.7\%)$ & \secondhighlight{$36.67$} $(-36.1\%)$ & $30.54$ $(-37.2\%)$ & \besthighlight{$38.03$} $(-24.9\%)$ \\
      \cline{2-7}
      & \multirow{4}*{III} & \textbf{ACC} & $27.87$ $(-55.4\%)$ & \secondhighlight{$42.27$} $(-31.8\%)$ & $31.10$ $(-43.7\%)$ & \besthighlight{$45.00$} $(-36.2\%)$ \\
      & & \textbf{NMI} & $19.47$ $(-62.3\%)$ & \secondhighlight{$31.28$} $(-47.2\%)$ & $15.60$ $(-69.6\%)$ & \besthighlight{$37.01$} $(-41.3\%)$ \\
      & & \textbf{ARI} & $11.85$ $(-75.5\%)$ & \secondhighlight{$22.34$} $(-51.6\%)$ & $13.15$ $(-78.5\%)$ & \besthighlight{$45.02$} $(-26.3\%)$ \\
      & & \textbf{F1} & $19.79$ $(-60.9\%)$ & \besthighlight{$34.15$} $(-40.5\%)$ & $19.71$ $(-59.4\%)$ & \secondhighlight{$30.08$} $(-40.6\%)$ \\
      \hline
    \end{tabular}
  }
\end{table}

\subsubsection*{\textbf{Stability Analysis}}
We evaluate the stability of our proposed method with the four best-performing baselines. We illustrate the line plot of the change of four metrics with the number of epochs in Figure \ref{fig:stability}. Although SCGC and CONVERT have high peaks in all metrics, they are not stable in the training process. We can't ensure that the model can converge to the optimal solution. AGC-DRR and SUBLIME, on the other hand, demonstrate consistent performance during training. However, AGC-DRR displays subpar performance on specific datasets, occasionally exhibiting metrics declines throughout training. In contrast, SUBLIME consistently delivers stable performance across all datasets, albeit not always achieving the highest metrics. Our proposed method consistently maintains stability across datasets and metrics, often outperforming the alternatives. These findings underscore the efficacy and reliability of our approach.

\begin{figure*}[ht]
  \begin{minipage}[t]{0.45\linewidth}
    \centering
    \caption{Performance comparison of the full model and its variants on four datasets (Cora, Citeseer, PubMed, and Amazon Computers). NL means the noise levels.}
    \label{figure:ablation}
    \includegraphics[width=0.49\textwidth]{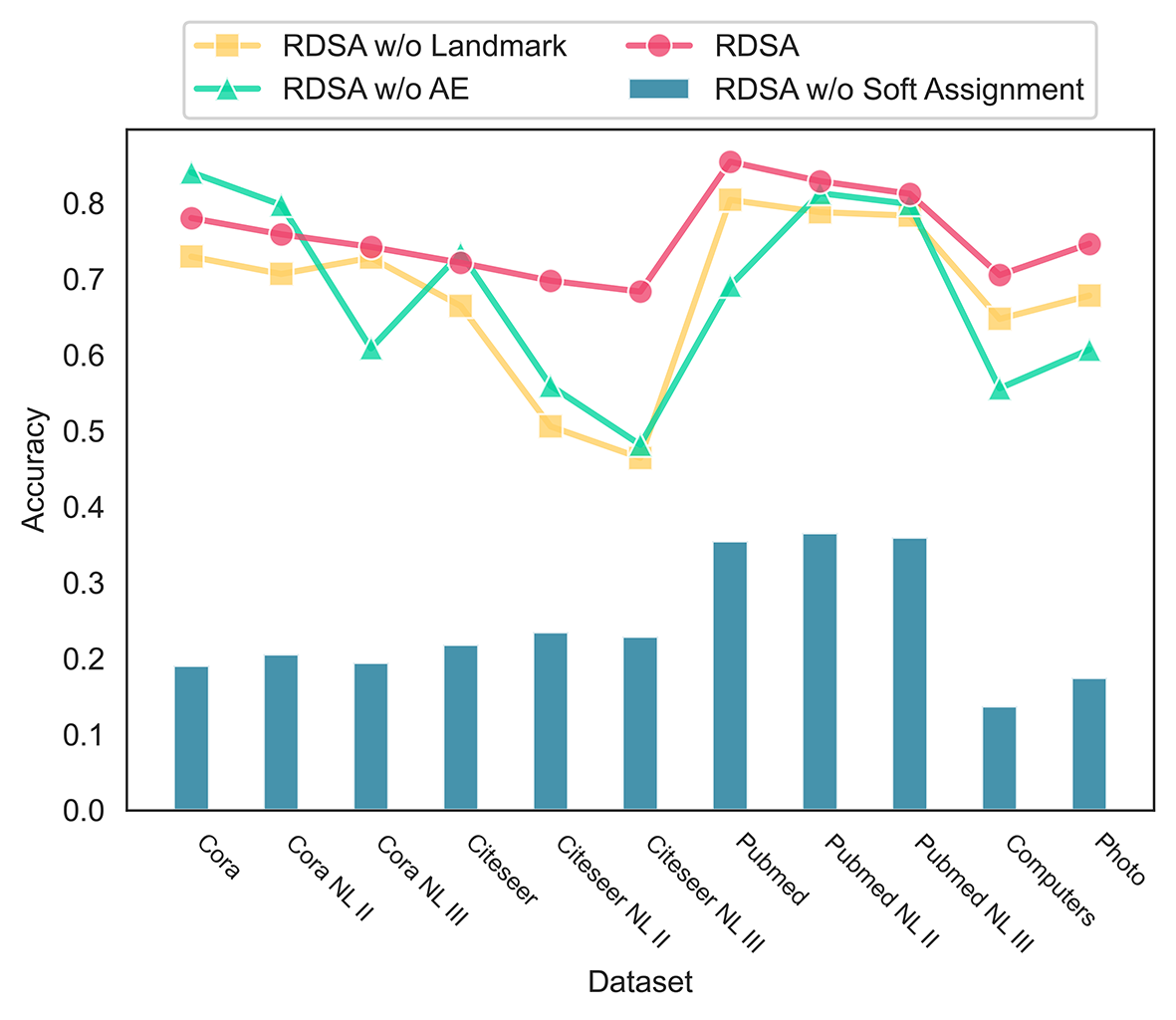}
    \includegraphics[width=0.49\textwidth]{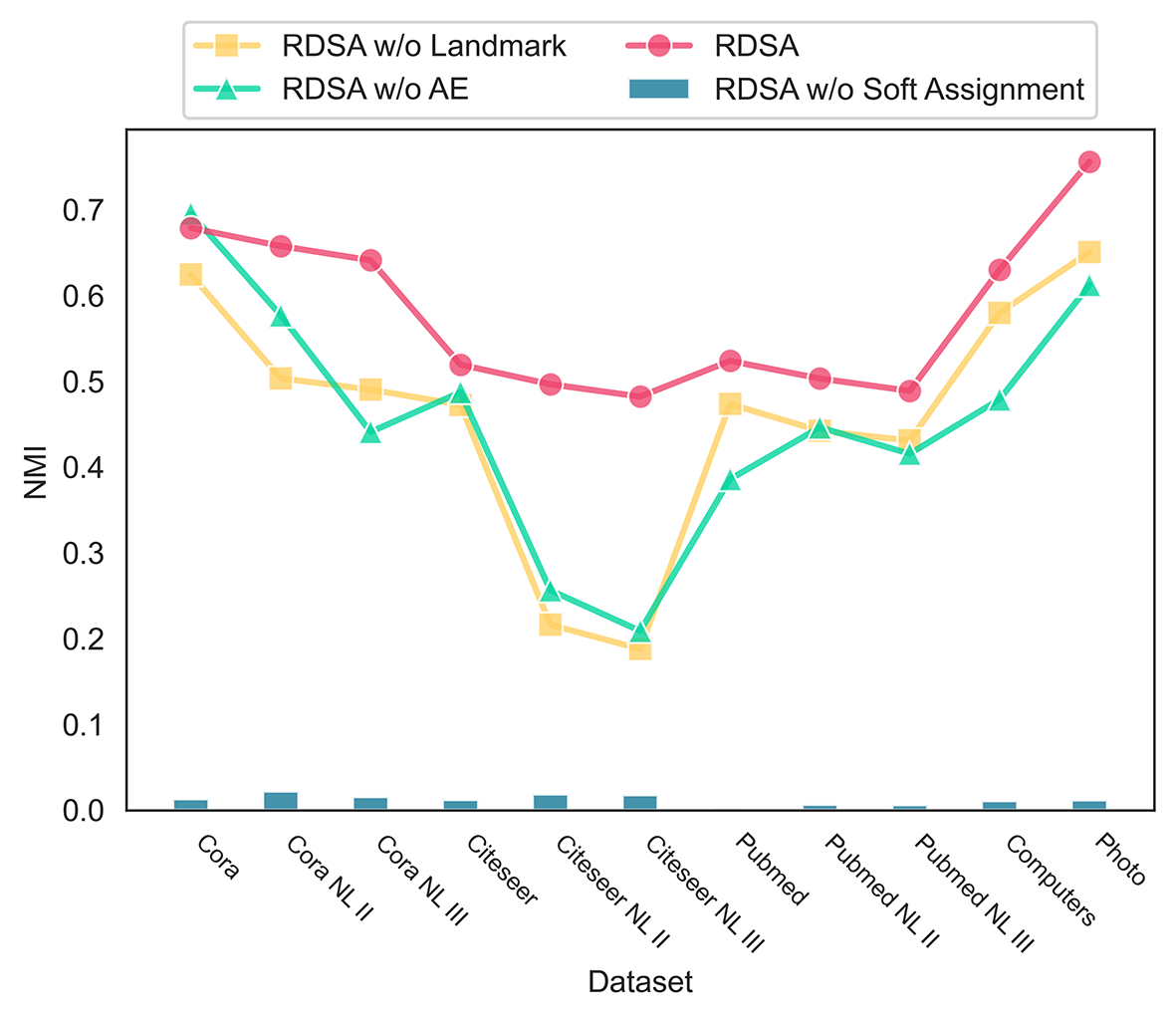} \\
    \includegraphics[width=0.49\textwidth]{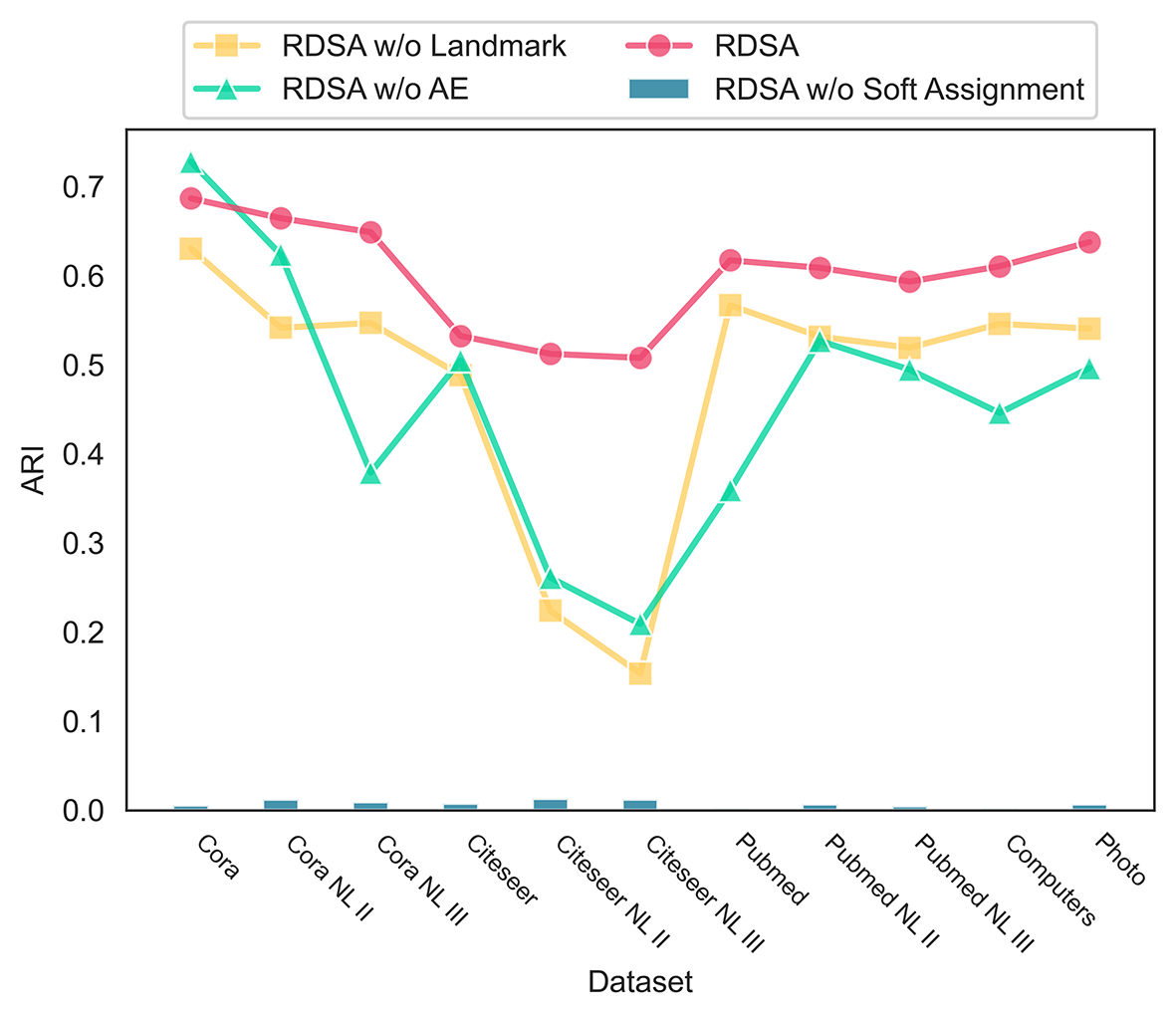}
    \includegraphics[width=0.49\textwidth]{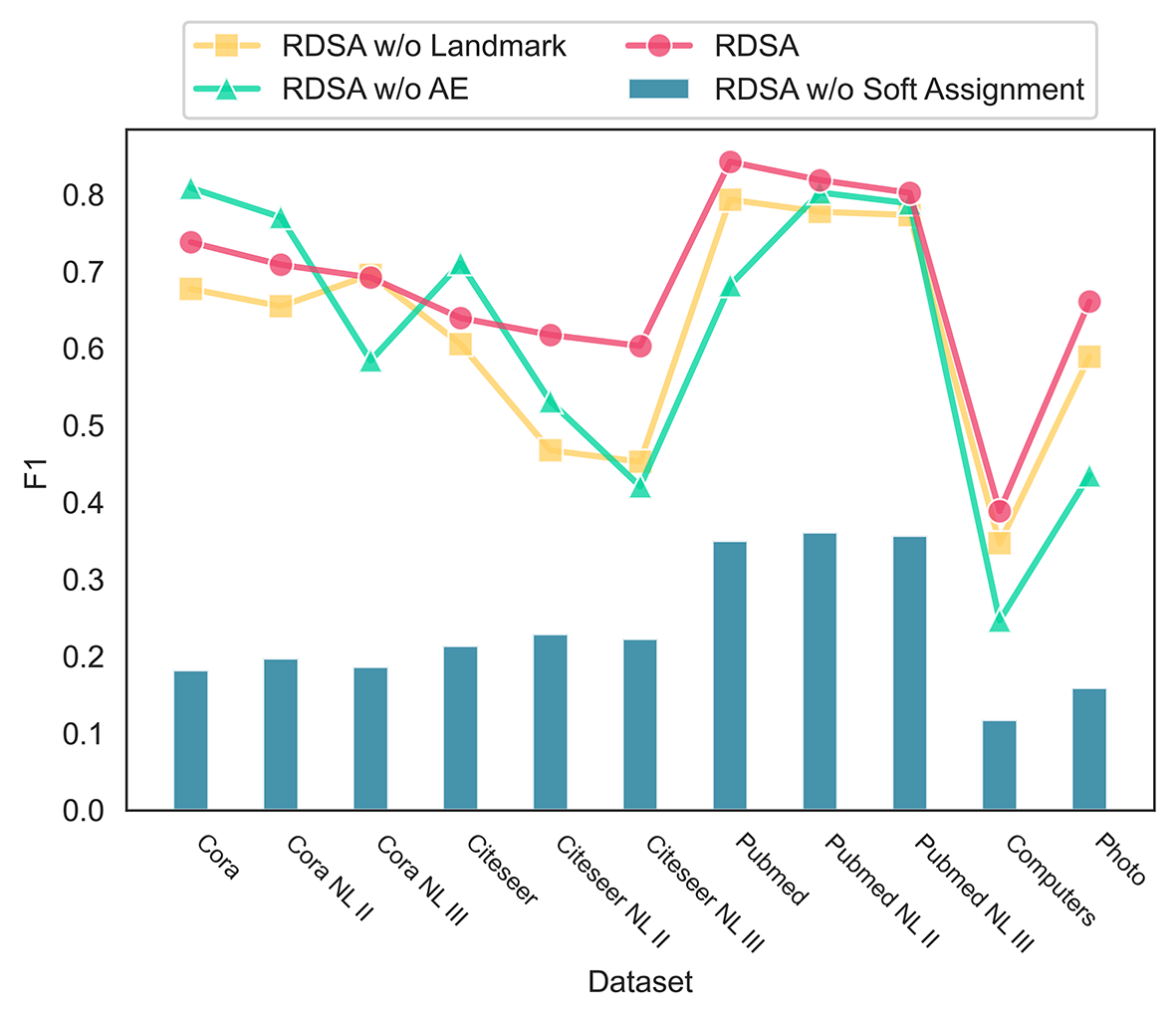}
  \end{minipage}
  \hspace{0.04\linewidth}
  \begin{minipage}[t]{0.48\linewidth}
    \centering
    \caption{The bar plot shows the hyperparameter analysis of the proposed framework on four datasets (Cora, Citeseer, PubMed, and Amazon Computers), different colors represent different datasets.}
    \label{figure:sensitivity}
    \includegraphics[width=0.49\textwidth]{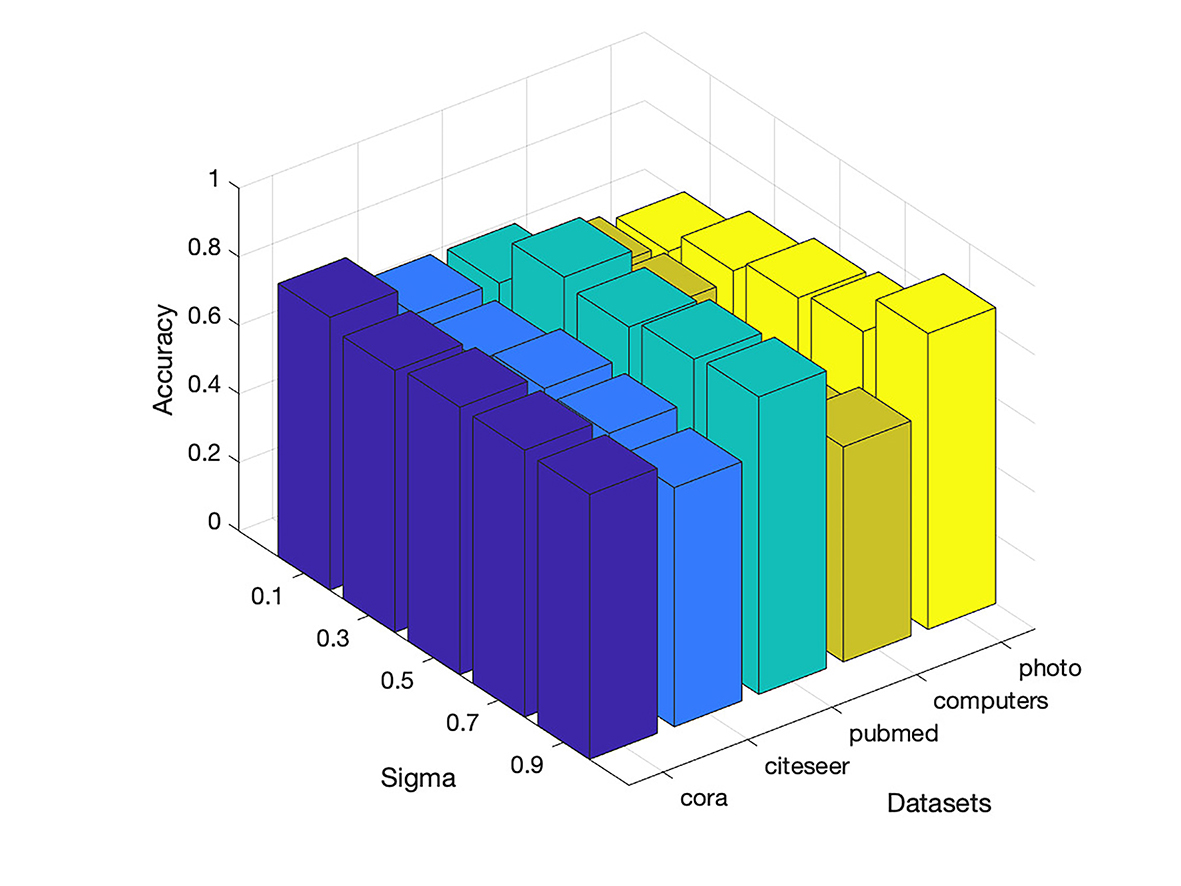}
    \includegraphics[width=0.49\textwidth]{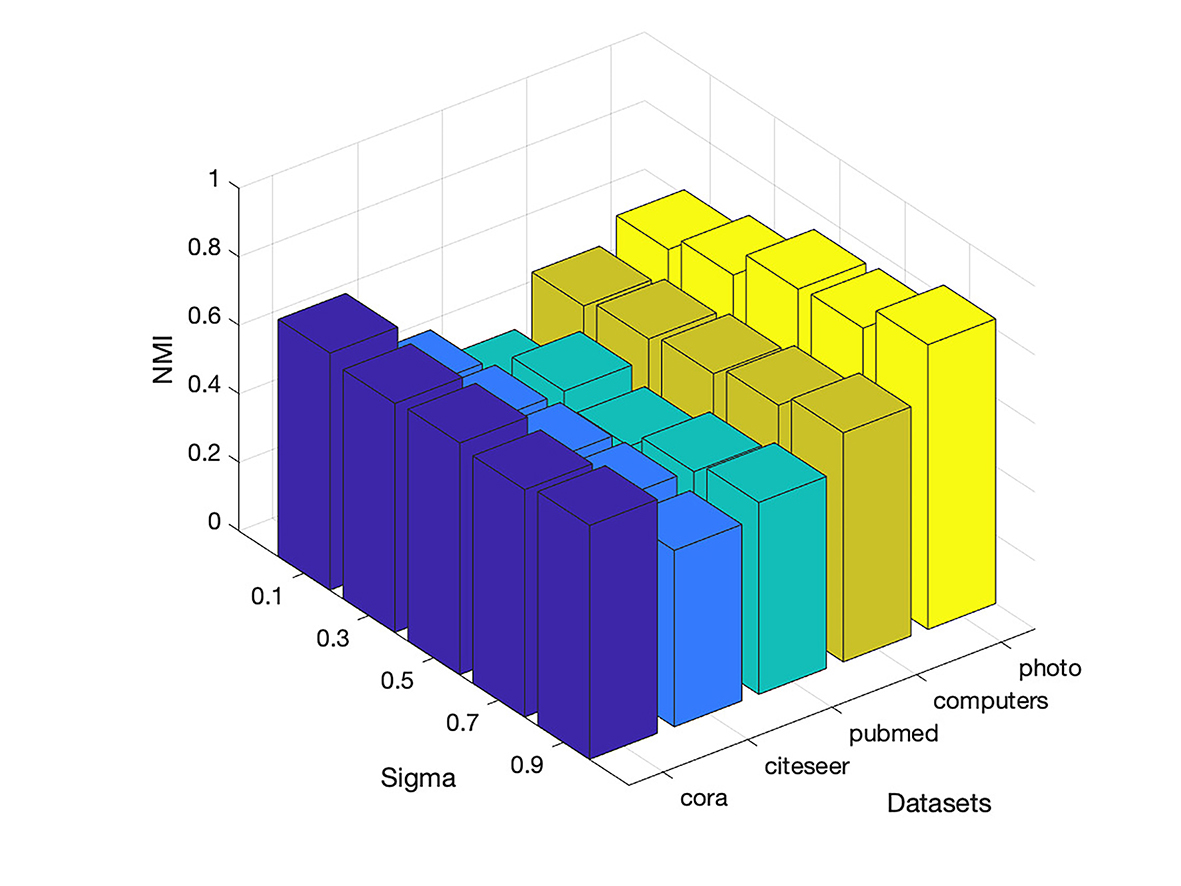} \\
    \includegraphics[width=0.49\textwidth]{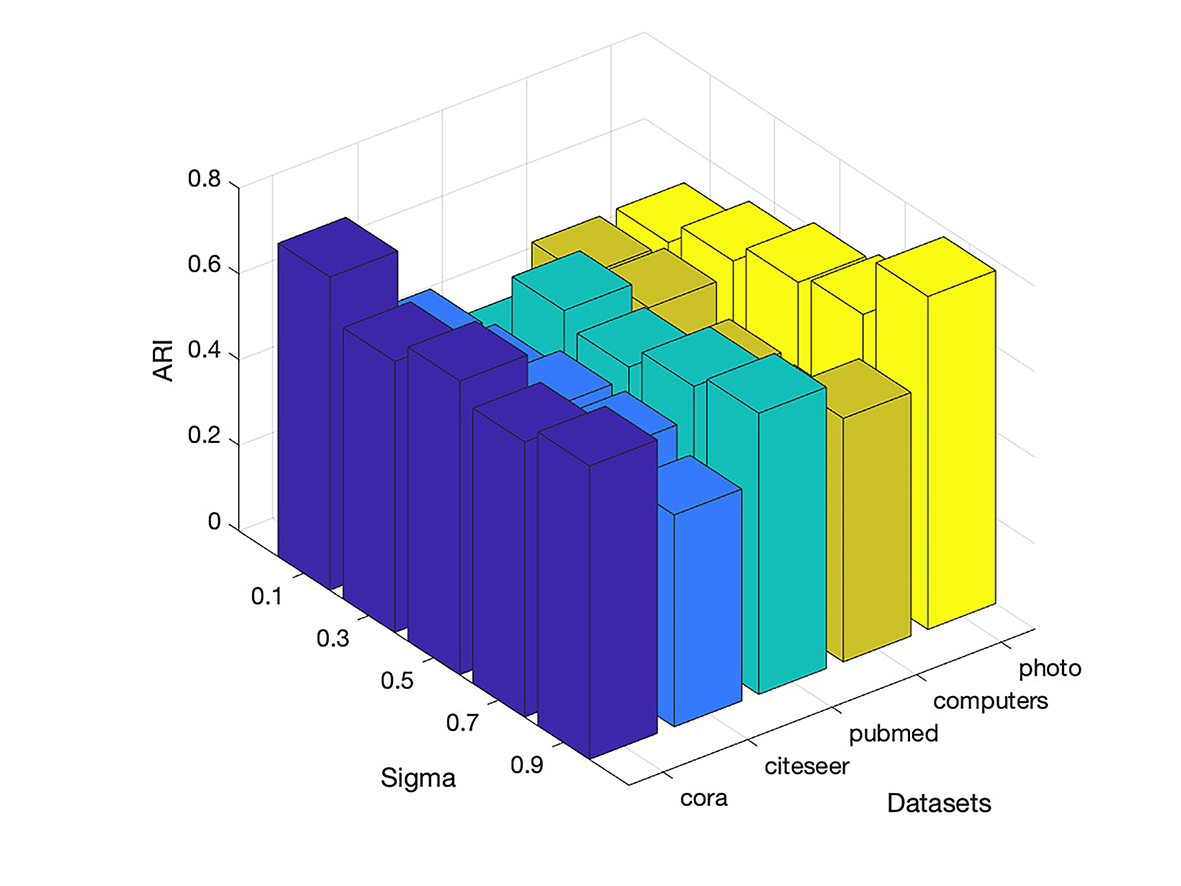}
    \includegraphics[width=0.49\textwidth]{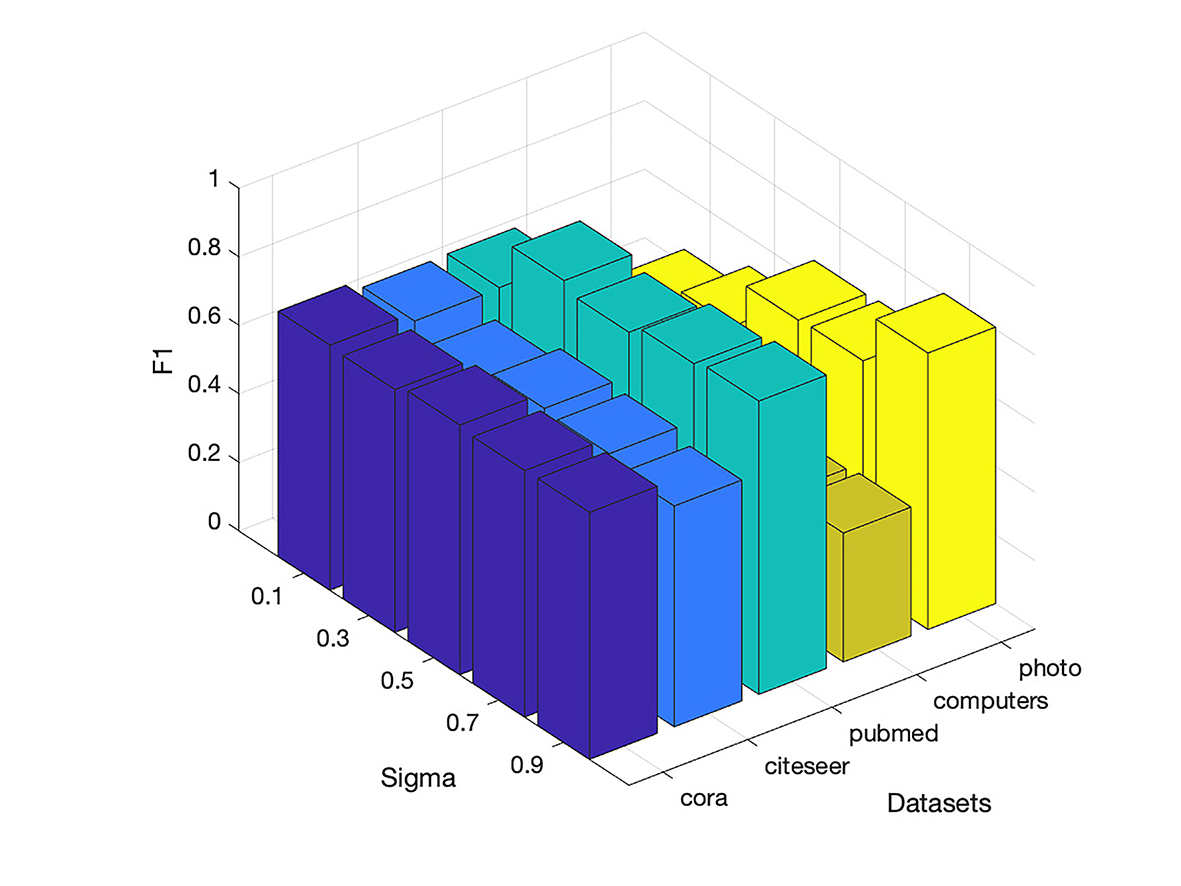}
  \end{minipage}
\end{figure*}

In summary, our robustness analysis comprehensively evaluates our proposed method against state-of-the-art approaches, focusing on noise adaptability, stability, scalability (Sec. \ref{sec:baseline}) and effectiveness (Sec. \ref{sec:baseline}). Regarding noise adaptability, our method consistently outperforms baselines across various datasets and noise levels, demonstrating its effectiveness in handling noisy data. Stability analysis reveals that while some alternatives exhibit sporadic peaks in performance, they lack consistency throughout training, whereas our method maintains stability across datasets and metrics, often surpassing others. Furthermore, our scalability analysis confirms the efficacy of our approach even on large-scale datasets, highlighting its robustness and versatility. Overall, these findings emphasize the superiority and reliability of our proposed method across different evaluation criteria, validating its potential for real-world applications.

\subsection{Ablation Studies}
In this section, we conduct ablation studies to assess the efficacy of the proposed components within our method. We juxtapose the performance of our complete model against variants lacking specific elements, namely the landmarks preservation soft assignment (RDSA w/o Landmark), the attribute embedding module (RDSA w/o AE), and both the modularity-optimized soft assignment and landmark preservation soft assignment (RDSA w/o Soft Assignment). The results are depicted in Figure \ref{figure:ablation}. Consistently, our full model outperforms its variants across all datasets and metrics, highlighting the significance of each constituent. Notably, the landmark preservation soft assignment bolsters the model's adaptability to noisy data, as indicated by the performance dip in RDSA w/o Landmarks under heightened noise levels. Furthermore, the attribute embedding module enhances the model's capacity to encapsulate node attributes, which is particularly beneficial for large-scale datasets like PubMed, Amazon Photo, and Computers. These findings underscore the efficacy of each component within our proposed method, underscoring their collective contributions to the model's superior performance.


\subsection{Hyperparameter Analysis}
In this section, we conduct a comprehensive analysis of the impact of hyperparameters on the performance of our proposed framework by systematically varying them within a reasonable range. The results, which are summarized in Figure \ref{figure:sensitivity}, indicate that the RDSA framework achieves optimal performance when the hyperparameter $\sigma$ is set between 0.3 and 0.7 across the majority of datasets. This range balances the emphasis on both structure and node information. However, when $\sigma$ values fall outside this range, performance tends to decline, either due to overemphasis on structural information or an over-reliance on node features. Notably, for the Amazon Photo dataset, which exhibits an imbalance between node and edge information, the optimal $\sigma$ is observed at 0.9, reflecting the need for a different balance between these factors in this specific context.

\section{CONCLUSION}
In this paper, we propose a novel robust deep graph clustering method, RDSA, which integrates structure based soft assignment, node based soft assignment, and attribute embedding to enhance the model's adaptability to noisy data, stability, and scalability. Extensive experiments on seven benchmark datasets, along with ablation studies, demonstrate the superior performance of our method compared to state-of-the-art approaches. These findings underscore its efficacy in handling noisy data, maintaining stability, and scaling to large datasets, while also validating the significance of each model component in achieving this performance. Future work will focus on extending our method to a broader range of graph-related tasks, such as node classification and link prediction, to further validate its versatility and robustness. Additionally, we plan to explore its application in real-world problems like recommendation systems, where graph-based representations play a crucial role, and molecular research, where accurate clustering of molecular structures is essential. These extensions will allow us to assess the practical impact of RDSA in diverse and complex domains.

\begin{acks}
This work is partially funded by the XJTLU Research Development Fund [Grant Number RDF-21-02-050] and the Young Data Scientist Program of the China National Astronomical Data Center [Grant Number NADC2023YDS08].
\end{acks}

\newpage
\bibliographystyle{ACM-Reference-Format}
\bibliography{ref}

\appendix

\end{document}